\PassOptionsToPackage{table,xcdraw}{xcolor}
\documentclass[10pt]{article} 
\usepackage[accepted]{tmlr}

\usepackage{amsmath,amsfonts,bm}

\def\eqref#1{equation~\ref{#1}}

\def\1{\bm{1}}

\DeclareMathAlphabet{\mathsfit}{\encodingdefault}{\sfdefault}{m}{sl}
\SetMathAlphabet{\mathsfit}{bold}{\encodingdefault}{\sfdefault}{bx}{n}

\usepackage{hyperref}
\usepackage{url}
\usepackage{graphicx}
\usepackage{amsmath}
\usepackage{amssymb}
\usepackage{amsthm}
\usepackage{booktabs}
\usepackage{multirow}
\usepackage{tabularx}
\usepackage[table,xcdraw]{xcolor}
\usepackage{subcaption}
\usepackage{epigraph}
\newif\ifshowrevisions
\showrevisionsfalse
\ifshowrevisions\colorlet{revisionblue}{blue}\else\colorlet{revisionblue}{black}\fi
\newcommand{\rev}[1]{{\color{revisionblue}#1}}
\hypersetup{hidelinks,pdftitle={Towards Generalist Game Agents: A Survey of Foundation Models in Virtual Worlds}}
\newtheorem{definition}{Definition}

\title{Towards Generalist Game Agents: A Survey of Foundation Models in Virtual Worlds}

\author{\name Kuan Zhang$^{1,\star}$, Dongchen Liu$^{1,\star}$, Qiyue Zhao$^{1,\star}$, Tianyu Xin$^{1,\star}$ \\
      \name Yue Su$^{2,\star}$, Haisheng Wang$^1$, Han Yin$^1$, Hongbo Ma$^1$ \\
      \name Peize Li$^1$, Tianjun Gu$^1$, Xiangnan Wu$^3$, Xinran Zhang$^1$ \\
      \name Yongxuan Li$^1$, Zirong Chen$^1$, Yiming Li$^{1,\dagger}$ \\[0.5ex]
      \addr $^1$ College of AI, Tsinghua University \\
      \addr $^2$ MMLab, The University of Hong Kong \\
      \addr $^3$ University of Chinese Academy of Sciences \\
      \addr \href{mailto:zhang-k26@mails.tsinghua.edu.cn}{zhang-k26@mails.tsinghua.edu.cn}, yimingli9702@mail.tsinghua.edu.cn \\
      \addr $^{\star}$ Equal contribution. $^{\dagger}$ Corresponding author.\normalfont\normalsize}

\def\month{10}  
\def\year{2026} 
\def\openreview{\url{https://openreview.net/forum?id=SDkEq8Fzvl}}

\begin{document}

\maketitle

\begin{abstract}
The real world unfolds along a single set of physics laws, yet human intelligence demonstrates a remarkable capacity to generalize experiences from this singular physical existence into a multiverse of games, each governed by entirely different rules, aesthetics, physics, and objectives. This omni-reality adaptability is a hallmark of general intelligence. As Artificial Intelligence progresses towards Artificial General Intelligence, the multiverse of games has evolved from mere entertainment into the ultimate ground for training and evaluating AGI. The pursuit of this generality has unfolded across four eras: from environment-specific symbolic and reinforcement learning agents, to current large foundation models acting as generalist players, and toward a future creator stage where the agent both creates new game worlds and continually evolves within them. In this survey, we trace the full lifecycle of a generalist game player along four interdependent pillars: Dataset, Model, Harness, and Benchmark. We distinguish competence depth, coverage across games, and transfer, and use published performance comparisons to show how data, model, harness, and evaluation choices shape the resulting generality claims. Building on this end-to-end review, we chart a five-step roadmap, progressing from single-game mastery to the ultimate creator stage in which the agent simultaneously creates and evolves within the theoretical game multiverse. A consolidated synthesis connects each step to five recurring trade-offs, current limitations, priorities across the four pillars, and evaluations that could demonstrate progress.
\end{abstract}

\begin{center}
\small
\textbf{Code:} \url{https://github.com/THUSI-Lab/Awesome-LFMs-Play-Games} \\
\textbf{Project page:} \url{https://thusi-lab.github.io/Awesome-LFMs-Play-Games/}
\end{center}

\begin{figure}[t]
  \centering
  \includegraphics[width=\linewidth]{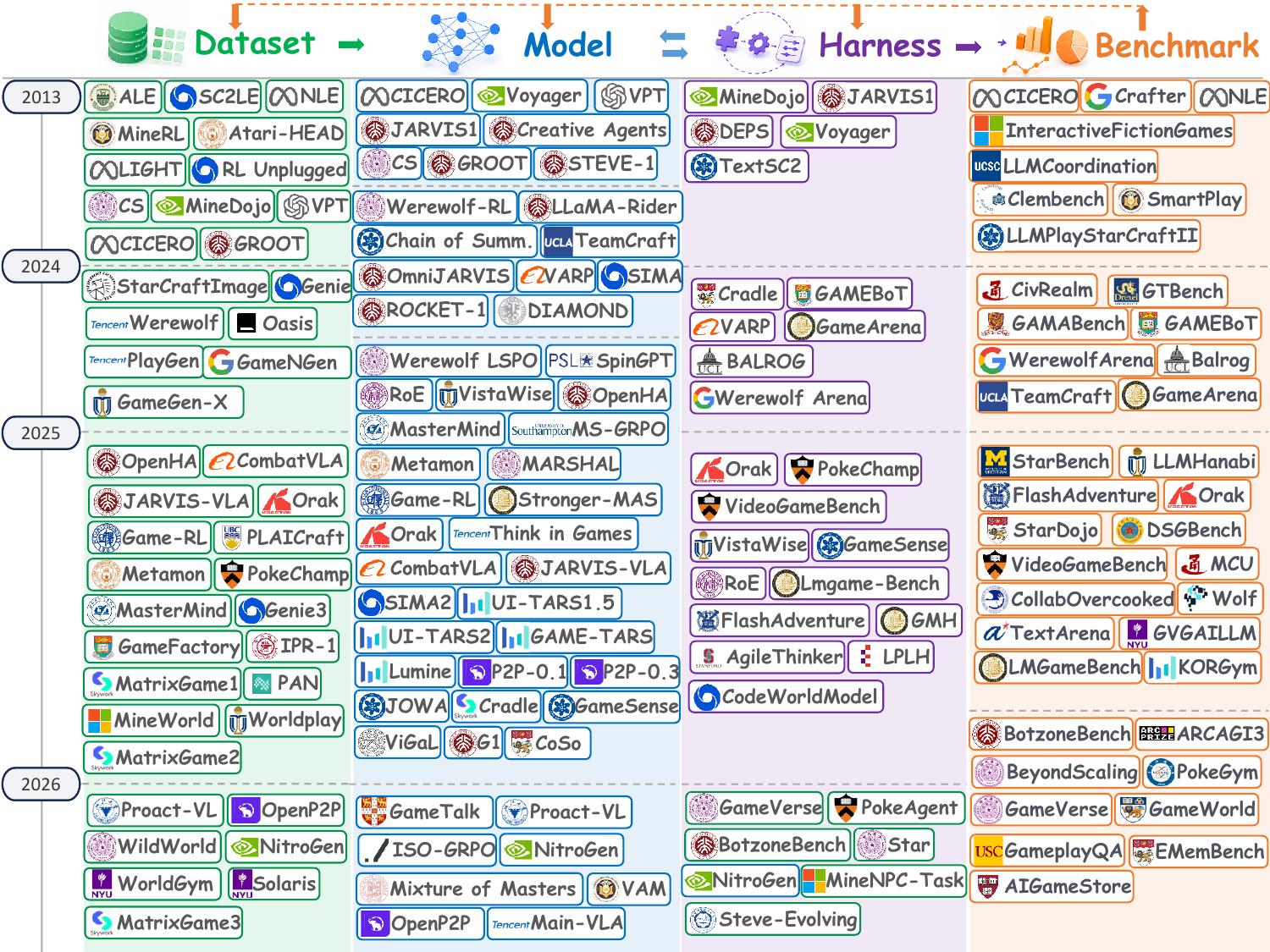}
  \caption{\textbf{A holistic overview of the research landscape toward generalist game players.} It organizes representative works along a temporal axis and unifies them into four key pillars: \textbf{Dataset, Model, Harness, and Benchmark}. By illustrating the evolution and interplay of these components over time, it reveals the field's transition from isolated, single-game studies to a unified paradigm of cross-game, multimodal generalist agents.}
  \label{fig:teaser_chapter1}
\end{figure}

\section{Introduction: From a Single Worldline to the Multiverse of Games}

\epigraph{\textit{``Man only plays when in the full meaning of the word he is a man, and he is only completely a man when he plays.''}}{--- \textbf{Friedrich Schiller}}

Games are far from mere pastimes; they are profound abstractions of reality---originating from life, yet transcending it. Collectively, they constitute a vast \textbf{\textit{multiverse}}, where each universe is governed by entirely distinct rules, aesthetics, physics and objectives, different from the single reality.

From a cognitive science perspective, human intelligence was developed to navigate the survival constraints of a single physical reality---a worldline characterized by consistent gravity, linear time, shared environment and fixed material properties \citep{cog72,cog73}. However, armed with the singular set of real-world experiences, humans can seamlessly adapt to the infinite \textit{\textbf{multiverse of games}} \citep{gameprior74,cognitivescience58}, from managing block-based ecosystems in Minecraft \citep{minedojo52} to orchestrating grand strategies in StarCraft \citep{ma2024large} and Civilization~\citep{civilbench61}. Within minutes, a human can adapt to a digital universe where the rules are different, the aesthetics surreal, the physics inverted, and the objectives entirely abstracted from physical survival. The \textit{\textbf{omni-reality adaptability}} to utilize localized experience to rapidly master a multiverse of entirely novel environments is the ultimate symbol of generalized intelligence \citep{cognitivescience58,univeralintelligence59,turing54}.

Driven by the pursuit of this generalized intelligence, researchers naturally turned to these digital games as the ultimate testbed for AI since its inception \citep{Deepblue64,turing54,gameforagi62}. \rev{This coupling between games and the pursuit of general intelligence is a premise the community has shared for decades: from Turing's early treatment of chess as a probe of machine intelligence \citep{turing54}, through Deep Blue \citep{Deepblue64}, AlphaGo \citep{AlphaGo65}, AlphaStar \citep{AlphaStar66}, and OpenAI Five \citep{DotaFive67}, to today's foundation-model agents such as Voyager \citep{voyager8}, Game-TARS \citep{gametars}, and Lumine \citep{lumine}.} Digital worlds offer high quality, dynamic, visually rich, and long-horizon environments that demand a synthesis of perception, planning, reasoning, and motor control, while AI pushes the boundaries of what can be automated and solved within these worlds. For decades, the AI community has utilized games as a crucible to train and measure capability~\citep{ale,Malmo68,minedojo52,gym22,flashadventure6,lmgbench9,oark5,gameverse69,aigamescore57,Deepblue64,AlphaGo65,AlphaStar66,DotaFive67,gametars}, yet it has historically failed to replicate this omni-reality adaptability. Generally, past paradigms have largely produced \textbf{\textit{specialists}}. Despite achieving superhuman performance through environment-specific Symbolic and Reinforcement Learning (RL) in certain complex games like DeepBlue in Chess \citep{Deepblue64}, AlphaGo in Go \citep{AlphaGo65}, AlphaStar in StarCraft II \citep{AlphaStar66} and OpenAI Five in Dota 2 \citep{DotaFive67}, these systems were developed and evaluated for specific games. Their training can combine human demonstrations and reinforcement learning, as in AlphaGo and AlphaStar, rather than always starting without prior data. Their strong game-specific results do not by themselves establish transfer to a different game. Moreover, these models almost operate as black boxes, heavily obscuring the underlying mechanisms that drive their decision-making processes \citep{interpre70,interpre71}. Ultimately, systems confined to this paradigm fall fundamentally short of the true AGI that AI community pursues.

Recently, the advent of Large Foundation Models (LFMs), including Large Language Models (LLMs) \citep{GPT-3-75,GPT-3.5-76,qwen1-78,yang2025qwen3technicalreport}, Vision-Language Models (VLMs) \citep{gpt4o25,gemini2.526,qwen3vl24,sima2-87}, Vision-Language-Action Models (VLAs) \citep{RT-2-79,combatvla,jarvis_vla,gametars}, and World Models (WMs) \citep{worldmodel82,oasis83,matrix1-84,matrix2-85,genie3-86}, has sparked a transformative paradigm shift. Rather than mastering a single game through billions of trial-and-error episodes from scratch, LFMs were born with vast open-world knowledge and emergent reasoning capabilities. By treating games not as isolated optimization problems, but as diverse instances of interactive environments, these models show the potential of \textbf{\textit{omni-reality adaptability}} as a true \textbf{\textit{generalist}} game player.

\begin{table}[t]
\centering
\caption{Comparison with representative surveys on foundation models and games. $\checkmark$: covered; $\sim$: partial; $\times$: not covered.}
\label{tab:survey_comparison}
{\color{revisionblue}
\small
\resizebox{\textwidth}{!}{%
\begin{tabular}{@{}llcccccc@{}}
\toprule
\textbf{Survey} & \textbf{Model families} & \textbf{Dataset} & \textbf{Model} & \textbf{Harness} & \textbf{Benchmark} & \textbf{Trade-offs} & \textbf{Roadmap} \\
\midrule
\citet{gallotta2024survey} & LLM & $\times$ & $\checkmark$ & $\sim$ & $\sim$ & $\times$ & $\checkmark$ \\
\citet{hu2024survey} & LLM & $\times$ & $\checkmark$ & $\checkmark$ & $\sim$ & $\times$ & $\sim$ \\
\textbf{Ours} & LLM / VLM / VLA / WM & $\checkmark$ & $\checkmark$ & $\checkmark$ & $\checkmark$ & $\checkmark$ (5) & $\checkmark$ (5-step) \\
\bottomrule
\end{tabular}%
}
}
\end{table}

This survey connects the literature through the lifecycle of a deployed game-playing agent. Relative to surveys of LLMs and games broadly \citep{gallotta2024survey} or LLM-based game agents \citep{hu2024survey}, we jointly examine LLMs, VLMs, VLAs, and world models, with particular attention to the harness that turns model outputs into gameplay (Table~\ref{tab:survey_comparison}). Our contributions are threefold:

\begin{itemize}
\vspace{-1.5mm}
    \item \textbf{Evolution and operational generality.} The standard goal-conditioned POMDP identifies which goals, observations, action semantics, and dynamics change across four eras of game AI. We distinguish competence depth, coverage across games, and transfer to unseen games, and apply the resulting coverage criterion to published model scores rather than treating the formalism as a standalone theoretical contribution.
    \vspace{-1mm}
    \item \textbf{A four-pillar analysis of the deployed agent.} Datasets, models, harnesses, and benchmarks determine different parts of the generality claim: training support, transferable representations and control, the effective interaction policy, and the evidence available to evaluate it. We use these dependencies to analyze failure mechanisms and identify controlled comparisons that separate model improvements from changes in data, interface privilege, or deployment budget.
    \vspace{-1mm}
    \item \textbf{A research roadmap with testable milestones.} We synthesize five research steps and five recurring trade-offs in one table, connecting current limitations to priorities across the four pillars and to evaluations that would demonstrate progress. This makes explicit why expanding game coverage, attaining expert performance, adapting online, and generating worlds require different evidence.
    \vspace{-1.5mm}
\end{itemize}

{\color{revisionblue}
\subsection{Survey Scope and Methodology}
\label{sec:survey-methodology}

\textbf{Sources.} Our primary window runs from January 2024 to April 2026, a period of rapid growth in foundation-model approaches to game playing. Across that window we tracked the proceedings of the venues in which this work appears: machine learning (NeurIPS, ICML, ICLR), computer vision (CVPR, ICCV, ECCV), natural language processing (ACL, EMNLP, NAACL), and games (IEEE Conference on Games, AIIDE, IEEE Transactions on Games). We also tracked new arXiv postings under cs.AI, cs.LG, cs.CL, and cs.CV, since in this area the frontier routinely appears as a preprint or technical report months before peer review, or instead of it. Targeted updates through 21 October 2026 supplement this primary window with 32 selected works across the four pillars in Tables~\ref{tab:paper_dataset}, \ref{tab:paper_model}, \ref{tab:paper_harness}, and~\ref{tab:benchmark_taxonomy}, together with GameVerse, GameWorld, OmniGameArena, and GameHorizon evidence for the performance comparison and roadmap. Versioned papers and separately identified official leaderboard snapshots are used; this update is not an exhaustive survey of all work released after April 2026. Work published before 2024 enters where the argument requires it rather than as part of the sweep: the era history of Section~\ref{sec:paradigm-evolution}, and the formal antecedents of the definitions in Section~\ref{sec:generalist-definition}.

\textbf{Procedure and inclusion criteria.} We seeded the corpus from two earlier surveys that focus specifically on large language models and games \citep{gallotta2024survey,hu2024survey}, then expanded it by expert curation and citation snowballing: for each retained work we followed both its reference list and the works citing it, iterating until new candidates stopped introducing categories that the four-pillar organization did not already contain. Every retained work was then checked against its original source, with venue or source type, year, system naming, and reported figures taken from the paper, technical report, or official release. A work is retained if it treats foundation models as game-playing agents, or contributes to one of the four pillars that support them, namely a dataset, a model, a harness component, or a benchmark, and if it additionally either appeared at one of the venues above, is widely recognized within the community, or is the first reported instance of a capability. The selection is expert-curated rather than produced by a fixed query string as in a PRISMA-style systematic review; its coverage is broad and representative rather than exhaustive, and limited to work published in English. The primary corpus contains 229 cited works, of which 153 appeared in 2024 or later and 119 in 2025 or later; roughly half are peer-reviewed venue papers, and the remainder are preprints, technical reports, or release notes. Together, the targeted updates add 34 references beyond this primary corpus, including two separately cited official result snapshots, for 263 cited works in total. The four-pillar additions are drawn from the survey's public reading list and checked against primary sources; one work may contribute to several tables.

\textbf{Scope of evidence.} The share of preprints, technical reports, and release notes reflects how this area publishes: the frontier is typically released in that form months ahead of peer review, and sometimes in place of it. We therefore prioritize widely recognized, high-impact works, report each figure under the conditions that produced it (model, interface, metric, opponent), and read such results as indicative of capabilities and failure modes rather than as settled findings; Section~\ref{sec:conclusion} returns to this limitation.
}

\section{Preliminary: AI, Games, and the Paradigm Shift}

\begin{figure*}[!t]
  \centering
  \includegraphics[width=1\linewidth]{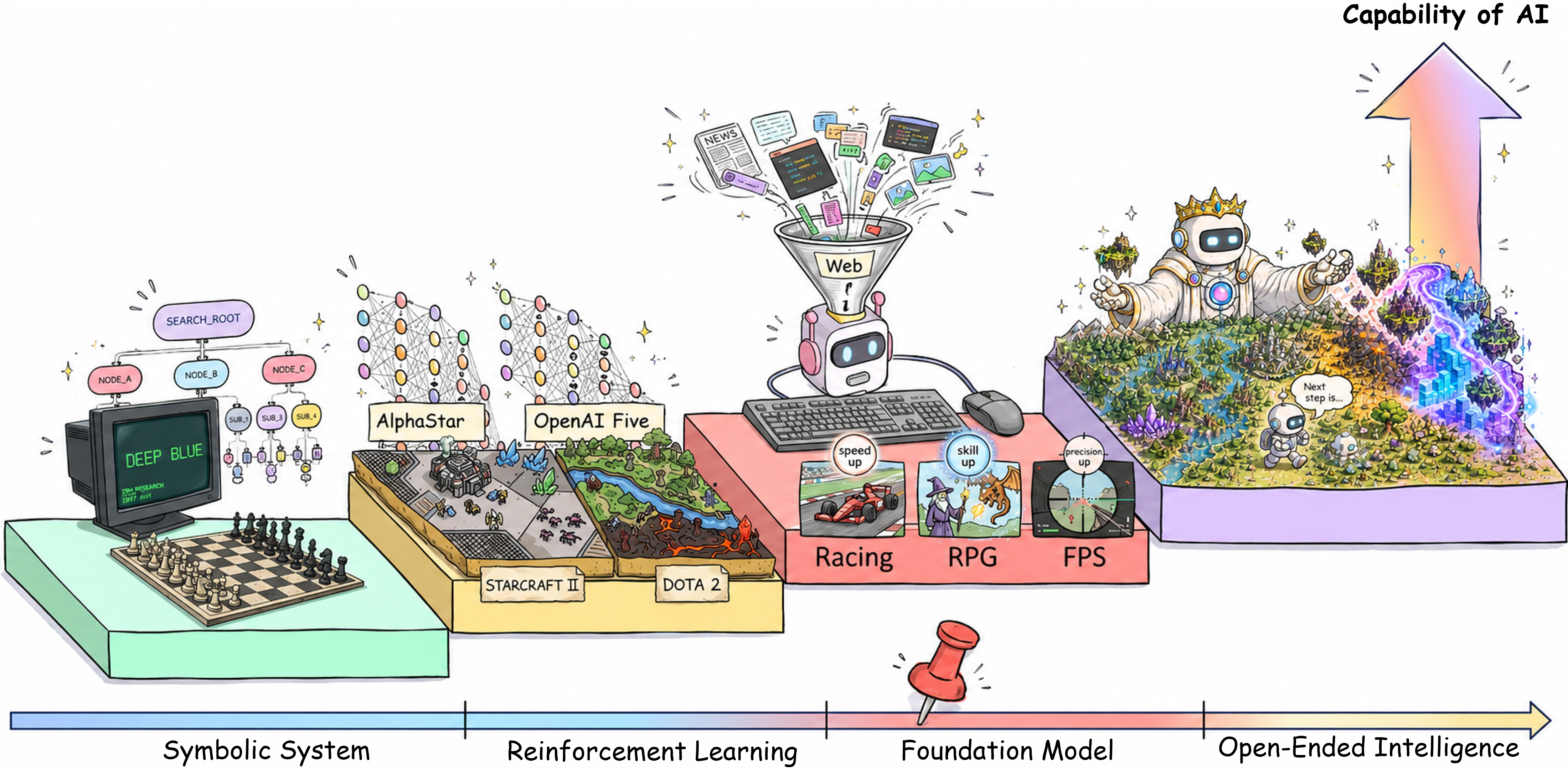}
  \caption{\textbf{The evolution of game-playing AI paradigms towards AGI.} We categorize the timeline into four eras: (1) Symbolic Systems, relying on hard-coded heuristics within isolated environments; (2) Deep Reinforcement Learning (DRL), achieving superhuman mastery in specific domains, yet constrained as narrow experts; (3) Large Foundation Models (LFMs), emerging as generalist agents capable of reasoning and adaptation across the human-crafted multiverse of games; and ultimately, (4) The Creator, the future where AI transcends playing to become the simulator itself, autonomously generating and evolving infinite game multiverses.}
  \label{fig:adaptive_scale}
\end{figure*}

\subsection{The Symbiosis and Formalization of AI in Games}

The quest for AI has always been closely linked to games. From the early days of chess programs \citep{Deepblue64} to modern highly complex strategy games \citep{ma2024large}, games serve as a standardized, quantifiable proxy for real-world decision-making. They include diverse challenges, such as partial observability, long-horizon planning, and real-time multimodal processing, while without the physical risks and costs of real-world robotics.

To establish a rigorous foundation, we formalize the interaction between an AI agent and a game environment as a Partially Observable Markov Decision Process (POMDP) \citep{markov-88}. A game can be described as a tuple $\mathcal{M} = \langle G, S, A, T, R, \Omega, O, \gamma \rangle$, where: 

\begin{itemize}
    \vspace{-1.5mm}
    \item $G$ is a set of potential goals or objectives. Each $g \in G$ is a task objective (e.g., natural language prompts, target images, or specific sub-tasks) that dictates the agent's current mission.
    \vspace{-1mm}
    \item $S$ is a set of states. Each $s \in S$ represents the internal state of the environment.
    \vspace{-1mm}
    \item $A$ is a set of actions. Each $a \in A$ is an executable input to the game. Internal reasoning, retrieval, and memory updates belong to the agent harness; a tool call is an environment action only when it is part of the declared game interface.
    \vspace{-1mm}
    \item $T: S \times A \rightarrow \Delta(S)$ is the state transition probability function which takes a state-action pair $(s, a)$ and outputs the probability distribution $T(s' \mid s, a)$ of the next state.
    \vspace{-1mm}
    \item $R: S \times A \times G \rightarrow \mathbb{R}$ is the feedback/reward function, conditioned on the specific goal $g \in G$. The reward $r = R(s, a, g)$ is scalar; textual or visual feedback is represented through observations.
    \vspace{-1mm}
    \item $\Omega$ is a set of observations accessible to the agent.
    \vspace{-1mm}
    \item $O: S \times A \rightarrow \Delta(\Omega)$ is the observation probability function which takes a state-action pair $(s, a)$ and outputs the distribution $O(o_{t+1} \mid s_{t+1},a_t)$ of the next observation, conditioned on the reached state $s_{t+1}$ and preceding action $a_t$.
    \vspace{-1mm}
    \item $\gamma \in [0,1]$ is the discount factor; $\gamma=1$ is permitted for finite-horizon evaluation.
    \vspace{-1.5mm}
\end{itemize}

Under this framework, by explicitly introducing $G$, the agent aims to find a goal-conditioned policy $\pi(a_t \mid h_t, g)$, where $h_t=(o_0,a_0,\ldots,a_{t-1},o_t)$ is the interaction history, that maximizes the expected discounted return $E[\sum_{t=0}^{\infty} \gamma^t R(s_t, a_t, g)]$. 

\subsection{Operationalizing the Generalist Game Player}
\label{sec:generalist-definition}

We use the standard goal-conditioned POMDP as an organizing and measurement device, not as a new model of intelligence. Following performance-based accounts of generality \citep{univeralintelligence59,xland2021,ada2023}, the question is whether one deployed agent maintains competence as goals, observations, action semantics, and dynamics change. Contextual MDPs and goal-conditioned values provide the formal antecedents \citep{hallak2015contextual,schaul2015universal}. A game distribution $\mathcal D$ specifies which worlds matter and their weights; genre labels alone do not establish structural diversity.

\textbf{Measure the deployed agent under a declared protocol.} Write $\Pi_{\theta,H}$ for the effective policy formed by model parameters $\theta$ and harness $H$, including prompting, memory, tools, and action translation. Its evaluation protocol $\mathcal P$ specifies goals, initial states, termination, seeds, co-players, memory reset or carryover rules, observation and action access, and interaction and wall-clock budgets, including whether the game pauses during inference \citep{machado2018revisiting}. For a goal $g$, its expected return is
\begin{equation}
V_{\mathcal M}^{\mathcal P}(\Pi,g)
=\mathbb E_{\Pi,\mathcal M,\mathcal P}\!\left[\sum_{t=0}^{\infty}\gamma^t R(s_t,a_t,g)\right].
\label{eq:game-value}
\end{equation}
Raw returns have incompatible units across games. With higher scores oriented as better, a lower anchor $b_{\mathcal M,g}$ and human or expert reference $h_{\mathcal M,g}>b_{\mathcal M,g}$ measured under the same protocol give
\begin{equation}
\mathcal C_{\mathcal M}^{\mathcal P}(\Pi)
=\mathbb E_{g\sim\mathcal D_G^{\mathcal P}}\!\left[
\frac{V_{\mathcal M}^{\mathcal P}(\Pi,g)-b_{\mathcal M,g}}
{h_{\mathcal M,g}-b_{\mathcal M,g}}\right].
\label{eq:normalized-competence}
\end{equation}
This is the familiar affine normalization used in Atari evaluation \citep{ale,DQN63,badia2020agent57}. The lower anchor may be random play or a published minimum, but the choice must be stated: zero means the chosen anchor, not necessarily random competence. The reference also matters; human and world-record normalization answer different questions \citep{dreamerv2_wm}. Scores remain unclipped, permitting values below zero or above one. Degenerate anchors require raw-score reporting rather than division. Clipped scores, geometric aggregates, and opponent-relative ratings cannot automatically be substituted into Eq.~\ref{eq:normalized-competence}.

\textbf{Separate breadth from depth.} A mean score describes average competence, but can conceal failures compensated by a few strong games. We therefore also report the fraction of games exceeding an explicit competence threshold.
\begin{definition}[$(\epsilon,\delta)$-Generalist Game Player]
\label{def:generalist}
For $\epsilon,\delta\in[0,1]$, a deployed policy $\Pi_{\theta,H}$ is an $(\epsilon,\delta)$-generalist on $(\mathfrak M,\mathcal D)$ under protocol $\mathcal P$ if
\begin{equation}
\Pr_{\mathcal M\sim\mathcal D}\!\left[
\mathcal C_{\mathcal M}^{\mathcal P}(\Pi_{\theta,H})\geq 1-\epsilon
\right]\geq 1-\delta,
\label{eq:generalist-criterion}
\end{equation}
with model parameters and harness code and prior resources fixed across evaluation games, without per-game engineering or parameter updates.
\end{definition}
The protocol may supply game manuals as inputs and permit episode memory or in-context adaptation; these differ from manually replacing the controller for each game. Fixed-weight agents such as Gato and AdA illustrate why parameter sharing and adaptation must be distinguished \citep{reed2022generalist,ada2023}. Here $1-\epsilon$ sets competence depth and $1-\delta$ its coverage. Unlike the confidence parameter in PAC learning \citep{valiant1984theory}, $\delta$ is failure mass over games. A uniform distribution on a finite suite is a valid $\mathcal D$; estimated returns then yield descriptive coverage on that suite, not a guarantee on unseen games. Results should report several thresholds alongside mean and per-game scores, because both thresholds and game weights affect the comparison. Section~\ref{sec:generality-comparison} applies this criterion to published results.

\textbf{Distinguish coverage from transfer.} Evaluating many games does not by itself establish that they were unseen in training. Within-family transfer varies levels or tasks within a shared environment family; structural transfer tests changes in action semantics, observation functions or modalities, or transition rules, rather than only new textures or titles. Granting extra privileged state is a protocol change and must be reported separately. With declared training and test distributions and comparable protocols, the conventional train--test difference is
\begin{equation}
\operatorname{Gap}(\Pi)=
\mathbb E_{\mathcal M\sim\mathcal D_{\mathrm{tr}}}[\mathcal C_{\mathcal M}^{\mathcal P}(\Pi)]
-\mathbb E_{\mathcal M\sim\mathcal D_{\mathrm{test}}}[\mathcal C_{\mathcal M}^{\mathcal P}(\Pi)].
\label{eq:generalization-gap}
\end{equation}
This follows established generalization evaluation \citep{cobbe2020leveraging,kirk2023survey}. The difference is descriptive: unequal game difficulty, goal mixtures, and reference anchors can affect it, so it does not isolate a cause of transfer loss. A small gap is useful only together with high test competence: uniformly weak play also yields a small gap. Undisclosed pretraining data may prevent a clean transfer claim even when test scores are measurable. Weight-updating adaptation requires a separate protocol reporting interaction cost, performance before and after adaptation, and retention.

These distinctions guide the four pillars. Data determines which goal--observation--action combinations the agent can learn; models determine whether representations and control transfer; the harness changes the effective policy and sometimes the information available to it; benchmarks determine which breadth, depth, and transfer claims the resulting evidence supports. The following sections use these dependencies to identify bottlenecks and the comparisons needed to distinguish them.

\subsection{The Evolution of Game-Playing AI Paradigms}
\label{sec:paradigm-evolution}

The evolution of game-playing AI can be mapped to how agents interact with, and control the elements of, the POMDP tuple $\mathcal{M}$. This progression marks a fundamental shift from rigid rule execution in constrained state spaces to open-ended multiverse simulation. Table \ref{tab:paradigm_shift} visualizes this paradigm shift across four distinct eras:

\begin{table*}[t]
\centering
\caption{The Paradigm Shift in Game-Playing AI Formalization. We frame the evolution through the lens of a Goal-Conditioned POMDP $\mathcal{M} = \langle G,S, A,  T, R, \Omega, O, \gamma \rangle$, highlighting the transition from rigid, pre-defined components to agent-generated open-ended dynamics.}
\label{tab:paradigm_shift}
\resizebox{\textwidth}{!}{
\begin{tabular}{@{}l l l l l@{}}
\toprule
\textbf{Feature} & \textbf{Era 1: Symbolic} & \textbf{Era 2: Deep RL} & \textbf{Era 3: Foundation Models} & \textbf{Era 4: The Creator, Demiurge} \\
\midrule
\textbf{Timeline} & Pre-2010 & 2010-2022 & 2022-Present & Post-Future\\
\addlinespace
\textbf{Scope} & Single Universe & Single Universe & Human-crafted Multiverse & Theoretical Multiverse\\
\addlinespace
\textbf{Mechanism} & Rules $\rightarrow$ Search & State/Pixels $\rightarrow$ Policy & Knowledge $\rightarrow$ Reason $\rightarrow$ Action & Simulate $\rightarrow$ Generate $\rightarrow$ Evolve \\
\addlinespace
\textbf{Goal ($G$)} & Fixed Condition & Maximize Reward & Dynamic Tasks & Self-assigned \& Evolving \\
\addlinespace
\textbf{Action ($A$)} & Symbolic API & Structured API & Human-like Interface & \textbf{Unconstrained / Generative} \\
\addlinespace
\textbf{Dynamics ($T$)} & Deterministic & Fixed Black-box & Fixed Black-box & \textbf{Dynamically Generated} \\
\addlinespace
\textbf{Reward ($R$)} & Terminal / Heuristic & Env-defined Scalar & Env Scalar + Semantic Evaluation & \textbf{Agent-generated} \\
\addlinespace
\textbf{Obs. ($O, \Omega$)} & Bypassed (Direct $S$) & Pixels / Structured ($\Omega_{px/ vec}$) & Multimodal ($\Omega_{text, vision, audio}$) & \textbf{Omniscient / Bird's-eye} \\
\addlinespace
\textbf{World Prior} & Hand-craft Logic & Game-specific Experience & Pre-trained Knowledge  & Self-defined Logic \\
\addlinespace
\textbf{Target} & Specialist & Specialist & \textbf{Generalist Player} & \textbf{Omnipotent Creator} \\
 
\bottomrule
\end{tabular}
}
\end{table*}

\rev{We assign a system to an era by which components of $\mathcal{M}$ it treats as fixed versus agent-controlled (Table~\ref{tab:paradigm_shift}), not by publication date; the eras are analytical stages rather than disjoint periods, and a system may combine features of adjacent eras.}

\textbf{Era 1: Symbolic Systems - \textit{Know the Rules, Search the Trees}.} Prior to 2000, early AI relied on the hard-coded rules and heuristic search algorithms \citep{minmax53,Deepblue64}. In this era, the observation function $O$ was often bypassed, giving the agent direct access to a fully observable and simplified state space $S$ (e.g., the exact positions of the chess pieces). The action space $A$ was rigidly defined by game-specific logical rules, and the goal $G$ was a singular, fixed terminal state, such as checkmate in chess. These systems required domain experts to hand-craft features and lacked the ability of true perception and generalization.

\textbf{Era 2: Deep Reinforcement Learning - \textit{Learn Policies from Game Experience}.} Since the 2010s, the integration of deep neural networks with RL shifted the paradigm toward learned perception and decision-making \citep{AlphaGo65,AlphaStar66,DotaFive67}. Agents learned policies $\pi(a_t | o_{\le t})$ from high-dimensional observation spaces $\Omega$ (such as raw pixels or structured input). However, the action space $A$ remained a fixed, structured API as pre-defined control vectors in the specific task, and the goal $G$ was implicitly bound to a static, environment-provided scalar reward $R$. These systems learned strong policies through game-specific training, sometimes combining human demonstrations with self-play. Their reported success establishes competence in their target games, rather than a shared policy with demonstrated cross-game transfer.

\textbf{Era 3: Foundation Models - \textit{Understand, Reason, Then Act}.} Now, the current era is defined by models pre-trained on internet-scale multimodal data \citep{gametars,lumine,uitars119,openp2p89,nitrogen}. Here, the observation space $\Omega$ expands to include rich multimodal inputs (vision, audio, text). Language-conditioned systems broaden the goal space $G$ to include natural-language tasks, while vision-action systems such as NitroGen can retain implicit goals without language conditioning. Furthermore, the action space $A$ shifts from game-specific APIs toward shared, human-facing interfaces such as keyboard, mouse, and gamepad inputs; their action semantics still differ across games. By leveraging pre-trained world knowledge to infer underlying transition dynamics $T$, these models transition the field toward generalist game players in the human-crafted multiverse of games.

\textbf{Era 4: The Creator, Demiurge - \textit{Simulate, Generate, Then Evolve}.} \rev{Era 4 (the Demiurge) is the forward-looking endpoint of the $\mathcal{M}$-control trajectory in Table~\ref{tab:paradigm_shift}; world models \citep{genie3-86,oasis83} and embodied agents are early, active steps in this direction.} In the future, the AI transcends the role of a participant constrained by a fixed game $\mathcal{M}$ \citep{sima2-87,genie3-86}. Instead of merely optimizing a policy within predefined boundaries, the agent becomes the simulator itself. It possesses the capability to actively generate and expand the state space $S$ and action space $A$ dynamically. It invents its own evolutionary goals $G$, and constructs the transition dynamics $T$ and reward structures $R$. The interface becomes completely free-form, shifting from existing games to simulating, evolving, and directing the progression of entire theoretical multiverses.

\subsection{The Current Frontier: The Nexus of Era 2 and Era 3}
Situating the current landscape within the evolutionary trajectory of game-playing AI, we find ourselves at a critical stage: transitioning from the specialized optimization of Era 2 to the generalist paradigms of Era 3. While Deep RL in Era 2 has yielded specialists capable of mastering complex POMDPs, these systems often remain confined by the generalization bottleneck and the cost of training and engineering $\pi$ for each game $\mathcal{M}$ is becoming increasingly prohibitive.

Consequently, the field is pivoting toward Era 3: Foundation Model as generalists. By leveraging the world knowledge priors of Large Foundation Models, emerging frameworks demonstrate remarkable multimodal integration, zero-shot adaptation and goal-conditioned reasoning. However, significant challenges remain across the board, spanning accurate perception, sparse reasoning, memory retrieval, precise control, and efficiency, affecting virtually every layer of the system.

This survey specifically focuses on this paradigm-shifting frontier. We comprehensively review the ongoing transition, exploring the datasets, models, harness, benchmarks, and persistent challenges that define this pivotal moment in AI evolution, \rev{and outlining the steps that lead from it toward Era 4.}

\section{Datasets: The Foundation Beneath Generalist Game Player}

\begin{figure*}[htbp]
  \centering
  \includegraphics[width=1\linewidth]{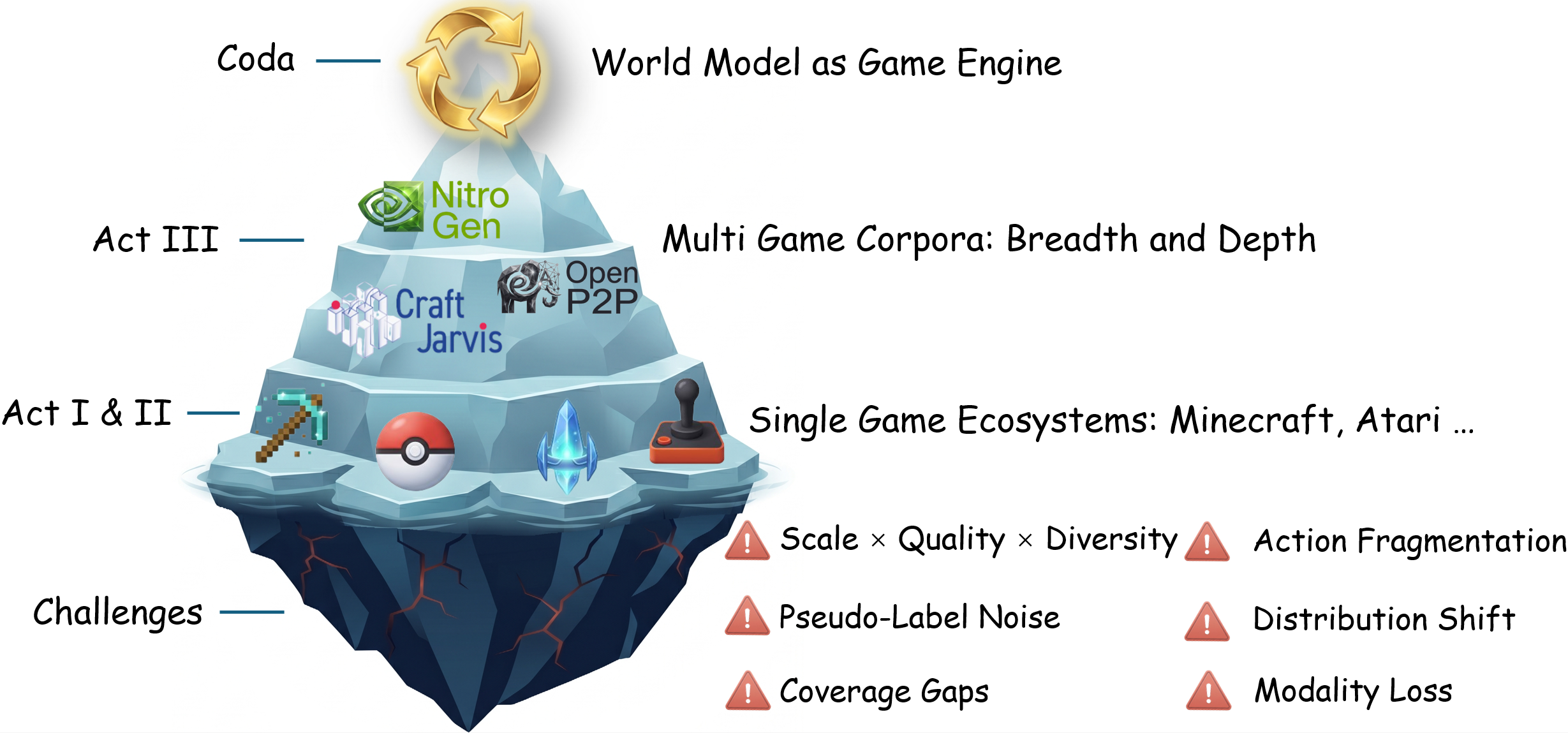}
  \caption{\textbf{Overview of Dataset in Game-playing AI.} The trajectory of data curation is shifting from isolated, single-game ecosystems (Acts I \& II) to comprehensive, multi-game corpora (Act III) in both depth and breadth, tailored for generalist agents. We highlight the future where world model can replace the game engine, and we show the critical bottlenecks in current dataset construction.}
  \label{fig:dataset_teaser}
\end{figure*}

For generality, the dataset question is whether training covers the decisions required in unfamiliar games, or mainly supplies more examples of familiar play. Among the four fundamental pillars of our pipeline, datasets occupy a uniquely structural position. They serve not merely as the starting point of the development loop but rather as the constraint surface that shapes all downstream processes. The model architectures a community can explore, the harness designs that prove viable, and the benchmark dimensions that remain measurable are mostly predetermined, frequently in an implicit manner, by the data available at the onset of training.

In the DeepRL Era 2 paradigm, agents \citep{AlphaGo65,AlphaStar66,DotaFive67} generated their own training data through direct environment interaction: each self-play loop was self-contained, the data it produced was inseparable from the game that generated it, and nothing transferred elsewhere. Era 3 breaks this closure. Foundation models require corpora that are collected before training begins, that span multiple games, and that carry value beyond the domain they came from. The question is no longer "\textit{how many environment steps can the agent take?}" but "\textit{what kind of data should we prepare, and where does it come from?}"

Table~\ref{tab:paper_dataset} summarizes training corpora, evaluation archives, and data-scaling studies. In the summary tables, NR denotes information not established by the source, and N/A means not applicable.

This section answers that question through three chapters and a coda. 

\begin{itemize}
    \item \textbf{Act I --- Requisition: Learning from Human Play.} Act I examines how the field learned to extract training signals from human gameplay, progressing from initial structured demonstration datasets to internet-scale pseudo-labeled video and finally to multimodal time-aligned streams.
    \item \textbf{Act II --- Excavation: Mining the Competitive Archive.} Act II investigates a complementary data source encompassing decades of replays archived on gaming platforms, alongside algorithmic synthesis methods that distill strategic knowledge directly from game mechanics.
    \item \textbf{Act III --- Scaling: Crossing the Single-Game Boundary.} Act III faces the fundamental challenge of transcending the single-game boundary, documenting the dual pathways of broad automated collection and deep structured curation that the community has pursued to construct multi-game corpora.
    \item \textbf{Coda --- Creation: The World Model as Data Engine}. Finally, the Coda looks beyond these three phases to the emerging paradigm of the \textit{world-model-as-data-engine}, effectively connecting the datasets pillar to the Era 4 Demiurge vision that frames our comprehensive survey.
\end{itemize}

\begin{table*}[!t]
\centering
\caption{Summary of representative large datasets in gaming AI}
\label{tab:paper_dataset}
\begingroup
\setlength{\tabcolsep}{4pt}
\renewcommand{\arraystretch}{1.08}
\resizebox{\textwidth}{!}{%
\begin{tabular}{llllll}
\toprule
\textbf{Paper} & \textbf{Modality} & \textbf{Data Source} & \textbf{Label Method} & \textbf{Scale} & \textbf{Game Set} \\
\midrule
ALE~\citeyearpar{ale} & Frame + action & Atari 2600 ROMs & Env-provided reward & 57 games & 57 Atari games \\
SC2LE~\citeyearpar{sc2le} & State + actions & Ladder replays & Auto-recorded & \rev{$\sim$800K games (SL set)} & StarCraft II \\
LIGHT~\citeyearpar{light} & Text dialogues & Crowdsourced roleplay & Human-authored & 663 scenes, 11K episodes & Text adventure \\
MineRL~\citeyearpar{minerl} & RGB + actions & Crowdsourced play & Auto-recorded & 60M frame-action pairs & Minecraft \\
Atari-HEAD~\citeyearpar{atari-head} & Frame + gaze + action & Human play + eye-tracker & Sync-recorded & 117 hrs, 328M gaze & 20 Atari games \\
NLE~\citeyearpar{nle} & Symbol + action & Procedural generation & Env-provided reward & Infinite (procedural) & NetHack \\
RL Unplugged~\citeyearpar{rl-unplugged} & Frame + action & RL replay buffers & Auto-recorded & 46 games $\times$ 200M frames & 46 Atari games \\
CS Deathmatch~\citeyearpar{csgo} & RGB + kbd/mouse & Server spectation & Rule-based IDM & \rev{$\sim$5.8M frames, 100 hrs} & CS:GO \\
VPT~\citeyearpar{vpt} & Video + actions & Web videos & IDM pseudo-label & 70K hrs (clean) & Minecraft \\
MineDojo~\citeyearpar{minedojo52} & Video + text & YouTube/Wiki/Reddit & None (natural) & 730K videos, 300K hrs & Minecraft \\
CICERO~\citeyearpar{cicero} & Text + board state & webDiplomacy.net & Auto-aligned & 40K+ human games & Diplomacy \\
StarCraftImage~\citeyearpar{starcraftimage} & Rendered images & Match replays & Auto-generated & 3.6M imgs / 60K replays & StarCraft II \\
GROOT~\citeyearpar{groot51} & Video + actions & Contractor + YouTube & IDM pseudo-label & 30 tasks (benchmark) & Minecraft \\
Werewolf-18K~\citeyearpar{wu2024enhancereasoninglargelanguage} & Text dialogues & Human play & Auto-recorded & 18.8K sessions & Werewolf \\
Genie~\citeyearpar{genie} & Unlabeled video & Web videos & None (unsupervised) & 30K hrs (filtered) & 100+ platformers \\
GameGen-X~\citeyearpar{gamegenx} & Video + captions & Web game videos & GPT-4o captioning & 1M+ clips, 150+ games & 150+ games \\
GameNGen~\citeyearpar{gamengen_wm} & Frame + action & RL-agent gameplay & Auto-recorded & 20 FPS, multi-minute play & DOOM \\
PlayGen~\citeyearpar{yang2024playablegamegeneration} & Frame + action & Agent-generated gameplay & Auto-recorded & 20 FPS, 1K+ Gen. frames & Mario \& DOOM \\
OpenHA~\citeyearpar{openha} & RGB + actions & VPT trajectories & Rule-based pipeline & \rev{5B tokens, 800+ tasks} & Minecraft \\
JARVIS-VLA~\citeyearpar{jarvis_vla} & Image + actions & Scripted agent & Self-supervised tokenizer & \rev{6.4M entries ($\sim$10B tokens), 1K+ tasks} & Minecraft \\
PLAICraft~\citeyearpar{plaicraft} & 5-modal aligned & Human play & None (time-aligned) & 10K hrs, 10K players & Minecraft \\
Game-RL~\citeyearpar{gamerl} & Image + QA & Code generation & Code-verified & \rev{$\sim$140K QA pairs, 158 tasks} & 30 games \\
Metamon~\citeyearpar{metamon} & RL trajectories & Platform replays & Auto-parsed & 2M human + 18M self-play & Pok\'{e}mon \\
Pok\'{e}Champ~\citeyearpar{karten2025pokechamp} & Text battle logs & Platform replays & Reverse-engineered & 3M+ battles & Pok\'{e}mon \\
Mastermind~\citeyearpar{wang2025mastermind} & Text (encoded) & Algorithm synthesis & Algorithm-labeled & 1.7M + 288K samples & 2 Chess games \\
CombatVLA~\citeyearpar{combatvla} & Screenshot + actions & Expert human play & Human AoT labels & 25K imgs, 200 hrs & Black Myth \\
GF-Minecraft~\citeyearpar{gamefactory_wm} & Video + kbd/mouse & Programmatic capture & Auto-recorded & 70 hrs, 2K clips & Minecraft \\
Orak~\citeyearpar{oark5} & Text/Image/Both & LLM-generated & LLM + human verified & 11K fine-tuning samples & 12 games \\
Matrix-Game~\citeyearpar{matrix1-84} & Video + kbd/mouse & MineDojo + auto-collected gameplay & Auto-labeled actions & $\sim$ 400 frames & Minecraft \\
Matrix-Game 2.0~\citeyearpar{matrix2-85} & Video + kbd/mouse & UE + GTA5 auto-production & Auto annotations & 25 FPS, minute-level gen & GTA5 + UE scenes \\
WildWorld~\citeyearpar{li2026wildworld} & Video + state & ARPG auto-capture & Auto-extracted & 108M frames, 450+ actions & Monster Hunter \\
NitroGen~\citeyearpar{nitrogen} & Video + gamepad & Web videos & CV overlay extraction & 40K hrs, 1K+ games & 1,000+ games \\
Open-P2P~\citeyearpar{openp2p89} & Video + kbd/mouse & Expert human play & Recorded + VLM text & 8.3K hrs, 600M pairs & \rev{42 3D games} \\
Proact-VL~\citeyearpar{yan2026proactvlproactivevideollmrealtime} & Video + ASR text & YouTube streams & Multi-stage pipeline & 561 hrs, 128K samples & 12 games \\
Solaris~\citeyearpar{solaris_wm} & Video + actions & Automated multiplayer collection & Sync-recorded & Minute-level play & Minecraft \\
Matrix-Game 3.0~\citeyearpar{matrix3_wm} & Video + action + pose + text & UE + auto collection + augmentation & Auto-generated & 40 FPS; minute-long play & UE + AAA games \\
Agent Island~\citeyearpar{agentisland2026} & Dialogue/votes/events & Live benchmark & Recorded traces; annotation NR & Dynamic; count NR & Agent Island \\
CrossFPS~\citeyearpar{scope2026} & Video + 10-DoF controls & FPS gameplay & Frame-aligned telemetry & 69K clips & 7 FPS games \\
MGC2025~\citeyearpar{mindgames2026} & Observations/actions/rewards & Multi-agent competition & Automatic trajectory logging & 29,571 matches & 4 social/strategic games \\
AgenticSTS~\citeyearpar{agenticsts2026} & Records + partial logs & Tagged agent runs & Automatic outcome logging & 312 records (298 completed) & Slay the Spire 2 \\
StatePlay~\citeyearpar{stateplay2026} & Video + actions/state/text & Recorded gameplay & Synchronized traces + prompts & 10K train rows & Street Fighter \\
BOSS-140K~\citeyearpar{worldmind2026} & Video + actions/state & State-conditioned agent & Synchronized state/actions & 144,631 clips & 3 games \\
GameWAM~\citeyearpar{gamewam2026} & Video + keyboard/mouse & VPT + scripts/APPO & Inherited / recorded actions & NR & Minecraft + ViZDoom \\
GameHorizon~\citeyearpar{gamehorizon2026} & Video + actions/instructions & Expert human play & Segmentation + VLM annotation & 5,000 h & 21 games \\
GameDirector~\citeyearpar{gamedirector2026} & Video + actions/state & Automated gameplay & Synchronized state/actions & 51,758 clips & 3 games \\
GameScaling~\citeyearpar{zhang-scaling-games} & Frames + actions/text & VPT/D2E/OpenP2P + general data & Inherited actions + reasoning & 50--1,600 h & Mixed-game corpora \\
AAArena~\citeyearpar{aaarena2026} & Policy programs + ranks & Competition archives & Ranking/evaluator metadata & 1,920 programs & 12 adversarial games \\
Video-Arena~\citeyearpar{videoarena2026} & Video + reference tuples & Recorded matches/replays & Parsers + engine traces & 180 cases; \textasciitilde{}74 h & 9 games \\
Player-1~\citeyearpar{player1_2026} & Video + actions/text & Gameplay/UI demonstrations & Recorded + goal/reasoning labels & 6,044 trajectories & 172 games \\
\bottomrule
\end{tabular}
}
\endgroup
\end{table*}

\subsection{Act I --- Requisition: Learning from Human Play}

Human players generate enormous behavioral traces every day, and they constitute the most natural source of training data for game-playing agents. The challenge of learning from human play is a tension between two quantities that move in opposite directions: the scale of available gameplay footage and the cost of attaching action labels to it.

The earliest efforts resolved this tension by choosing precision over scale. MineRL~\citep{minerl} compiled over 60 million state-action pairs from human Minecraft demonstrations. A parallel effort collected \rev{5.5 million frames (95 hours) of scraped online} Counter-Strike gameplay through the same strategy~\citep{csgo}\rev{, released together with smaller clean expert-demonstration subsets for a total of about 5.8 million frames over 100 hours}. Atari-HEAD~\citep{atari-head} captured not only frames and actions but also human gaze data via eye-tracking across 20 Atari games. More recently, CombatVLA~\citep{combatvla} invested in collecting 200 hours of Black Myth: Wukong aligned at sub-15-millisecond precision, further annotated with Action-of-Thought labels. These datasets founded imitation learning for game agents, yet their construction cost grew linearly with data volume, leaving the resulting corpora orders of magnitude smaller than the total gameplay available on the internet.

To decouple data scale from human labeling costs, recent works increasingly employ learned models as automated annotators at both perceptual and semantic levels. At the perceptual level, Video PreTraining (VPT) \citep{vpt} utilizes an Inverse Dynamics Model (IDM) to infer low-level actions from large-scale unlabeled internet videos, a paradigm also adopted by GROOT \citep{groot51} to curate task-oriented benchmarks. Complementing this, semantic-level approaches leverage the reasoning capabilities of foundation models: Orak \citep{oark5} deploys strong LLMs to autonomously generate structured expert trajectories across various game genres, while Proact-VL \citep{yan2026proactvlproactivevideollmrealtime} and GameGen-X \citep{gamegenx} employ VLMs to autonomously label and caption the video. Despite enabling internet-scale data collection, these automated methods share a critical limitation: residual noise, from IDM inaccuracies or LLM hallucinations, inevitably propagates into the training objective, degrading downstream policies over long horizons.

While VPT demonstrated that vision and action can be recovered at scale, human gameplay is far richer than frames paired with keyboard inputs. Atari-Head \citep{atari-head} collects 117 hours of Atari-2600 games with human K\&B action and gaze. PLAICraft \citep{plaicraft} pushes the boundary by collecting 10,000 hours of Minecraft gameplay with five modalities aligned in time: video, audio, natural language (voice and text), actions, and K\&M signals. It captures not only what the player did but also what the player said and aspects of the player's cognitive state. Training a model requires observations spanning the full perceptual space the agent will operate in. Multimodal temporal alignment is therefore not optional but a prerequisite for agents that perceive and reason about the game world in a human-like manner.

At the opposite end of the cost spectrum, \rev{Open-P2P} \citep{openp2p89} invested in over 8,300 hours of meticulous human annotation \rev{(around 600 million image--action pairs) across the 42 3D titles it lists}. Its core contribution is empirical: by varying model scale and data volume, the authors showed that behavior cloning follows a predictable scaling law and that increasing both dimensions leads to emergent causal reasoning. This reframes the quality-versus-quantity debate. Internet-scale pseudo-labeled data  and expertly annotated data at moderate scale are not interchangeable. They serve different roles in the training pipeline. The former provides broad behavioral coverage at low per-sample cost, while the latter supplies the high-fidelity signal needed to unlock qualitatively new capabilities. \textbf{\textit{How to optimally combine these two data regimes remains one of the central open problems in corpus design for game-playing agents.}}

Viewed as a whole, the evolution of human-sourced game data follows a clear trajectory: from expensive synchronous annotation \citep{minerl,csgo}, through internet-scale pseudo-labeling \citep{vpt}, to multimodal temporal alignment \citep{plaicraft} and quality-centric scaling experiments \citep{openp2p89}. Each advance was driven by a specific limitation of its predecessor, and the limitations that remain collectively, most notably the noise ceiling of pseudo-labels and the prohibitive cost of high-quality annotation at scale, motivate the alternative data sources explored in the following sections.

\subsection{Act II --- Excavation: Mining the Competitive Archive}

Competitive gaming platforms have accumulated decades of match replays, battle logs, and ranked statistics, forming a vast reservoir whose value for training game agents has only recently been recognized. Unlike the human demonstrations of Section 3.1, competitive replays are byproducts of the competitive ecosystem, not produced for teaching AI. Yet they carry a distinctive advantage: match outcomes and player rankings provide implicit quality signals requiring no additional annotation.

The most direct use of this archive is to extract structured training data from historical replays. The practice traces back to SC2LE \citep{sc2le}, which opened Blizzard's anonymised 1v1 ladder as a standardized research resource and reported supervised baselines on \rev{800K replays from the initial 2017 release, an archive that has kept growing since}, establishing the template of leveraging platform recording infrastructure and competitive outcomes as implicit supervision. StarCraftImage \citep{starcraftimage} later drew on 60,000 replays from the same ladder to compile 3.6 million images with spatial reasoning labels derived from the game state, using the ladder's built-in rating system as a natural skill-level filter. Metamon \citep{metamon} mines a full decade of Pok\'{e}mon Showdown competitive replays, capturing the complete evolutionary record of the metagame---from casual play to tournament-level strategy through repeated cycles of innovation, counter-adaptation, and balance patches. This combination of quality stratification and temporal diversity enables robust policy learning without active environment interaction. The common thread is that competitive replays embed implicit reward signals---rankings and win rates---that directly inform offline RL reward shaping, bypassing both the expensive manual annotation and the noisy pseudo-labeling.

Beyond replays that preserve visual or state-level information, a parallel line of work extracts strategic knowledge from textual battle logs and dialogue records at even larger scale. Pok\'{e}champ \citep{karten2025pokechamp} compiled over three million competitive Pok\'{e}mon battle logs, one of the largest corpora of structured competitive interactions. CICERO's Diplomacy corpus \citep{cicero} contributes over 40,000 full-press Diplomacy games in which natural language negotiation messages are aligned with board-state actions, encompassing the full spectrum of strategic communication. Werewolf-18K \citep{wu2024enhancereasoninglargelanguage} provides 18,800 sessions of nine-player social deduction games with complete dialogue logs and voting records. At this scale, strategic patterns and team-composition preferences emerge from aggregate distributions rather than manual expert identification. 

Evaluation archives also contribute reusable data, with incomplete public coverage: AAArena releases 909 packages from a 1,920-program pool, and AgenticSTS provides 312 records (298 completed runs) with partial logs~\citep{aaarena2026,agenticsts2026}.

For rule-complete games with tractable state spaces, training data can be algorithmically synthesized or simulated, bypassing the need for human or competitive data collection. Prominent examples include RL Unplugged \citep{rl-unplugged}, which compiles diverse agent-generated replay buffers across Atari games; NLE \citep{nle}, which provides procedurally generated NetHack episodes with automated reward annotations; Game-RL \citep{gamerl}, which utilizes Code2Logic to programmatically generate verifiable QA pairs; and Mastermind \citep{wang2025mastermind}, which synthesizes Q-value-annotated trajectories and optimal moves for Doudizhu and Go via extensive game-tree search. By enumerating optimal solutions, synthesized data often surpasses human demonstrations in quality, offering verifiable correctness without expensive manual annotation or noisy pseudo-labeling. However, its application is fundamentally bounded by computational tractability. As environments become partially observable, continuous, or open-ended, algorithmic synthesis becomes prohibitive. Consequently, this strategy serves as a robust complement to human- and competition-sourced data, rather than a universal replacement.

Taken together, the data sources examined in this section, including competitive replays with implicit quality gradients \citep{sc2le,starcraftimage,metamon}, large-scale battle logs and dialogue archives \citep{karten2025pokechamp,cicero,wu2024enhancereasoninglargelanguage}, agent-generated replay buffers and procedural environments \citep{nle,rl-unplugged}, and algorithmic synthesis with verifiable labels \citep{gamerl,wang2025mastermind}, substantially expand data volume and diversity. Yet they introduce new challenges: distribution shift from game updates, limited applicability of algorithmic synthesis to open-ended games, and the single-game focus that characterizes most competitive archives. These limitations motivate the cross-game scaling in the following section.

\subsection{Act III --- Scaling: Crossing the Single-Game Boundary}

The datasets of Sections 3.1 and 3.2 share a common limitation: most are confined to a single game or a narrow family of related games. For Era 3's goal of building generalist agents, this single-game isolation is a fundamental bottleneck. Two complementary strategies have emerged for constructing multi-game corpora: a breadth path that maximizes game coverage through automated collection, and a depth path that maximizes annotation quality within a carefully curated set of games.

The breadth path has progressed through several generations of increasingly ambitious collection. ALE \citep{ale} first standardized 57 Atari 2600 titles under a uniform frame-action-reward interface, establishing the principle that a common data format spanning dozens of games could accelerate research. Genie \citep{genie} scaled this to the internet era, harvesting over 30,000 hours of unlabeled 2D platformer videos spanning hundreds of titles and using unsupervised learning to discover latent action spaces without action annotation. \rev{NitroGen \citep{nitrogen} represents the current breadth frontier, assembling 40,000 hours across more than 1,000 games through an automated pipeline that harvests web-scale video and recovers gamepad actions with computer vision; Open-P2P \citep{openp2p89} occupies the same generation but trades coverage for fidelity, recording 8,300 hours (600 million image--action pairs) across 42 3D titles with actions logged rather than inferred.} NitroGen's significance lies in demonstrating that heterogeneous multi-game corpora can serve as the data foundation for generative pre-training architectures, including diffusion and autoregressive models. Yet breadth comes at a cost: as game coverage grows, action heterogeneity grows correspondingly. The same input in CS:GO carries entirely different semantics from that input in Dota 2. Unifying these disparate action representations remains a pressing open challenge, directly motivating the unified action tokenization.

The depth path takes the opposite stance, focusing on fewer games where every task provides rich, verifiable training signals. Game-RL \citep{gamerl} instantiates this with a corpus spanning 30 games and 158 tasks, each paired with a Code2Logic verifiable reward function---eliminating the reward misspecification that destabilizes RL on noisy data. \rev{OpenHA \citep{openha} and JARVIS-VLA \citep{jarvis_vla} contribute hierarchical VLA trajectories with layered annotation from high-level natural-language goals down to atomic keyboard commands: OpenHA releases five action- and chain-of-action datasets totalling about 5 billion tokens over 800+ tasks, and JARVIS-VLA reports 6.4 million expert entries (roughly 10 billion tokens) over 1k+ atomic tasks.} This hierarchy provides the training signal for hierarchical policy learning at multiple temporal and semantic granularities. The value of the depth path is a quality leverage effect: verifiable signals eliminate the ambiguity that leads to reward hacking, so even moderately sized datasets can produce training stability that much larger noisy corpora cannot match.

Surveying the cross-game coverage landscape reveals a highly uneven map. The Minecraft ecosystem \citep{minerl,minedojo52,vpt,groot51,openha,jarvis_vla} is the most mature, forming a multi-year chain of increasingly sophisticated resources. Text strategic games \citep{light,wu2024enhancereasoninglargelanguage,karten2025pokechamp,metamon,cicero} have likewise received substantial coverage, whose text-native interfaces align naturally with LLM input formats. The Atari ecosystem \citep{ale,atari-head,nle,rl-unplugged} provides another mature data chain available for Era 3 reuse. By contrast, several important categories remain underrepresented. Real-time 3D games, including FPS, MOBAs, and action RPGs, demand millisecond-level response times and high-dimensional continuous observations yet lack large-scale annotated datasets comparable to Minecraft. WildWorld \citep{li2026wildworld} and CombatVLA \citep{combatvla} represent early steps toward filling this gap for the ARPG subdomain, but FPS and MOBA remain data-scarce. Continuous action-space games pose a related challenge: the majority of existing datasets encode discrete keyboard inputs, leaving the continuous control regime of racing games, flight simulators, and physics-based environments with almost no dedicated training data. Social deduction games, despite session-level datasets such as Werewolf-18K \citep{wu2024enhancereasoninglargelanguage}, lack the temporally extended records needed for online adaptation and real-time belief modeling. These gaps reflect structural mismatches between data collection infrastructure and domain requirements, and closing them will likely require new recording frameworks for high-frequency, continuous-control, and socially interactive environments.

In summary, the breadth path \citep{ale,genie,nitrogen,openp2p89} provides the data substrate for generative pre-training across a thousand games but exposes the action-space unification problem. The depth path \citep{minedojo52,gamerl,jarvis_vla,openha} provides stable, verifiable reward signals for reliable RL but within a narrower scope. The human annotation or model-as-judge paradigm \citep{minedojo52,gamegenx,oark5,gamefactory_wm} offers a scalable middle ground but remains bounded by teacher-model capability. No single corpus achieves simultaneous leadership in scale, annotation quality, and game diversity. This three-way trade-off is the defining unresolved tension of the dataset pillar, driving the exploration of world-model-based data generation in the next section.

\subsection{Coda --- Creation: The World Model as Data Engine}
Game datasets in Era 3 are still largely produced by human-authored, code-based game engines. This substrate naturally limits scale by engine-side collection efficiency, annotation quality by the gap between the AI learner and the game engine, and diversity by the closed set of already existing human-made games.
Therefore, to push beyond the current boundary of game data acquisition, \emph{the key is no longer to propose another dataset, but to propose another data engine}---one in which AI itself becomes the driver of data generation.
This paradigm shift essentially unifies data generation and model learning, creating a self-amplifying data flywheel that can jointly expand scale, improve quality, and broaden diversity.

\paragraph{Breaking the Code-Engine Ceiling.}
Manual code engines are hard-coded systems to ensure a stable gameplay experience: they provide human-friendly control interfaces, maintain preset game world boundaries for robust interaction, and are typically optimized for a single game world. By contrast, game world models are learned neural simulators that model game dynamics and predict action-conditioned futures, thereby making data generation more flexible and opening a path to relaxing the ceilings imposed by code engines.

First, they relax the \emph{interface ceiling}. In code-based engines, collectable trajectories are restricted to pre-defined keyboard, mouse, controller, or API interfaces. Recent world models begin to absorb action conditioning into the model itself. \textit{Genie}~\citep{genie} is an early landmark in this direction, as it learns interactive environments with latent actions from unlabeled videos rather than relying on explicitly recorded engine-side controls. \textit{MineWorld}~\citep{mineworld_wm} and \textit{Matrix-Game 2.0}~\citep{matrix2-85} further strengthen this trend by emphasizing action following and action injection under explicit control signals, while \textit{PAN}~\citep{pan_wm} extends the action channel beyond fixed low-level interfaces to language-conditioned actions. In this sense, what counts as an admissible action for data generation is increasingly less confined to traditional low-level interfaces, even if some systems still rely on predefined controls.

Second, world models loosen the \emph{preset-boundary ceiling}. Code engines can only operate within mechanics, maps, and interaction patterns that have been explicitly authored in advance. By contrast, recent generative systems begin to instantiate and recombine interactive situations beyond the exact boundaries of logged gameplay. \textit{GameFactory}~\citep{gamefactory_wm} explicitly frames this as creating new games through generative interactive videos, while \textit{GameGen-X}~\citep{gamegenx} pushes toward open-world interactive generation with multimodal control. \textit{PAN}~\citep{pan_wm} broadens this further to general, interactable, and long-horizon world simulation, and \textit{IPR-1}~\citep{ipr1_wm} complements this trend by moving toward game-to-unseen generalization with prediction-reinforced physical reasoning. These works indicate that training worlds no longer need to remain fully confined to the exact environments and interaction patterns specified in advance.

Third, world models begin to push against the single-game ceiling. Traditional datasets are ultimately limited by the finite set of already existing human-made games, whereas recent neural game engines increasingly expand toward more diverse and open-ended worlds. \textit{GameFactory}~\citep{gamefactory_wm}, \textit{GameGen-X}~\citep{gamegenx}, and \textit{PAN}~\citep{pan_wm} all broaden the scope of generated environments beyond narrow single-game settings, while \textit{Solaris}~\citep{solaris_wm} extends this expansion to multiplayer shared-world simulation with cross-player consistency.

\paragraph{Towards Self-Amplifying Data Flywheel.}
Beyond serving as a data engine that relaxes the constraints of code-based environments, the deeper promise of world models lies in entering the training loop itself as a self-amplifying data flywheel. In this setting, the world model is no longer a passive predictor of next states; instead, it becomes a training component that supports \emph{adaptive generation} tailored to the player's demands, while also enabling \emph{closed-loop co-evolution} with the player model.

Early evidence in robotics has already shown that such a paradigm is not only feasible, but also powerful.
On the one hand, recent works of \emph{adaptive generation} suggest that: rather than producing a fixed pool of synthetic rollouts, the world model could generate data conditioned on the learner's weaknesses and demands, thereby improving policy models in long-tail failure-prone scenarios. Ctrl-World~\citep{ctrlworld} shows that imagined successful trajectories can be selectively synthesized for policy improvement, while WMPO~\citep{wmpo} demonstrates that on-policy post-training can be carried out inside a learned world model, allowing the generated training distribution to evolve together with the policy.
On the other hand, world models can also support \emph{closed-loop co-evolution} once they become part of the training loop itself. World-Gymnast~\citep{worldgymnast} explicitly couples world-model rollouts with policy refinement in a reinforcement learning loop, showing that simulator quality and policy capability can be iteratively improved together. A similar perspective is also emerging in embodied systems in Psi-R2 and Psi-W0 \citep{psibotblog}, where the world model is not treated as a standalone predictor, but as an integrated component for policy evaluation, optimization, and autonomous self-improvement within a closed-loop pipeline. This evidence demonstrates the power of world models to function as a self-amplifying data flywheel that co-evolves with policy models.

Game LFM training naturally calls for a data flywheel that is tailored to and evolves with the model.
Games are goal-directed, feedback-driven environments that naturally push players to improve through interaction. This makes adaptive training data especially valuable, as it can target the player's current weaknesses, failure modes, and learning demands. Since game difficulty often evolves with player competence, games also naturally motivate player--simulator co-evolution.
However, current game world models still face substantial challenges before they can serve as such a data flywheel. High-quality training data require neural game engines to maintain temporal persistence and spatial consistency across the game multiverse, rather than merely generating locally plausible clips. Moreover, shared-world modeling that supports coherent multi-player rollouts remains underexplored, yet it is essential for scaling the flywheel toward industrial-level rollout throughput and richer collaborative or competitive training scenarios.
Further advances in game world models will enable neural data engines to internalize more faithful game dynamics, ultimately forming a self-amplifying data flywheel that is tailored to and co-evolves with the game LFMs.

\paragraph{Synthesis: support, supervision, and transfer.}
Generality depends on support over decisions, not only the number of titles. NitroGen's extracted actions carry overlay delays, parsing errors, and player-specific bindings~\citep{nitrogen}; these can couple similar observations to different control labels. More hours may therefore improve frequent motor patterns without covering rare goals or unfamiliar mechanics. Open-P2P's logged inputs avoid inferring actions from pixels, but its collection policy still determines which states and goals are represented~\citep{openp2p89}. A discriminating experiment would hold model size, training compute, and deployment protocol fixed, then vary mechanics diversity and annotation source separately at matched data volume. Testing on held-out games and goal compositions, with per-game competence and threshold coverage as in Eqs.~\ref{eq:normalized-competence}--\ref{eq:generalist-criterion}, would distinguish broader support from deeper mastery of familiar play. Label audits stratified by game and action type would test whether failures concentrate where supervision is unreliable. For synthetic trajectories, evaluation must return to the original game engine: improvements confined to the learned simulator cannot establish transfer across dynamics.

\section{Models: The Evolving Brain of Generalist Game Player}

\begin{figure}
    \centering
    \includegraphics[width=0.9\linewidth]{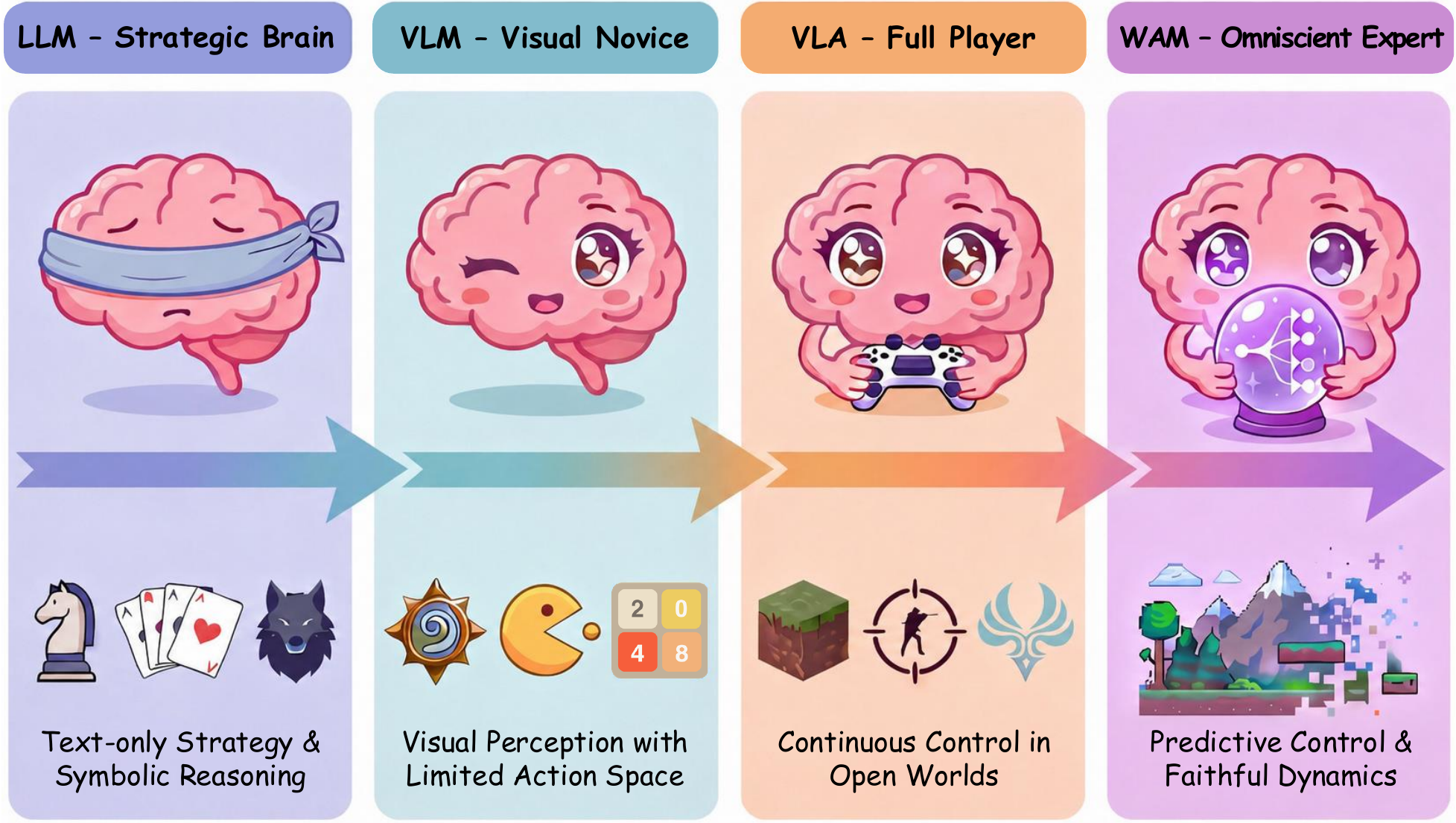}
    \caption{\textbf{Overview of Models in Game-playing AI.} Game-playing models progress from LLMs as strategic brains for text-based reasoning, to VLMs as visual novices that perceive game screens, then to VLAs as full players that execute low-level controls, and finally to WAMs as predictive experts that internalize game dynamics for world-aware decision-making.}
    \label{fig:model_teaser}
\end{figure}

Having established the paradigm shift towards generalist game players, we now turn to the cognitive core of these agents: the \textbf{Model}. The architectural question is which capabilities transfer when observations, action meanings, or transition dynamics change. In traditional paradigms, game AI was locked to rigid interfaces. A chess engine takes a fixed board vector and returns a move index; a StarCraft agent reads a state tensor and emits an action ID. These systems are powerful within their contracts, but the contract is the cage. Change the board size, swap the piece types, or ask the engine to explain its reasoning in natural language, and the entire architecture collapses. Its input parser, action head, and reward function are all fixed to a single game. In contrast, building foundation agents requires architectural unification, including designing a singular model capable of processing unconstrained multimodal observations ($\Omega$), reasoning over open-ended goals ($G$), and executing through universal human-computer interfaces ($A$).

Within the current landscape, this pursuit of a unified architecture has evolved through four progressive stages. Each stage addresses a critical bottleneck in closing the perception-action loop:

\begin{itemize}
    \item \textbf{Large Language Models (LLMs):} Acting as the foundational cognitive engine, LLMs introduce world knowledge and strategic planning to games. However, restricted to textual inputs and outputs, they remain visually disconnected---capable of complex reasoning but blind to raw visual environments.
    
    \item \textbf{Vision-Language Models (VLMs):} By integrating visual encoders, VLMs equip the agent with direct pixel-level perception. While achieving multimodal grounding, their outputs typically remain semantic, creating a disconnect between visual understanding and continuous motor execution, such as keyboard, mouse and controller.
    
    \item \textbf{Vision-Language-Action Models (VLAs):} VLAs bridge the execution gap by mapping multimodal observations directly to low-level continuous control (e.g., unified keyboard and mouse operations). This end-to-end architecture seamlessly unifies perception, reasoning, and action within a single stream.
    
    \item \textbf{World Action Models (WAMs):} Representing the emerging frontier, WAMs move beyond reactive control by internalizing the environment's transition dynamics ($T$). By simulating future states and evaluating potential trajectories, these models enable predictive, model-based planning prior to execution.
\end{itemize}

\rev{We assign a model to a class by the most advanced stage of the perception--action loop it closes \emph{natively}: text reasoning (LLM), visual perception (VLM), low-level action $A$ (VLA), or learned dynamics $T$ (WAM). These classes are cumulative rather than mutually exclusive, so we classify a hybrid system by its most advanced native capability and cross-reference it under the others; a pipeline that merely wraps a VLM with external memory or planning is treated as a Harness contribution (Section 5), not a new model class.}

In the following subsections, we systematically review these four architectural paradigms. We explore their structural designs, training methodologies, perception-action loop designs, and the inherent limitations that continue to drive the evolution of generalist models.

Table~\ref{tab:paper_model} includes world models (WMs), non-player character (NPC) systems, and renderers alongside player models. Coachable Agents and ActSWM provide task-specific control comparisons~\citep{coachableagents2026,actswm2026}.

\begin{table*}[!t]
\centering
\caption{Summary of representative models in gaming AI}
\label{tab:paper_model}
\begingroup
\setlength{\tabcolsep}{4pt}
\renewcommand{\arraystretch}{1.08}
\resizebox{\textwidth}{!}{%
\begin{tabular}{lllllll}
\toprule
\textbf{Paper} & \textbf{Arch.} & \textbf{Base Model} & \textbf{Input} & \textbf{Action} & \textbf{Training Recipe} & \textbf{Game Set} \\
\midrule
CICERO~\citeyearpar{cicero} & LLM & BART-like 2.7B & Text state + dialogue & NL messages + orders & RL + planning (piKL) & Diplomacy \\
Voyager~\citeyearpar{voyager8} & LLM & GPT-4 & Text API & Code (skills) & Training-free & Minecraft \\
Werewolf-RL~\citeyearpar{DBLP:conf/icml/Xu000W24} & LLM & GPT-4 + RL Net & Text log & Dialogue & MAPPO & Werewolf \\
LLaMA-Rider~\citeyearpar{llamarider90} & LLM & LLaMA-2-70B & Text API & Text API calls & SFT & Minecraft \\
Chain of Summ.~\citeyearpar{ma2024large} & LLM & GPT-3.5/4 + Qwen-7B & Text (TextSC2) & RTS commands & SFT (open-source) & StarCraft II \\
Werewolf LSPO~\citeyearpar{xu2025learning} & LLM & LLaMA-3-8B & Text dialogue & Strategic utterances & CFR + DPO & Werewolf \\
Refl.\ of Episodes~\citeyearpar{xu2026reflectionepisodeslearningplay} & LLM & GPT-3.5-Turbo & Text (keyframes) & RTS commands & Training-free & StarCraft II \\
Mastermind~\citeyearpar{wang2025mastermind} & LLM & LLaMA-2-7B & Text board state & Game moves & SFT (synth.\ data) & Doudizhu, Go \\
Metamon~\citeyearpar{metamon} & LLM & Transformer 200M & Text battle state & Move selection & Offline RL & Pok\'{e}mon \\
Orak~\citeyearpar{oark5} & LLM & Qwen-2.5-72B & Text / image & Game actions & SFT & 12 genres \\
Think in Games~\citeyearpar{liao2025think} & LLM & Qwen-2.5-7B/32B & Text state & Game policies & SFT + RL & Multiple \\
SpinGPT~\citeyearpar{maugin2025spingpt} & LLM & LLaMA-3 & Text game state & Poker decisions & SFT + RL & Poker \\
MS-GRPO~\citeyearpar{dilkes2025reinforced} & LLM & Qwen2.5-3B & Text state & Sequential decisions & MS-GRPO & Snake, Frozen Lake \\
MARSHAL~\citeyearpar{yuan2025marshal} & LLM & Qwen3-4B & Text scenarios & Strategic decisions & Self-play RL & Multiple \\
Stronger-MAS~\citeyearpar{zhao2025strongermas} & LLM & Qwen3-1.7B/8B & Text prompts & Multi-agent actions & AT-GRPO & Multiple \\
GameTalk~\citeyearpar{conchellovendrell2026gametalk} & LLM & LLaMA-3-3B & Text dialogue & Strategic dialogue & GRPO + DPO & Dialogue games \\
VAM~\citeyearpar{zhang2026vam} & LLM & Qwen2.5-3B/7B & Text (chess pos.) & Chess moves & RL + action masking & Chess \\
ISO-GRPO~\citeyearpar{xia2026implicit} & LLM & Qwen2.5-0.5B & Text game state & Poker decisions & SFT + ISO-GRPO & Poker, Pok\'{e}mon \\
Mixture of Masters~\citeyearpar{frisoni2026mixture} & LLM & Custom GPT (small) & Text (chess pos.) & Chess moves & RL + MoE routing & Chess \\
Creative Agents~\citeyearpar{creative_agents} & VLM & GPT-4V + Diffusion & Image+Text+Goal image & Code action & Prompting & Minecraft \\
TeamCraft~\citeyearpar{long2024teamcraft} & VLM & Vicuna-7B/13B & Vision + text & Multi-agent actions & Imitation Learning & Minecraft \\
VARP~\citeyearpar{wukong55} & VLM & 3 Models & Game Screenshot & Python kbd/mouse actions & Prompting+Retrieval & Black Myth: Wukong \\
ROCKET-1~\citeyearpar{rocket1} & VLM & GPT-4o + SAM-2 & Pixel image+Mask & Keyboard\&Mouse & SFT & Minecraft \\
GameSense~\citeyearpar{gamesense} & VLM & Qwen2.5-VL & Pixel image & Keyboard\&Mouse & Prompting+Module RL & 3 action games \\
Cradle~\citeyearpar{cradle14} & VLM & GPT-4V & GUI Screenshot & Keyboard\&Mouse code & Prompting & 4 commercial games \\
ViGaL~\citeyearpar{xie2026play} & VLM & Qwen2.5-VL-7B & Image + text & Game actions & RL & Arcade games \\
G1~\citeyearpar{g1} & VLM & Qwen2.5-VL-7B & Game Screenshot & Text action & SFT+GRPO & 4 visual games \\
CoSo~\citeyearpar{coso} & VLM & LLaVA-1.6 & Screenshot+Text & Text action & Online RL & Gym Cards \\
Game-RL~\citeyearpar{gamerl} & VLM & Qwen2.5-VL-7B & Game image+Text QA & Text answer & GRPO & 30 games \\
VistaWise~\citeyearpar{vistawise} & VLM & GPT-4o & Text API + KG & API calls & Training-free & Minecraft \\
Orak~\citeyearpar{oark5} & VLM & Qwen-2.5-72B & Text / image & Game actions & SFT & 12 genres \\
VL-DAC~\citeyearpar{vldac} & VLM & Qwen2-VL-7B & Screenshot+Text & Text action & PPO & MiniWorld\&Gym-Cards \\
Proact-VL~\citeyearpar{yan2026proactvlproactivevideollmrealtime} & VLM & LiveCC-7B-Base & Video stream & Commentary / guidance & SFT & 12 live-streamed games \\
BC-CSGO~\citeyearpar{csgo} & VA & EfficientNet\&ConvLSTM & Pixel image & Keyboard\&Mouse & SFT & CSGO \\
VPT~\citeyearpar{vpt} & VA & from scratch(Transformer-XL-style) & Pixel image & Keyboard\&Mouse & PT+SFT+PPG & Minecraft \\
GROOT~\citeyearpar{groot51} & VA & VPT & Pixel image & Keyboard\&Mouse & SFT & Minecraft \\
\rev{Pixels2Play-0.1}~\citeyearpar{p2p} & VA & from scratch(Decoder-only Transformer) & Pixel image & Keyboard\&Mouse & PT & Roblox \& MS-DOS games \\
NitroGen~\citeyearpar{nitrogen} & VA & Siglip\&DiT & Pixel image & Gamepad & PT+SFT & 1000+ video games \\
STEVE-1~\citeyearpar{steve1} & VLA & VPT\&MineCLIP & Pixel image+Text & Keyboard\&Mouse & SFT & Minecraft \\
OmniJarvis~\citeyearpar{omnijarvis} & VLA & LLaVA-7B & Pixel image+Text+Action & Action token & SFT & Minecraft \\
SIMA~\citeyearpar{raad2024scaling} & VLA & SPARC\&Phenaki\&Transformer-XL & Pixel image+Text & Keyboard\&Mouse & SFT & Multiple 3D games \\
\rev{Pixels2Play-0.3}~\citeyearpar{p2p03} & VLA & from scratch(EfficientNet\&Transformer) & Pixel image+Text & Keyboard\&Mouse & PT+SFT & Roblox \& MS-DOS \& FPS games \\
SIMA2~\citeyearpar{sima2-87} & VLA & Gemini Flash-Lite & Pixel image+Text & Keyboard\&Mouse & SFT+RL & Multiple 3D games \\
JarvisVLA~\citeyearpar{jarvis_vla} & VLA & Qwen2-VL-7B & Pixel image+Text & Action token & CPT+SFT & Minecraft \\
OpenHA~\citeyearpar{openha} & VLA & Qwen2-VL-7B & Pixel image+Text & Multiple action space & PT+SFT & Minecraft \\
CombatVLA~\citeyearpar{combatvla} & VLA & Qwen2.5-VL-3B & Pixel image+Text & Text Action & SFT & Wukong \& Sekiro \\
Lumine~\citeyearpar{lumine} & VLA & Qwen2-VL-7B & Pixel image+Text & Text K\&M Action & CPT+SFT & Genshin Impact \\
Game-TARS~\citeyearpar{gametars} & VLA & Qwen2.5-VL-7B\&Seed-VL-1.5 & Pixel image+Text & Code Action & CPT+SFT+RFT & 500+ video games \\
UI-TARS1.5~\citeyearpar{ui-tars-15} & VLA & Qwen2.5-VL-7B & Pixel image+Text & Code Action & SFT+RL & 14 diverse games \\
UI-TARS2~\citeyearpar{uitars119} & VLA & Qwen2.5-VL-7B & Pixel image+Text & Text Action & SFT+PPO & 15+ 2D games \\
\rev{Open-P2P}~\citeyearpar{openp2p89} & VLA & from scratch(Decoder-only Transformer) & Pixel image+Text & Keyboard\&Mouse & PT+SFT & Multiple 3D games \\
Main-VLA~\citeyearpar{main_vla} & VLA & Qwen2-VL-7B & Pixel image+Text & Keyboard\&Mouse & SFT & Minecraft \& Peace \& Valorant \\
Odysseus~\citeyearpar{odysseus2026} & VLA & Qwen3-VL-8B-Instruct & Screenshot + prompt & Controller-button combinations & SFT + PPO & Mario Land / Mario Bros. \\
GameScaling~\citeyearpar{zhang-scaling-games} & VLA & Qwen3.5 (0.8B--27B) & Frame + rules + history & 200-ms keyboard/mouse & CPT & 80-game corpus \\
Player-1~\citeyearpar{player1_2026} & VLA & Qwen3.5-9B / 35B-A3B & Image/text + history & Keyboard/mouse & SFT + PPO & 172 demonstration + 16 RL games \\
Genie~\citeyearpar{genie} & WM & Tokenizer + Transformer & Unlabeled video & Latent actions & Video pretraining & 2D platformers \\
Oasis~\citeyearpar{oasis83} & WM & Autoregressive Transformer & Game frames & Keyboard\&Mouse & Autoregressive video modeling & Minecraft-like world \\
GameNGen~\citeyearpar{gamengen_wm} & WM & Diffusion model & DOOM frames + actions & Game actions & Diffusion next-frame modeling & DOOM \\
MineWorld~\citeyearpar{mineworld_wm} & WM & Autoregressive Transformer & Frames + actions & Keyboard\&Mouse & Action-conditioned pretraining & Minecraft \\
GameFactory~\citeyearpar{gamefactory_wm} & WM & Video diffusion & Text/image/video prompts & Interactive actions & Diffusion fine-tuning & Generated games \\
Matrix-Game 3.0~\citeyearpar{matrix3_wm} & WM & Diffusion Transformer & Frames + actions + memory & Keyboard\&Mouse & AR diffusion + distillation & Minecraft \\
GameGen-X~\citeyearpar{gamegenx} & WM & Video diffusion & Multimodal prompts & Interactive actions & Diffusion fine-tuning & Open-world games \\
PAN~\citeyearpar{pan_wm} & WM & Long-horizon WM & History + text actions & Language actions & Long-horizon simulation & General worlds \\
WorldCam~\citeyearpar{worldcam_wm} & WM & Autoregressive Transformer & Frames + camera pose & Camera controls & Pose-aware AR training & 3D game worlds \\
ReactiveGWM~\citeyearpar{reactivegwm2026} & NPC/WM & Wan2.2-TI2V-5B & Frame/mask + ctrl/rules & Controls -> reactive video & Flow matching + role binding & Hikou no Mizu / SF III \\
Agora-1~\citeyearpar{agora1_2026} & WM & State model + DiT renderer & State + player actions & Controls -> state/video & Dynamics + rendering training & GoldenEye simulation \\
SCOPE~\citeyearpar{scope2026} & WM & Wan2.2-TI2V-5B & Video/text + telemetry & Gamepad -> video & Spatial video fine-tuning & 7 FPS games \\
StatePlay~\citeyearpar{stateplay2026} & WM & Wan2.2-5B + state branch & Frame/state + text/action & Controls -> state/video & Video flow + state regression & Street Fighter \\
WorldMind~\citeyearpar{worldmind2026} & NPC/WM & Gemma-E2B + Wan2.2-5B & RGB + history + controls & NPC plans + video & Perception + video distill. & Game A / Hollow Knight / Isaac \\
Odyssey-3~\citeyearpar{odyssey32026} & WM+policy & Proprietary AR diffusion Transformer & Recent frames & Keyboard/mouse & Frozen WM + decoder training & GTA V / RDR2 / Sleeping Dogs \\
Agora-2~\citeyearpar{agora2_2026} & WM & State Transformer + 1B renderer & State + actions + geometry & Controls -> state/video & Dynamics supervision + flow matching & Diablo II simulation \\
GameDirector~\citeyearpar{gamedirector2026} & NPC/WM & Gemma-E2B + Wan2.2-5B & Video + state + controls & Boss skills + video & Perception + video distill. & 3 games (N/V/H) \\
Matrix-Game 4.0~\citeyearpar{matrixgame4_2026} & Renderer & Wan2.2-Fun-5B Control & Proxy/text + history & Controls -> video & SFT + causal distillation & Shooter / driving \\
DreamerV3~\citeyearpar{dreamerv3_wm} & WAM & RSSM & Pixels / states & Game actions & World-model RL & Atari, Minecraft, DMLab \\
JOWA~\citeyearpar{jowa_wm} & WAM & World-action Transformer & Offline trajectories & Future actions & Joint WM-action pretraining & Atari \\
DIAMOND~\citeyearpar{diamond_wm} & WAM & Diffusion WM & Atari pixels & Atari actions & In-model policy training & Atari \\
GameWAM~\citeyearpar{gamewam2026} & WAM & Wan2.2-5B + Action DiT & Vision + text + history & Keyboard/mouse + GUI & Joint video/action flow matching & Minecraft / ViZDoom \\
Coachable Agents~\citeyearpar{coachableagents2026} & RL & Domain-specific actor--critic & Observations + style parameters & Domain-specific controls & SAC / QR-SAC / Cat-RAC & HFW / GT7 / humanoid \\
ActSWM~\citeyearpar{actswm2026} & WM+MPC & ViT-tiny + predictor & Pixels + history + goals & CEM action blocks & Latent prediction + constraints & Minecraft; 3 offline games \\
\bottomrule
\end{tabular}
}
\endgroup
\end{table*}

\subsection{LLMs as Game Brain: Reasoning the Strategies in the Textual Multiverse}

Large Language Models (LLMs) ~\citep{GPT-3-75,GPT-3.5-76,qwen1-78} remove the constraint in Deep Reinforcement Learning. They take free-form text in and produce free-form text out. The same model can read a chess position \citep{gamerl}, a Werewolf transcript \citep{wu2024enhancereasoninglargelanguage}, a Pokemon battle log \citep{PokeGym}, or a Minecraft inventory \citep{llamarider90}, all without architectural change. Both observation and action live in natural language, making them in principle unbounded. Moreover, LLMs also bring broad world knowledge. Pre-training on game manuals, strategy forums, tournament commentary, and community wikis means the model already understands bluffing before it plays its first poker hand, and knows where diamonds spawn before it enters a Minecraft cave \citep{voyager8}. These two properties,\textbf{\textit{ flexible interface}} and \textbf{\textit{pre-trained knowledge}}, open the possibility of a single reasoning engine that generalizes across the textual multiverse: all games whose states and actions can be faithfully expressed in text. 

This section traces how far training-based methods have pushed LLMs toward this goal, along two threads: natively textual games, where the challenge is strategic depth; and visual games accessible through text, where the challenge shifts to model design under perceptual constraints.

\textbf{The Native Territory: Text Multiverse of Games.} In text games, the observation is text, the action is text, and the core challenge is reasoning. This makes them the natural arena for sharpening LLM strategic capability through training.

Social deduction games demand unconstrained natural language as the action space: the agent persuades, deceives, and detects lies through dialogue, making LLMs irreplaceable. The landmark is CICERO \citep{cicero}, which first achieved human-level performance in Diplomacy via game-theoretic planning. Subsequent shifts moved from modular designs \citep{DBLP:conf/icml/Xu000W24} to unified alignment, where Werewolf LSPO \citep{xu2025learning} and GameTalk \citep{conchellovendrell2026gametalk} integrate strategies directly into LLMs via DPO and GRPO.

In imperfect information games (e.g., Poker), research focuses on long-horizon credit assignment. The GRPO \citep{deepseekr1} family, including MS-GRPO \citep{dilkes2025reinforced} and ISO-GRPO \citep{xia2026implicit}, addresses sparse rewards, while SpinGPT \citep{maugin2025spingpt} standardizes the SFT-RL pipeline. Offline learning also thrives, with Metamon \citep{metamon} reaching expert levels in Pokemon via ranked replays. For board games, models like Mastermind \citep{wang2025mastermind} and Mixture of Masters \citep{frisoni2026mixture} leverage algorithmic synthesis and MoE architectures to provide explainable strategic reasoning. VAM \citep{zhang2026vam} masks illegal chess moves during RL to focus exploration on strategic quality. However, text games also expose reasoning limits. Credit assignment over hundreds of turns remains unsolved. 

Moreover, MARSHAL \citep{yuan2025marshal} and Stronger-MAS \citep{zhao2025strongermas} train on self-play from multiple games, testing cross-domain transfer. Game-RL \citep{gamerl} spans 30 games with Code2Logic verifiable rewards; ViGaL \citep{xie2026play} and Think in Games \citep{liao2025think} show that game RL improves general reasoning beyond games, suggesting games are training grounds for broader intelligence. Yet, multi-agent coordination degrades with agent count and belief-nesting depth. Most fundamentally, tracking what each player knows, believes, and intends across long interactions imposes a combinatorial cognitive load that current context windows struggle to support. These reasoning bottlenecks define the frontier in the textual multiverse.

\textbf{Expanding the Frontier: LLMs in Visual Games.} Most modern games are visual. LLMs cannot see pixels but can play when game states reach them as text, through APIs (Mineflayer \citep{mineflayer}, TextStarCraft II \citep{ma2024large}), text protocols (Pokemon Showdown), or perception modules. In this pipeline, the LLM acts as the reasoning core for planning and decomposition. Minecraft remains the central benchmark: Voyager \citep{voyager8} introduced autonomous skill libraries, followed by LLaMA-Rider \citep{llamarider90} for exploration and JARVIS-1 \citep{wang2023jarvis1} for memory-augmented planning. Advances in VistaWise \citep{vistawise} and TeamCraft \citep{long2024teamcraft} further integrated cross-modal knowledge and multi-agent scaling. In complex RTS environments like StarCraft II, Chain of Summarization \citep{ma2024large} and Reflection of Episodes \citep{xu2026reflectionepisodeslearningplay} tackle context limits through hierarchical abstraction and episodic memory.

Despite successes in multi-game training \citep{oark5}, \textbf{\textit{a perceptual ceiling persists}}. Textual mediation often discards critical spatial relations and real-time visual cues, limiting the agent's ability to master continuous dynamics. This bottleneck necessitates the transition to VLMs and VLAs, which integrate pixel-level perception and end-to-end motor control.

\subsection{VLMs as Game Novice: Opening the Eyes to the Visual Multiverse}

While LLM-based agents excel at planning and reasoning in text-mediated games, their reliance on text-converted state representations inevitably discards the spatial, temporal, and perceptual details that human players routinely extract from visual feedback. This perceptual bottleneck limits generalization across visually diverse game environments. Vision-Language Models (VLMs)~\citep{llava,qwen2.5-vl,qwen3vl24} address this gap by integrating visual encoders with language understanding, enabling agents to perceive and reason about game worlds directly from pixels.

\textbf{VLMs as Perceptual Reasoners for Game Decision-Making.} An early exploration in this direction is MineDojo~\citep{minedojo52}, which trains a contrastive video-language model, MineCLIP, to score the alignment between an agent's video trajectory and a language instruction, serving as a dense reward for reinforcement learning without manual reward design. This work demonstrates that vision-language alignment can effectively guide agents across diverse open-ended tasks, establishing the potential of joint vision-language learning for gameplay.

Subsequent works further leverage VLMs as the perceptual and reasoning backbone. Creative Agents~\citep{creative_agents} pairs language and diffusion-based goal image generation with a VLM code controller for creative building in Minecraft. Cradle~\citep{cradle14} introduces the General Computer Control setting, where a VLM reasons over screenshots and outputs keyboard-mouse actions as code, generalizing across commercial games without game-specific APIs. VARP~\citep{wukong55} applies VLM-driven action planning to ARPG combat in \textit{Black Myth: Wukong}, combining a visual action planning module with a visual trajectory tracking system that jointly translates raw screen observations into timed combat decisions. ROCKET-1~\citep{rocket1} addresses the spatial information bottleneck of language by introducing visual-temporal context prompting, where a VLM communicates interaction targets to a low-level policy via segmentation masks. GameSense~\citep{gamesense} shifts the VLM from a per-step controller to a developer that synthesizes reusable action-feedback modules for real-time play. Proact-VL~\citep{yan2026proactvlproactivevideollmrealtime} further extends VLMs to real-time streaming, building a proactive video language model that processes continuous gameplay and autonomously decides when to respond, serving as an interactive AI companion.

\textbf{Reinforcement Learning for Sharper Visual Game Reasoning.} The recent success of reinforcement learning (RL) in improving VLM reasoning has also influenced game agent research. G1~\citep{g1} trains VLMs via RL self-evolution in a multi-game environment, showing that perception and reasoning mutually bootstrap during training. CoSo~\citep{coso} improves exploration efficiency by using counterfactual reasoning to focus RL updates on action-critical tokens. Game-RL~\citep{gamerl} synthesizes verifiable reasoning data from game source code and applies GRPO-based training~\citep{grpo}, finding that RL on game data alone transfers to broader vision-language benchmarks. VL-DAC~\citep{vldac} decouples token-level PPO~\citep{ppo} updates from environment-step value estimation, demonstrating that RL training in lightweight simulators generalizes to real-image agentic control. Collectively, these efforts confirm that RL post-training consistently sharpens VLM decision-making in game settings and can yield benefits beyond the game domain.

Despite these advances, VLM-based game agents still rely on indirect execution mechanisms such as code generation, API calls, or predefined skill libraries and functions to translate high-level understanding into concrete actions. This decoupled architecture introduces latency, limits control granularity, and confines VLM agents to the role of slow-frequency planners. Realizing a truly general game agent thus still requires closing the perception-action loop within an end-to-end framework.

\subsection{VLAs as Game Player: Closing the Perception-Action Loop}

While VLMs provide visual perception, their outputs remain semantic and require external modules for motor execution. Vision-Language-Action models (VLAs)~\citep{rtx,openvla} close this gap by mapping multimodal observations directly to low-level control~\citep{DifusionPolicty81,Su_2025_ICCV} within end-to-end architectures. This paradigm builds on a sustained line of vision-to-action (VA) research that learns sensorimotor policies from gameplay.

\textbf{Scaling Vision-to-Action Pre-training from Videos.} Early VA research has explored video-based pre-training as a scalable route to sensorimotor policy learning. CSGO~\citep{csgo} demonstrates that behavioral cloning (BC)~\cite{bcz} can scale to millions of human gameplay frames to produce competent FPS agents. VPT~\citep{vpt} then extends BC by leveraging an inverse dynamics model to pseudo-label massive internet gameplay videos and training a causal policy network on the resulting corpus, confirming the potential of internet-scale video pre-training for action generation. \
STEVE-1~\citep{steve1} builds on VPT with goal-conditioned fine-tuning, showing that the paradigm scales to multi-step task compositions. 
GROOT~\citep{groot51} extends goal conditioning by using reference gameplay videos as instructions, demonstrating that video pre-training can simultaneously provide the policy prior.
NitroGen~\citep{nitrogen} further broadens the scope of VA paradigm, training a joint ViT-DiT vision-action architecture on 40K hours of gameplay spanning over 1,000 titles to achieve general-purpose continuous action generation for smoother and higher-resolution control. In light of the above VA advances, \rev{Open-P2P~\citep{openp2p89}\footnote{\normalcolor The same group released \emph{Pixels2Play-0.1}~\citep{p2p}, \emph{Pixels2Play-0.3}~\citep{p2p03}, and \emph{Open-P2P}~\citep{openp2p89}; we refer to each by its release name.}} provides a systematic investigation of BC scaling laws, showing that increasing model capacity and data volume not only improves imitation fidelity but also promotes the emergence of causal reasoning over spurious correlation.

\textbf{Leveraging Multimodal Understanding for Action Generation in VLAs.} Building on the VA foundations, contemporary VLAs retain the end-to-end sensorimotor backbone while further integrating language comprehension into a unified multimodal policy, OmniJARVIS~\citep{omnijarvis} pioneers the paradigm by processing multimodal inputs through a shared encoder-decoder architecture and training action outputs via BC, with 
SIMA~\citep{raad2024scaling} validating the generality of this paradigm by training a language-conditioned agent across diverse commercial video games. 
This framework is quickly inherited by JARVIS-VLA~\citep{jarvis_vla} and OpenHA~\citep{openha}, which further leverage pre-trained VLMs as the perceptual encoder and append lightweight policy decoders to generate actions. 

The adoption of VLMs raises whether reinforcing their domain-specific understanding can benefit downstream action generation. CombatVLA~\citep{combatvla} distills tactical reasoning into learned representations through a progressive \textit{Action-of-Thought} curriculum; MAIN-VLA~\citep{main_vla} prunes redundant visual and linguistic semantics into compact, action-critical features; and Lumine~\citep{lumine} unifies language reasoning with action generation in a closed-loop framework, achieving instruction-adherent long-horizon control with cross-game transfer. These results indicate that deepening the perceptual and reasoning capacity of VLM encoders consistently improves both long-horizon interaction and cross-game generalizability.

Building on these supervised foundations, recent work introduces RL post-training to enable self-improvement beyond demonstration data---whether through self-generated gameplay across 3D virtual worlds~\citep{sima2-87} or multi-turn RL within GUI environments at scale~\citep{uitars119}. This supervised pre-training followed by RL post-training pipeline has become a prevalent recipe for generalist VLA agents, yielding policies that continuously adapt across heterogeneous interactive environments.

\textbf{Toward Real-Time Control: Evolving Action Representations.} To achieve better real-time control, the development of game VLAs has also focused on improving action representation. Early VLAs such as JARVIS-VLA~\citep{jarvis_vla} and Game-TARS~\citep{gametars} discretize control into per-step tokens generated autoregressively alongside language output. To reduce the resulting inference overhead, subsequent work shifts toward \emph{action chunks}: CombatVLA~\citep{combatvla} applies truncated decoding of action-of-thought sequences for a 50-fold speedup in 3D ARPG combat, while Lumine~\citep{lumine} predicts six 33,ms textual chunks per step to sustain 30,Hz keyboard-mouse control. This trajectory culminates in NitroGen~\citep{nitrogen}, which bypasses language-space encoding entirely and employs a flow-matching diffusion transformer to generate 16-step chunks of \emph{continuous} gamepad actions from a single RGB frame. Collectively, these advances chart a path from discrete, per-token action decoding toward unified continuous representations capable of accommodating diverse control modalities at real-time rates.

\subsection{WAMs as Game Expert: Internalizing the Dynamics of the Game Multiverse}

VLAs have achieved strong semantic reasoning and fine-grained control. However, their understanding of game-world evolution still remains largely semantic, and their control generation is predominantly reactive. WAMs offer a new paradigm to overcome these limitations in two aspects:
\begin{itemize}
    \item \textbf{Predictive Control:} By internalizing action-conditioned dynamics, WAMs support \emph{prediction before action}, so that decisions are guided by future anticipation rather than only current states.
    \item \textbf{Faithful Dynamics:} WAMs learn \emph{world models} of games from visual and control signals, leading to more faithful dynamics modeling of how game worlds evolve under interaction.
\end{itemize}
More importantly, these two properties also make WAMs inherently more \emph{general}: instead of merely fitting game-specific action patterns, they capture transferable regularities of how game worlds evolve under intervention across diverse games. In this sense, WAMs shift game agents from reactive perceive-and-act systems to predictive plan-and-execute systems, marking a crucial step toward more generalist game players.

\paragraph{Early Prototypes of Predictive Control.}
Early world-model-based game agents marked the first step toward predictive game playing by introducing dynamics internalization and future modeling into control. In this sense, they can be viewed as prototypes of gaming WAMs.
Action-Conditional Video Prediction~\citep{acvp_wm} first made the dependency between future frames and control variables explicit, training convolutional and recurrent predictors to roll out Atari futures conditioned on candidate actions. World Models~\citep{worldmodel82} compressed visual observations with a VAE, learned latent dynamics with an MDN-RNN, and evolved a compact controller inside the learned ``dream'' environment. SimPLe~\citep{simple_wm} turned this idea into an iterative Atari training loop, repeatedly fitting a stochastic video model from real interaction and training PPO on short imagined rollouts. The Dreamer series~\citep{dreamer_wm,dreamerv2_wm,dreamerv3_wm} jointly trains an encoder, representation function, and dynamics function within recurrent latent state-space models, enabling implicit predictive control through imagined actor-critic learning across increasingly diverse control tasks. IRIS~\citep{iris_wm} replaced recurrent latent prediction with a transformer world model over discrete image tokens, and DIAMOND~\citep{diamond_wm} further showed that diffusion-based dynamics can preserve action-relevant visual details and train Atari agents entirely inside a learned world model.

Their shared contribution lies not in a modern unified world-action architecture that jointly maps historical observations and actions to future observations and actions, but in demonstrating that prediction-guided control can substantially improve control effectiveness. By grounding control in predicted futures, these works establish the early prototype of gaming WAMs. Nevertheless, limited generative capacity and insufficient dynamics-modeling ability prevented these early models from fully internalizing complex, visually grounded dynamics across a broader set of modern games with complex spatial layouts and rich interaction mechanisms.

\paragraph{Gaming World Models with Faithful Dynamics.}
While early prototypes had already demonstrated the effectiveness of predictive control, recent gaming WMs have strengthened the dynamics model itself.
Genie~\citep{genie} established this direction by learning latent actions and autoregressive dynamics from unlabeled videos, showing that interactive environments can emerge without explicit action labels. This autoregressive paradigm was later grounded in Minecraft-like sandbox worlds by Oasis, MineWorld, and the Matrix-Game series~\citep{oasis83,mineworld_wm,matrix1-84,matrix2-85,matrix3_wm}, which progressively improved action-conditioned generation, action following, streaming rollout, and long-horizon memory, moving gaming WMs from video continuation toward controllable neural environments.
As domains become visually richer, diffusion- or flow-style neural engines further improve fidelity and controllability. GameNGen~\citep{gamengen_wm} shows that a diffusion model trained on DOOM traces can operate as a real-time neural game engine for FPS dynamics, while GameFactory and GameGen-X~\citep{gamefactory_wm,gamegenx} use video-generation backbones and multi-stage training to extend world modeling toward open-domain, modern 3D, and open-world games.
Beyond real-time controllability, recent works increasingly target persistent dynamics. PAN and Model as a Game~\citep{pan_wm,model_as_a_game_wm} emphasize long-horizon interaction, mechanic consistency, and numerical/spatial reliability; WorldPlay and WorldCam~\citep{sun2025worldplay,worldcam_wm} strengthen geometric consistency and spatial revisitation, with WorldCam using camera-pose control for long-horizon 3D grounding; and Solaris~\citep{solaris_wm} extends world modeling to synchronized multiplayer Minecraft with coherent shared states.
These advances suggest that dynamics modeling of modern games with complex spatial structures and interaction mechanisms is no longer fragile. Rather, world modeling for modern games is becoming increasingly reliable, providing the foundation on which gaming WAMs can internalize, predict, and ultimately control.


\paragraph{Towards Modern Gaming World Action Models.}
World Action Models (WAMs) jointly model future world states and future actions from interaction history, enabling prediction and control to be learned within a unified framework rather than as separate modules. 
Promising evidence has already emerged in robotics, where works such as DreamZero \citep{dreamzero} couple predictive world prediction with action generation, improving embodied control and cross-task generalization.
In games, the Dreamer series~\citep{dreamer_wm,dreamerv2_wm,dreamerv3_wm} provides an important prototype of this paradigm. Although it does not fully unify world-action anticipation, it demonstrates the effectiveness of acting on latent states shaped by internalized future dynamics. JOWA~\citep{jowa_wm} takes a more direct step toward modern gaming WAMs by jointly pretraining world and action predictions, showing that world-action coupling can improve decision-making in game environments.
GameWAM jointly trains video and action prediction but uses action-only online inference, while the Odyssey-3 preview trains a policy decoder on a frozen world model~\citep{gamewam2026,odyssey32026}.
Nevertheless, gaming WAMs remain largely underexplored. Existing efforts have not yet fully leveraged the stronger world-model substrate emerging from recent gaming WMs, where faithful dynamics are learned from visual and control signals across richer environments. In this setting, WAMs could further enhance predictive control by grounding action generation in deeper anticipations of world evolution. More importantly, it may enable agents to learn transferable dynamics rather than game-specific action patterns, moving gaming WAMs toward a more fundamental world understanding and ultimately toward \emph{omni-reality adaptability} across the game multiverse.

\paragraph{Synthesis: shared representations versus transferable control.}
Architectural unification makes inputs and outputs compatible; it does not by itself establish shared action semantics. OmniJARVIS couples learned behavior tokens to a policy decoder, with tokenizer ablations evaluated within Minecraft~\citep{omnijarvis}. This supports a representation choice, while cross-game transfer remains a separate test. The same control can have different consequences under a different transition function $T$, so a shared latent space must preserve distinctions relevant to goal progress. To separate motor reuse from strategic reasoning, compare learned behavior tokens with direct low-level action outputs at matched parameter and data budgets, then freeze model weights and harness and test disclosed key remappings, new goal compositions, and unseen mechanics separately. Remapping success supports action grounding; sustained success under changed mechanics provides different evidence about planning and adaptation. For WAMs, ablate imagined rollouts at a fixed inference budget and evaluate on held-out transition rules. Improved prediction is useful for agent generality only insofar as it improves task competence or coverage under the same observation access and evaluation protocol (Section~\ref{sec:generality-comparison}).

\section{Harness: The Nervous System of Generalist Game Player}

\begin{table*}[!t]
\centering
\caption{Summary of representative harness designs in gaming AI}
\label{tab:paper_harness}
\begingroup
\setlength{\tabcolsep}{4pt}
\renewcommand{\arraystretch}{1.08}
\resizebox{\textwidth}{!}{%
\begin{tabular}{lllllll}
\toprule
\textbf{Paper} & \textbf{Perception} & \textbf{Action} & \textbf{Reasoning} & \textbf{Real-time Reactivity} & \textbf{Memory} & \textbf{Adaptive Learning} \\
\midrule
DEPS~\citeyearpar{wang2023describe} & Textual Env. & Semantic & Workflow & -- & Working Only & Self-Reflection \\
AgentVerse~\citeyearpar{chen2024agentverse} & Textual Env. & Semantic & Multi-Agent & -- & Working Only & -- \\
Voyager~\citeyearpar{voyager8} & Textual Env. & Code Control & Workflow & -- & Both & Evolving Skills \\
StarCraft Bench~\citeyearpar{ma2024large} & Textual Env. & Semantic & Prompt ENGR & -- & Working Only & -- \\
Cradle~\citeyearpar{cradle14} & Raw & GUI & Workflow & Pausing & Both & Self-Reflection \\
GAMEBoT~\citeyearpar{gamebot} & Textual Env. & Semantic & Prompt ENGR & -- & Working Only & -- \\
GameSense~\citeyearpar{gamesense} & Visual Scaffold & Code Control & Workflow & Decoupling Reasoning from Action & Both & Self-Reflection \& Evolving Skills \\
VideoGameBench~\citeyearpar{videogamebench4} & Raw & GUI & Workflow & Pausing & Working Only & -- \\
LPLH~\citeyearpar{lplh} & Textual Env. & Semantic & Workflow & -- & Both & Self-Reflection \\
Orak~\citeyearpar{oark5} & Visual Scaffold \& Textual Env. & Semantic & Workflow & Pausing & Both & Self-Reflection \\
GMH~\citeyearpar{gmh} & Visual Scaffold \& Textual Env. & Semantic & Workflow & -- & Working Only & Prompt Opt. \\
VistaWise~\citeyearpar{vistawise} & Visual Scaffold & GUI & Prompt ENGR & Pausing & Both & -- \\
FlashAdventure~\citeyearpar{flashadventure6} & Raw & GUI & Workflow & -- & Both & Self-Reflection \\
TITAN~\citeyearpar{videogametesting} & Textual Env. & Semantic & Workflow & -- & Both & Self-Reflection \\
CWM~\citeyearpar{codeworldmodel} & Textual Env. & Code Control & Workflow & -- & Both & Self-Reflection \\
AgileThinker~\citeyearpar{realtimereasoning} & Textual Env. & Semantic & Workflow & Dual Thread & Working Only & -- \\
PERIL~\citeyearpar{licato2025persona} & Textual Env. & Semantic & Prompt ENGR & -- & Working Only & -- \\
DSGBench~\citeyearpar{dsgbench} & Textual Env. & Semantic & Workflow & Pasuing & Working Only & Self-Reflection \\
Lmgame-Bench~\citeyearpar{lmgbench9} & Textual Env. & Semantic & Prompt ENGR & Pausing & Working Only & Prompt Opt. \& Self-Reflection \\
DeepPHY~\citeyearpar{DBLP:conf/aaai/XuBWKWSZSDZ26} & Visual Scaffold & Semantic & Prompt ENGR & -- & Working Only & Self-Reflection \\
Steve-Evolving~\citeyearpar{xie2026steveevolvingopenworldembodiedselfevolution} & Visual Scaffold & Code Control & Workflow & -- & Both & Evolving-Skills \\
PokeAgent Challenge~\citeyearpar{karten2026pokeagentchallenge} & Textual Env. & Sematic & Multi-Agent & -- & Both & Self-Reflection \\
STAR~\citeyearpar{BeyondScaling} & Textual Env. & Sematic & Prompt ENGR & Batched Command Execution & Working Only & -- \\
NitroGen~\citeyearpar{nitrogen} & Raw & GUI & -- & Pausing & Working Only & -- \\
MineNPC-Task~\citeyearpar{MineNPCTask} & Textual Env. & Code Control & Workflow & -- & Both & Self-Reflection \\
BotzoneBench~\citeyearpar{BotzoneBench} & Textual Env. & Sematic & Prompt ENGR & -- & Working Only & -- \\
GameVerse~\citeyearpar{gameverse69} & Raw & GUI/Sematic & Workflow & -- & Both & Self-Reflection \\
Continual Harness~\citeyearpar{continualharness2026} & Screenshots + text maps & Buttons + code skills & Multi-Agent + refiner & NR & Both & Harness edits + co-train. \\
SPIKE~\citeyearpar{spike2026} & Screen + derived feedback & Primitives (subtype NR) & Strategic / reactive workflow & Event-triggered dual controller & Both & Reflection / memory updates \\
APEX~\citeyearpar{apex2026} & Textual Env. & Semantic text actions & Strategy-DAG exploration & N/A (turn-based) & Both & Reflection / DAG evolution \\
PTCG-Bench~\citeyearpar{ptcgbench2026} & Text (partial state) & Semantic tools / masking & ReAct workflow & N/A (turn-based) & Working / Both & Reflection / prompt / skills \\
OmniGameArena~\citeyearpar{omnigamearena2026} & Raw screenshots & Native control primitives & Reflect / judge / distill & Paused / latency-charged tracks & Both & Skill-prompt refinement \\
Env.-Grounded APO~\citeyearpar{environmentgroundedapo2026} & Text + descriptor & Semantic (BabyAI) & Multi-Agent workflow & N/A (step-based) & Working Only & Reward-gated Prompt Opt. \\
AgenticSTS~\citeyearpar{agenticsts2026} & Text + typed rules & Semantic API & Typed-retrieval workflow & N/A (turn-based) & Both & Summaries / prose skills \\
Gauntlet~\citeyearpar{compiledagency2026} & Descriptions + raw APIs & Code Control (frozen) & Develop / test / freeze & Zero runtime model calls & Controller artifacts & Pre-freeze code evolution \\
WorldMind~\citeyearpar{worldmind2026} & Frames  ->  compact state & NPC action conditioning & Four-layer workflow & NR & Working state & NR \\
GameDirector~\citeyearpar{gamedirector2026} & Visual  ->  explicit state & NPC commands / prompts & Rules / NPC planning & NR & Explicit world state & NR \\
SpireBench~\citeyearpar{spirebench2026} & Text + state diffs & Semantic tools / batches & Planning + reflection & N/A (turn-based) & Both (variants) & Reflection / consolidation \\
AAArena~\citeyearpar{aaarena2026} & Rules / SDKs / replays & Code Control (development) & Test / revise workflow & Standalone policy; timing NR & Development state & Executable-policy evolution \\
Preference Model~\citeyearpar{preferencemodel_sts2026} & Textual Env. & Semantic tools / batches & Workflow + rule lookup & N/A (turn-based) & Transcript + rules & NR \\
Matrix-Game 4.0~\citeyearpar{matrixgame4_2026} & Engine state + proxies & Code / renderer & Coding-agent workflow & Causal rendering & World state/references & NR \\
\bottomrule
\end{tabular}
}
\endgroup
\end{table*}

A game's observed difficulty depends on the harness: exposing hidden state, packaging skills, or pausing execution changes what the foundation model must solve.

Classical cognitive science frameworks, such as the Unified Theories of Cognition (UTC), posit that intelligence is fundamentally a closed-loop system intertwined with essential functions like perception, memory, reasoning, and action \citep{utc92,standardMoM93}. However, although the foundation models demonstrate powerful representation capabilities, they are stateless by design and suffer from inherent limitations, such as memory volatility \citep{bridging94,tang2026vlm}. Consequently, they fail to function as complete intelligent systems capable of navigating dynamic and complex environments independently. To bridge this gap, researchers have engineered various Modular Harnesses designed to decouple and augment the foundational functions required by autonomous agents.

In the current landscape, the most prominent harnessing components can be categorized into six core dimensions: 

\begin{figure*}[t]
  \centering
  \includegraphics[width=1\linewidth]{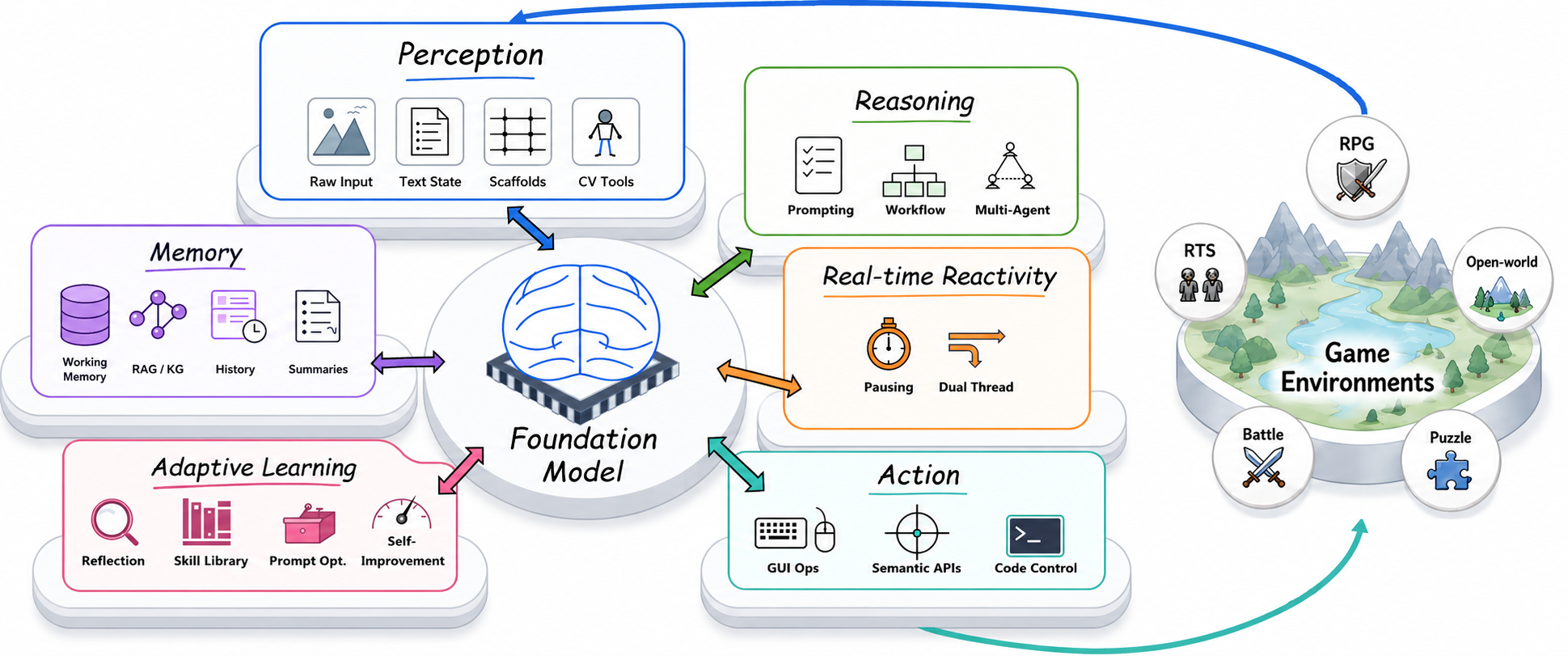}
  \caption{\textbf{Overview of Harness in Game-playing AI.} The harness bridges foundation models and game environments through six interdependent dimensions: Perception parses raw observations into model-readable representations, Action grounds decisions into executable interfaces, and Reasoning synthesizes goals with perceived states into multi-step plans. Real-time Reactivity mitigates inference latency in dynamic environments, Memory persists interaction histories beyond stateless inference, and Adaptive Learning distills trajectories into reusable experiences for continual self-evolution.}
  \label{fig:harness_teaser}
\end{figure*}

\begin{itemize}
    \item \textbf{Perception:} The perception module extracts multi-modal information from raw environmental observations ($\Omega$) and transforms it into structured representations---such as textual prompts---that the underlying reasoning engine (e.g., LLMs) can process.
    \item \textbf{Action:} The action module translates the agent's internal decisions into executable operations. It involves defining the allowable action space ($A$), designing the invocation mechanisms (e.g., APIs, code execution, or GUI operations), and applying these actions to alter the environment state.
    \item \textbf{Reasoning:} Serving as the cognitive core, the reasoning module synthesizes the predefined goals ($G$) with perceptual inputs to estimate the current state ($S$). Instead of executing actions directly, it performs task decomposition, multi-step planning, and selects the optimal action trajectory.
    \item \textbf{Real-time Reactivity:} This mechanism is specifically designed to mitigate the critical gap between the inherent inference latency of foundation models and the temporal constraints of dynamic, real-time environments, ensuring timely and robust responses.
    \item \textbf{Memory:} To overcome the stateless nature of foundation models, the memory module persistently stores and manages key interaction histories, environment knowledge, and procedural experiences. It provides the agent with a coherent context for consistent decision-making.
    \item \textbf{Adaptive Learning:} The adaptive learning module drives the continuous evolution of the agent. By reflecting on successes and failures, it abstracts interaction trajectories into reusable heuristic experiences, enabling the system to self-correct and autonomously adapt to novel or dynamic environments over time.
\end{itemize}

In the following sections, we systematically review the harness designs, practical efficacies, and current limitations corresponding to each of these six functional dimensions.

\subsection{Perception}
In game interactions, vision serves as the most fundamental perceptual channel for humans, enabling them to observe dynamic environments, read textual information, confirm game states, and guide precise action execution. However, this natural process poses a significant challenge for foundation models. Large Language Models (LLMs) are inherently restricted to the text modality; meanwhile, although Vision-Language Models (VLMs) can process visual inputs and comprehend global semantics, they still exhibit a substantial gap compared to human visual adaptation, particularly in fine-grained spatial alignment. Recent studies reveal that when raw game screenshots are used directly as input, VLMs frequently encounter issues such as spatial localization bias, semantic ambiguity, and severe perceptual hallucinations \citep{gameverse69,videogamebench4}. To bridge this "spatial-semantic gap" and alleviate the models' cognitive load, researchers have designed various perception harnesses. These module designs primarily include: \textbf{Textual Environment} and \textbf{Visual Scaffolding}.

\paragraph{\textbf{Textual Environment.}} The textual environment paradigm bypasses the inherent spatial grounding limitations of vision models by converting complex, high-dimensional game states into structured natural language, tabular formats, or symbolic code representations. Recent studies implement this paradigm through distinct, framework-specific designs to optimize cognitive load. For instance, the LPLH framework utilizes Interactive Fiction games, which are natively text-based, thereby entirely circumventing visual perception challenges and allowing the agent to focus purely on semantic understanding and narrative reasoning \citep{lplh}. For visually complex Real-Time Strategy games, DSGBench and Reflection of Episodes leverage the TextStarCraft II environment to parse dense environmental states---such as unit counts, resource levels, and building statuses---into comprehensive textual summaries \citep{ma2024large,xu2026reflectionepisodeslearningplay,dsgbench}. Similarly, The PokeAgentChallenge and Pok\'{e}champ Agent abstract the intricate battle mechanics of Pok\'{e}mon into structured text via the Pok\'{e}mon Showdown simulator \citep{karten2025pokechamp,karten2026pokeagentchallenge}. For spatially structured tasks, frameworks like LmgameBench extract backend data from grid-based games (e.g., Sokoban, 2048, Tetris) to convert visual layouts into textual tables, explicitly listing object coordinates and their attributes \citep{lmgbench9}. For open-world 3D sandbox games, frameworks like Voyager and the Orak benchmark utilize the Mineflayer JavaScript API to entirely bypass raw visual processing \citep{voyager8,oark5,mineflayer}. Taking abstraction a step further, Code World Model translates natural language game rules and historical trajectories into an executable Python-based world model. This approach transforms the game environment into formal code, serving as a reliable simulator to support rigorous algorithmic planning, such as Monte Carlo Tree Search \citep{codeworldmodel}.

\paragraph{\textbf{Visual Scaffolding.}} In scenarios where visual modalities need to be retained, frameworks deploy visual scaffolding or external computer vision (CV) tools to bridge the spatial-semantic gap. To directly enhance the raw visual input, General Modular Harness explicitly overlays grid lines and coordinate labels onto the game images to reduce perception errors \citep{gmh}. Alternatively, Steve-Evolving anchors continuous visual experiences into structured state snapshots (e.g., precise coordinates, inventory summaries, health, and GUI states), pairing them with fine-grained execution diagnoses to facilitate reliable recall and agent evolution \citep{xie2026steveevolvingopenworldembodiedselfevolution}. In privilege-free settings, modules employ external CV models as visual parsers. For example, GameSense operates entirely on real-time game screenshots without API access, deploying a suite of CV tools (e.g., Grounding DINO for object detection and OpenCV for state reading) to parse highly dynamic environments in FPS and action games \citep{gamesense}. Similarly, VistaWise and the Orak benchmark deploy lightweight object detection models (such as YOLOv10 and YOLOv11) to extract bounding boxes and entity positions from raw screenshots \citep{vistawise,oark5}. Crucially, they perform perception abstraction to translate continuous spatial data into discrete semantic concepts (e.g., categorizing the distance between fighters as "very close" or "far").

\subsection{Action}
When playing games, human players seamlessly translate high-level semantic intentions (e.g., "aim and shoot the enemy") into fine-grained motor executions (e.g., moving the mouse to specific coordinates and left-clicking). While current foundation models exhibit robust reasoning and tool-invocation capabilities, they lack native motor skills to directly manipulate digital environments. Therefore, defining an appropriate action space and encapsulating environmental interactions into executable interfaces serves as the critical bridge connecting the model's cognition to the game engine. As action spaces have rapidly evolved from predefined function calls toward more flexible interface-level navigation and code execution, we categorize current Action modules into three progressive paradigms: \textbf{GUI-level operations} via General Computer Control (GCC) \citep{cradle14}, \textbf{Semantic-designed action spaces} through API and skill encapsulation, and \textbf{Programmatic control} via automated code generation.

\begin{table}[!t]
    \centering
    \caption{Example of GUI Action Space}
    \label{tab:gui_actions}
    \resizebox{\textwidth}{!}{
        \begin{tabular}[!]{l l p{10cm}}
            \toprule
            \textbf{Action Type} & \textbf{Description} & \textbf{Parameters} \\
            \midrule
            
            \multicolumn{3}{l}{\textit{\textbf{Mouse Movement}}} \\
            \midrule
            MOVE\_TO & Move to position & \texttt{x}, \texttt{y} (int, req): Target coordinates (window relative) \\
            MOVE\_BY & Move relatively & \texttt{dx}, \texttt{dy} (int, req): Offset; \texttt{duration} (float, opt): Seconds \\
            
            \midrule
            \multicolumn{3}{l}{\textit{\textbf{Mouse Click}}} \\
            \midrule
            CLICK & Click mouse & \texttt{x}, \texttt{y} (opt); \texttt{button} (left/right/middle); \texttt{num\_clicks} \\
            RIGHT\_CLICK & Right click & \texttt{x}, \texttt{y} (int, opt): Click coordinates \\
            DOUBLE\_CLICK & Double click & \texttt{x}, \texttt{y} (int, opt): Click coordinates \\
            
            \midrule
            \multicolumn{3}{l}{\textit{\textbf{Mouse Drag}}} \\
            \midrule
            MOUSE\_DOWN & Press button & \texttt{button} (str, opt): "left", "right", or "middle"; \texttt{duration} (float, opt) \\
            MOUSE\_UP & Release button & \texttt{button} (str, opt): "left", "right", or "middle";  \\
            DRAG\_TO & Drag to target & \texttt{x}, \texttt{y} (int, req): Target coordinates \\
            
            \midrule
            \multicolumn{3}{l}{\textit{\textbf{Mouse Scroll}}} \\
            \midrule
            SCROLL & Scroll wheel & \texttt{dx}, \texttt{dy} (int, req): Scroll amount (+/-); \texttt{duration} (float, opt) \\
            
            \midrule
            \multicolumn{3}{l}{\textit{\textbf{Keyboard Input}}} \\
            \midrule
            TYPING & Type text & \texttt{text} (str, req); \texttt{interval} (float, opt) \\
            PRESS & Press key & \texttt{key} (str, req); \texttt{duration} (float, opt) \\
            KEY\_DOWN & Key down & \texttt{key} (str, req): Key name; \texttt{duration} (float, opt) \\
            KEY\_UP & Key up & \texttt{key} (str, req): Key name;  \\
            HOTKEY & Key combo & \texttt{keys} (list, req): e.g., \texttt{['ctrl', 'c']}; \texttt{duration} (float, opt) \\
            
            \midrule
            \multicolumn{3}{l}{\textit{\textbf{Control Flow}}} \\
            \midrule
            WAIT & Wait time & \texttt{duration} (float, req): Seconds to wait \\
            DONE & Task success & No parameters \\
            FAIL & Task failure & No parameters \\
            \bottomrule
        \end{tabular}
    }
\end{table}

\paragraph{\textbf{GUI-level Operations}} General Computer Control (GCC), a paradigm formalized by the Cradle framework, aims to build foundational agents that master computer tasks via a universal human-style interface---receiving inputs from screens and audio, and outputting native keyboard and mouse actions \citep{cradle14}. A representative design of such a GCC action space used by GameVerse is listed in Table.\ref{tab:gui_actions} \citep{gameverse69}. Frameworks such as Cradle demonstrate that, by utilizing pure GUI actions, agents can navigate complex commercial games like Red Dead Redemption 2 without relying on any built-in APIs \citep{cradle14}. Similarly, benchmarks like FlashAdventure evaluate agents on completing full story arcs in diverse web-based games using mouse and keyboard interfaces \citep{flashadventure6}. While this paradigm offers ultimate generalizability across any software, it exposes a critical limitation of current Vision-Language Models (VLMs) known as the "knowing-doing gap" or "semantic-execution gap" \citep{videogamebench4,gameverse69}. As observed in evaluations, even when an agent correctly reasons about the optimal high-level strategy, it frequently fails to map this semantic intent to the precise action with the right parameters, such as coordinates. This profound disconnect between strategic reasoning and action execution leads to significant performance degradation, particularly in real-time or precision-heavy scenarios. Furthermore, GUI operations like \texttt{drag} and \texttt{hold\_key} pose significant spatial and temporal challenges, as the drag trajectory and holding time require spatial and temporal intelligence, or even a combination of both. VideoGameBench noted that current foundation models perform poorly in the corresponding game environments \citep{videogamebench4}.

\paragraph{\textbf{Semantic-designed Actions}} Semantic actions are mostly pre-defined according to the common operational logic within the environment, utilizing conceptual or discrete parameters (e.g., target IDs or semantic locations) instead of precise pixel coordinates and physical distances \citep{gameverse69,videogametesting,gmh,gamebot}. For instance, in open-world MMORPGs, the TITAN framework abstracts the overwhelmingly vast continuous action space into high-level templates such as \texttt{Moveto(Location)} and \texttt{Attack(target)} \citep{videogametesting}. Similarly, the Orak benchmark abstracts the exact frame-perfect joystick combinations required in fighting games (e.g., Street Fighter III) into interpretable, discrete semantic commands like "Move Closer" or character-specific skills like "Fireball" \citep{oark5}. To systematically implement these semantic invocations, recent advancements leverage standardized tool-calling interfaces. In the Pok\'{e}Agent Challenge and Orak, the semantic actions are wrapped into independent Model Context Protocol (MCP) server to achieve plug-and-play functionality \citep{karten2026pokeagentchallenge,oark5}.
\paragraph{\textbf{Programmatic Control}} Programmatic control decouples high-level reasoning from low-level execution by utilizing executable code as the action space. Unlike semantic APIs that rely on predefined functions, this paradigm allows foundation models to synthesize their own execution logic. For instance, to mitigate the inference latency associated with direct key-mouse control in fast-paced environments, the GameSense framework utilizes VLMs to generate specialized Python scripts known as "Game Sense Modules" (GSMs). These scripts encapsulate real-time interactive logic and integrate external vision tools, running locally to manage tasks such as combat \citep{gamesense}. In open-world settings, Voyager leverages the Mineflayer API to generate JavaScript control primitives, verifying and storing these code snippets in an extensible skill library for future reuse \citep{voyager8}. Expanding this approach to environment simulation, Code World Models (CWM) prompts LLMs to translate natural language game rules and trajectories into an executable Python world model. This synthesized code includes functions for state transitions, legal move enumeration, and termination checks, functioning as a simulation engine to facilitate planning algorithms like Monte Carlo Tree Search (MCTS) \citep{codeworldmodel}.

\subsection{Reasoning} While perception modules extract environmental states and action modules execute commands, reasoning and planning modules serve as the central cognitive engine that bridges the two. Foundation models typically excel at rapid, intuitive responses but often struggle with the deliberate, multi-step logic required for complex strategic tasks. To elicit higher-order reasoning, recent research leverages various inference-time cognitive harnesses. Based on the structural design of the reasoning process, we categorize current Reasoning modules into three progressive paradigms: \textbf{Prompt Engineering}, \textbf{Workflow Orchestration}, and \textbf{Multi-Agent Systems}.

\paragraph{\textbf{Prompt Engineering}} At the most fundamental level, reasoning capabilities are elicited through carefully designed prompt formulations. Techniques such as Chain-of-Thought (CoT) prompting explicitly guide the model to generate intermediate reasoning traces before executing an action. For instance, the GAMEBoT framework leverages domain-specific CoT prompts to guide LLMs in addressing predefined modular subproblems prior to action selection \citep{gamebot}. Similarly, DEPS provides few-shot demonstrations to prompt the model into generating self-explanations for complex open-world tasks \citep{wang2023describe}, while environments like \texttt{lmgame-Bench} are specifically designed to support modern reasoning paradigms by allowing models to be evaluated with or without long chain-of-thought (long-CoT) reasoning \citep{lmgbench9}. Furthermore, to steer strategic reasoning in adversarial or cooperative settings, techniques like Persona Prompting inject specific strategic profiles into the model to reshape its decision-making process, as demonstrated in strategic board games like PERIL \citep{licato2025persona}.

\paragraph{\textbf{Workflow Orchestration}} To tackle long-horizon tasks and dynamic environments, raw reasoning is structured into systematic workflows. A foundational paradigm is ReAct, which interleaves reasoning traces with executable actions, allowing the agent to continuously ground its thoughts in environmental observations rather than relying purely on internal logic \citep{yao2022react}. For complex goals, explicit task decomposition is employed to break down macro-objectives into manageable sub-goals
. For instance, DEPS incorporates an interactive planning process with a trainable goal selector to rank parallel candidate sub-goals \citep{wang2023describe}, while Voyager utilizes an iterative prompting mechanism that incorporates environment feedback, execution errors, and self-verification to build and refine an executable skill library \citep{voyager8}. Similarly, TITAN employs a plan-execute-memorize-diagnose-replan workflow that tracks action histories and state coverage to diagnose execution stalls and dynamically adjust testing strategies in complex MMORPGs \citep{videogametesting}. Taking structural reasoning a step further, frameworks incorporate traditional search algorithms directly into the LLM's reasoning loop, such as Pok\'{e}Champ, which utilizes LLMs for heuristic position evaluation to guide depth-limited minimax search \citep{karten2025pokechamp}.

\paragraph{\textbf{Multi-Agent Systems}} Shifting the focus from individual intelligence to collective rationality, reasoning is distributed across multi-agent ecosystems. In complex environments, architectures often adopt an orchestrator with specialized sub-agents. The Pok\'{e}Agent Challenge implements a central orchestrator that maintains high-level route plans while dynamically dispatching specialized sub-agents for battle strategy, puzzle-solving, and self-reflection \citep{karten2026pokeagentchallenge}. Beyond internal cognitive division, AgentVerse deploys multiple independent agents in Minecraft \citep{chen2024agentverse}. In this shared open-ended environment, this multi-player collaboration enables agents to efficiently tackle multi-step tasks while exhibiting emergent social behaviors, such as volunteering unallocated time and resources to accelerate team progress.

\subsection{Real-time Reactivity} The inherent inference and execution latency of foundation models severely restricts their performance in real-time, dynamic environments. To mitigate this latency in interactive games, environment pausing is conventionally used to skip the reasoning time. For instance, Cradle, Orak, and VideoGameBench Lite pause the game during reasoning \citep{cradle14,oark5,videogamebench4}, while NitroGen intercepts the system clock of the game engine to freeze simulation time
\citep{nitrogen}. To bypass the reliance on environment pausing, recent frameworks focus on decoupling reasoning from action execution. In GameSense, by shifting the VLM's role from a direct controller to a module developer, the VLM generates task-specific executable code, termed Game Sense Modules (GSMs), which autonomously take over high-frequency real-time operations \citep{gamesense}. Besides, AgileThinker introduces a dual-thread architecture consisting of a planning thread for multi-step reasoning and a parallel reactive thread for outputting actions in real-time \citep{realtimereasoning}. Furthermore, to achieve real-time action generation, Lumine utilizes hybrid thinking to skip unnecessary reasoning and employs action chunking for high-frequency control, alongside latency optimizations including StreamingLLM, tensor parallelism, W8A8 quantization, and speculative decoding \citep{lumine,xiao2024efficient,speculativedecoding,xiao2023smoothquant}.

\subsection{Memory} Since foundation models are inherently stateless, memory modules are essential to store and maintain information for multi-step interactions of agents with dynamic environments \citep{sumers2024cognitive}. Following the Cognitive Architectures for Language Agents (CoALA) framework, we systematically organize memory modules into \textbf{Working Memory} and \textbf{Long-Term Memory}. In Table~\ref{tab:paper_harness}, \emph{Both} denotes these two memory types.

\paragraph{\textbf{Working Memory}} Working memory maintains active and readily available information, persists across foundation model calls, and enables the models to be stateful agents \citep{sumers2024cognitive}. A common approach for working memory is keeping a fixed steps of recent states and actions using a sliding window, such as the memory modules in lmgame-Bench and GMH \citep{lmgbench9,gmh}. Furthermore, VistaWise designs the memory stack based on the concept of Last-In-First-Out (LIFO), allowing the agent to recall decisions from the most recent to earlier ones with controllable recall steps \citep{vistawise}. Another form is history summarization, which maintains a fixed-length or compacted summary from history to avoid context 
overflow, as utilized in GameVerse and the Pok\'{e}Agent Challenge \citep{gameverse69,karten2026pokeagentchallenge}.

\paragraph{\textbf{Long-Term Memory}} Following the CoALA architecture, long-term memory can be systematically categorized into semantic, episodic, and procedural memory \citep{sumers2024cognitive}. Semantic memory provides the agent with factual background knowledge about the environment. For instance, LPLH and VistaWise utilize knowledge graphs to store world knowledge, entities, and connections, dynamically evolving them with the agent's exploration \citep{lplh,vistawise}. Similarly, to mitigate the lack of domain-specific knowledge, the Pok\'{e}Agent Challenge injects usage statistics and competitive strategies directly from Smogon, an online Pok\'{e}mon community \citep{karten2026pokeagentchallenge}. Episodic memory records the agent's past experiences and historical trajectories. To effectively maintain and retrieve these experiences, GameVerse employs a vector database (ChromaDB) to store the multimodal experiences that the agent decides to save \citep{gameverse69}. Finally, procedural memory stores the executable skills and action patterns. For example, Voyager maintains an ever-growing skill library of executable code to retain useful skills for future interactions \citep{voyager8}, while GameSense constructs a procedural memory database to store and retrieve historical action implementations and their corresponding execution codes \citep{gamesense}.

\subsection{Adaptive Learning} In a training-free paradigm, agents must adapt to dynamic environments entirely through their external harness rather than parameter updates. We categorize these adaptive learning mechanisms into three hierarchical levels: \textbf{Self-Reflection} for text-based experiential learning, \textbf{Evolving Skills} for executable action-level evolution, and \textbf{Prompt Optimization} for system-level instruction refinement. Continual Harness also explores model-weight co-training as a separate extension to harness editing~\citep{continualharness2026}.

\paragraph{Self-Reflection} Reflection utilizes the in-context learning capacity of foundational models, prompting the agent to analyze historical information and generate reflective text, which is then injected into the next-turn prompt to guide future actions. For state-level diagnosis, Orak compares the current state, previous state, and executed actions to generate reflections on state transitions \citep{oark5}. Similarly, TITAN generates diagnostic reflections by evaluating the abstracted state, current screenshots, action history, and task objectives \citep{videogametesting}. To further augment the reflection process, recent frameworks incorporate expert experiences as reference baselines. Reflection of Episodes (ROE) combines pre-defined expert experiences with self-experiences based on keyframe selection, generating updated self-experiences through post-episode reflection \citep{xu2026reflectionepisodeslearningplay}. Furthermore, GameVerse employs Vision-Language Models (VLMs) to simultaneously analyze the agent's failure videos and human expert tutorials, contrasting the two visual trajectories to generate concentrated experience prompts for policy refinement \citep{gameverse69}.

\paragraph{Evolving Skills} To better adapt to dynamic game environments, agents can continuously learn and memorize new skills. For example, in Voyager, once a generated piece of code is self-verified to achieve its target goal, it is treated as a new skill and stored in an ever-growing skill library for future reuse \citep{voyager8}. Furthermore, Steve-Evolving advances this concept by structurally anchoring interactive experiences with fine-grained diagnostic signals, including state differences and enumerated failure causes \citep{xie2026steveevolvingopenworldembodiedselfevolution}. Based on these anchored experiences, successful actions are distilled into reusable skills with explicit preconditions, while failed trajectories are distilled into executable guardrails to strictly prevent the agent from repeating similar mistakes.

\paragraph{Prompt Optimization} Agents' performance heavily depends on the quality of prompts. To address this, the General Modular Harness (GMH) introduces a two-stage prompt optimization pipeline leveraging objective reward signals from the environment \citep{gmh}. In the first stage, GMH applies empirical prompt engineering to construct a baseline template that integrates trajectory histories (past states, actions, and rewards) and reflective summaries. In the second stage, it standardizes the optimization process using the DSPy framework and the SIMBA algorithm. By treating the game's cumulative reward as the objective function, this process iteratively refines the baseline prompt templates.

\paragraph{Synthesis: attributing competence to the deployed system.}
State APIs change the observation channel $O$, and semantic skills change the effective action space $A$; gains from either concern the deployed policy $\Pi_{\theta,H}$, not the backbone alone. In lmgame-Bench's o4-mini ablation, the reported mean Sokoban score is unchanged with memory alone and higher with visual scaffolding~\citep{lmgbench9}; this supports separating perception assistance from base-model reasoning rather than treating their combined score as a model property. Section~\ref{sec:generality-comparison} examines such controlled evidence. A fair comparison swaps models within one fixed harness and separately ablates perception, memory, and planning within one model, keeping goals, opponents, seeds, access privileges, and resource budgets, including auxiliary-model calls, matched. Memory reset and update rules must also be declared, so improvement over repeated attempts can be distinguished from competence on first contact. Timing requires a separate control: NitroGen freezes simulation during inference~\citep{nitrogen}, which removes the cost of delayed decisions under real-time dynamics. Reporting both matched-step performance and performance under wall-clock deadlines would distinguish deliberation benefits from executable real-time control. These comparisons let threshold coverage describe a specified model--harness system and help attribute its failures, instead of conflating assistance, adaptation, and native capability.

\section{Benchmarks: The Diverse Arena for Game-Playing Agents}
\label{sec:benchmarks}

A benchmark determines which generality claim can be tested: competence on known tasks, coverage across a game list, or transfer under a specified structural change. Holding the deployed policy $\Pi_{\theta,H}$ fixed is necessary for attributing differences to games rather than to a new controller. Conversely, paired model or harness changes within the same protocol can identify which pillar limits performance; Section~\ref{sec:generality-comparison} gives concrete comparisons.

\begin{figure*}[t]
  \centering
  \includegraphics[width=0.9\linewidth]{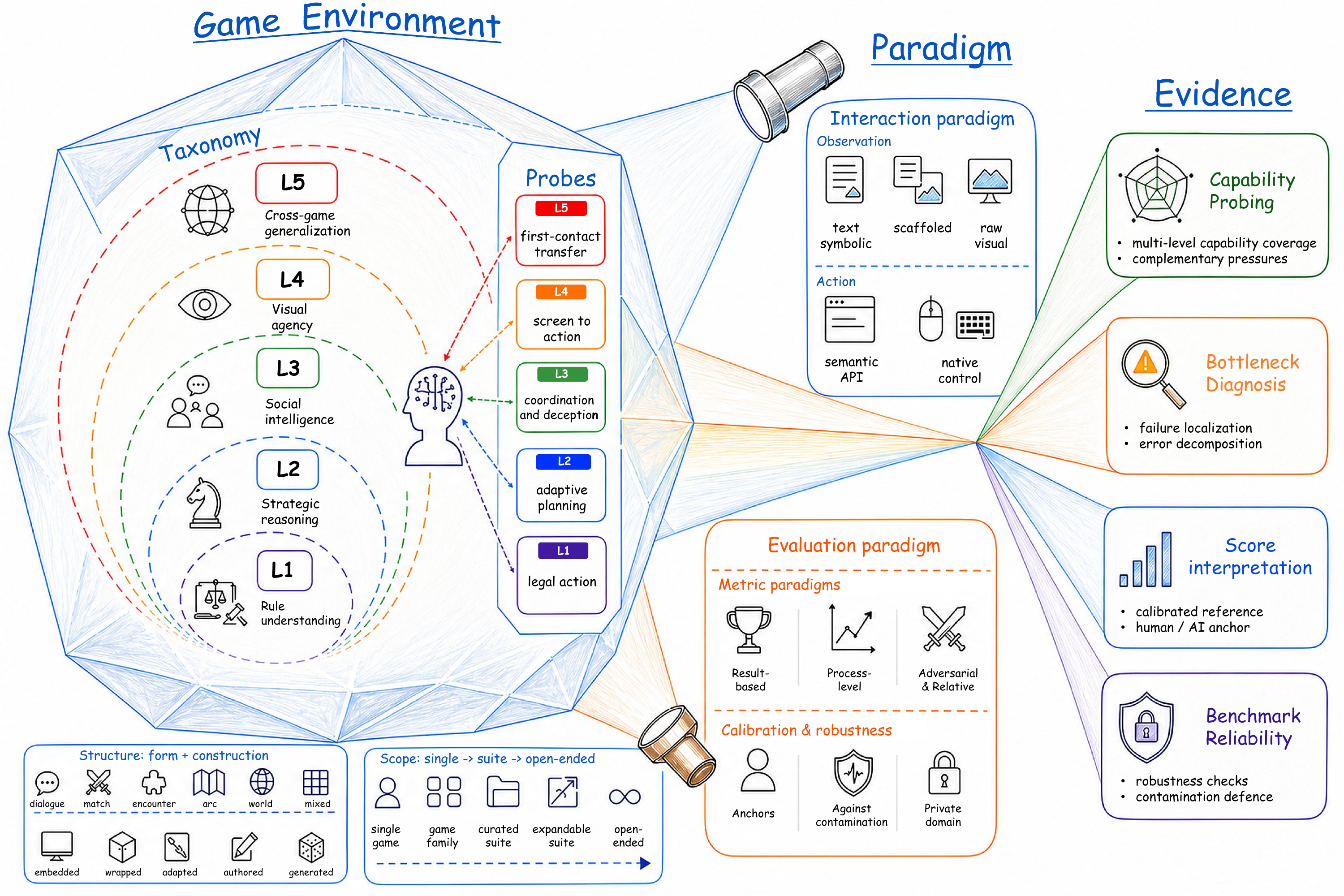}
  \caption{\textbf{Overview of conceptual framework for game benchmarks}. We frame game benchmarks through three connected dimensions: five-level environment organization \rev{(ordered by the minimum model class each interface admits)}, capability probing over agents, and the interaction and evaluation paradigms that shape gameplay into benchmark evidence.}
  \label{fig:benchmark_teaser}
\end{figure*}

The rapid advancement of large language and vision-language models has brought a fundamental evaluation challenge to the forefront: how should we assess systems that are increasingly studied not merely as answer generators, but as agents that must act over time? While many widely used benchmarks remain valuable for testing bounded knowledge, logical reasoning, or one-shot multimodal understanding, they provide only partial evidence of the intelligence required for continuous interaction. True agentic competence emerges in closed-loop settings, where a system must track an evolving state, make decisions under uncertainty, adapt to feedback, and recover from mistakes while pursuing a goal across multiple steps. The central evaluation gap, therefore, is not simply that existing tests are too easy or too narrow, but that many of them abstract away the dynamic structure in which intelligent behavior must actually unfold.

Games provide a compelling substrate for this kind of evaluation because they preserve that dynamic structure in forms natively designed for human play. Rather than mere collections of themed tasks, games are carefully engineered challenge systems characterized by explicit rules, goals, feedback loops, state progression, and consequences that accumulate over time. These properties make games natural probes of human-relevant capability. They effectively expose a broad range of reasoning demands, including rule understanding, planning, memory, uncertainty handling, coordination, and learning from interaction, while also supporting visually grounded agent tasks in which success depends on sustained perception, action, and progress toward an objective \citep{SmartPlay,GTBench,Balrog}. In this sense, games are valuable not because they are simply ``harder'' than static benchmarks, but because they retain the interactive, goal-directed, and human-oriented structure that static benchmarks frequently eliminate.

This perspective also changes how we should understand game diversity. In the context of this survey, the value of games does not lie in assembling a loose catalog of genres, but in viewing the multiverse of human games as a structured space of capability coverage. Different game structures impose different demands on reasoning, memory, coordination, adaptation, and long-horizon control. Thus, diversity matters not as surface-level variation, but as a principled mechanism for applying complementary cognitive and operational pressures on AI agents. This is precisely why games are unusually promising as benchmarks: they span a broad, continually expanding, and difficult-to-saturate design space. As highlighted by AI GameStore~\citep{aigamescore57}, the point is not to identify a small set of canonical titles to solve once and for all, but to treat the multiverse of games as an extensible evaluation substrate for increasingly general, human-relevant intelligence.

However, a game environment does not automatically constitute a benchmark. It becomes one only when a protocol specifies how gameplay is presented, what qualifies as progress, and how behavior is converted into evidence. This section develops that argument through three dimensions, followed by a quantitative comparison:
\begin{itemize}
    \item \textbf{Taxonomy} examines the types of game benchmarks the field has developed, organizing them into an evolutionary design space that spans rule-bounded settings, broader reasoning abilities, and increasingly open-ended environments.
    \item \textbf{Purpose} investigates what these environments are designed to measure, mapping different game structures to their motivating capability targets---ranging from rule grounding and strategic reasoning to social intelligence, visual agency, and cross-game generalization.
    \item \textbf{Paradigm} explores how gameplay is operationalized into benchmark evidence, analyzing the interaction interfaces and evaluation metrics that determine what a benchmark score truly represents.
\end{itemize}
  
Together, these dimensions move from benchmark structure, to capability target, to measurement logic; Section~\ref{sec:generality-comparison} then applies that logic to reported scores. This framing provides a comprehensive understanding of how existing game benchmarks are organized and how the field has evolved. It also helps clarify how benchmark design can more effectively reveal model capabilities and limitations, thereby informing future model and benchmark development.

\subsection{Taxonomy: The Evolutionary Levels of Game Environments}

\begin{table*}[!t]
\centering
\caption{Taxonomy of representative game benchmarks}
\label{tab:benchmark_taxonomy}
\begingroup
\setlength{\tabcolsep}{4pt}
\renewcommand{\arraystretch}{1.08}
\resizebox{\textwidth}{!}{%
\begin{tabular}{lllllllll}
\toprule
\textbf{Paper} & \textbf{Date} & \textbf{Level} & \textbf{Form} & \textbf{Build} & \textbf{Scope} & \textbf{Input} & \textbf{Action} & \textbf{Game Set} \\
\midrule
Jericho~\citeyearpar{InteractiveFictionGames} & 2019/09 & L2 Strategy & Arc & Wrapped & Curated & Text & Semantic & 56 IF games \\
NLE~\citeyearpar{NetHackLearningEnvironment} & 2020/06 & L2 Strategy & World & Wrapped & Single & Text & Semantic/native & NetHack \\
Crafter~\citeyearpar{Crafter} & 2021/09 & L4 Visual & World & Authored & Single & Visual & Native & 22 achievements \\
CICERO~\citeyearpar{cicero} & 2022/11 & L3 Social & Dialogue & Embedded & Single & Text & Semantic & Diplomacy \\
clembench~\citeyearpar{Clembench} & 2023/05 & L3 Social & Dialogue & Authored & Curated & Text & Semantic & 7 dialogue datasets \\
SmartPlay~\citeyearpar{SmartPlay} & 2023/10 & L1 Rule & Mixed & Wrapped & Curated & Text & Semantic & 6 games \\
AvalonBench~\citeyearpar{AvalonBench} & 2023/10 & L3 Social & Dialogue & Adapted & Single & Text & Semantic & Avalon \\
LLM StarCraft II~\citeyearpar{ma2024large} & 2023/12 & L2 Strategy & Match & Wrapped & Single & Text & Semantic & StarCraft II \\
CivRealm~\citeyearpar{civilbench61} & 2024/01 & L2 Strategy & World & Wrapped & Open-ended & Text & Semantic & Civilization \\
GTBench~\citeyearpar{GTBench} & 2024/02 & L1 Rule & Match & Adapted & Curated & Text & Semantic & 10 game-theory tasks \\
GAMABench~\citeyearpar{GAMABench} & 2024/05 & L2 Strategy & Match & Adapted & Curated & Text & Semantic & 8 scenarios \\
GameBench~\citeyearpar{GameBench} & 2024/06 & L2 Strategy & Match & Wrapped & Curated & Text & Semantic & 9 games \\
Werewolf Arena~\citeyearpar{WerewolfArena} & 2024/07 & L3 Social & Dialogue & Adapted & Single & Text & Semantic & Werewolf \\
BALROG~\citeyearpar{Balrog} & 2024/11 & L4 Visual & Mixed & Wrapped & Curated & Mixed & Semantic & 6 env. families \\
TeamCraft~\citeyearpar{long2024teamcraft} & 2024/12 & L3 Social & World & Adapted & Single & Mixed & Semantic & Minecraft \\
GameArena~\citeyearpar{gamearena27} & 2024/12 & L3 Social & Dialogue & Adapted & Curated & Text & Semantic & 3 live games \\
GAMEBoT~\citeyearpar{gamebot} & 2024/12 & L2 Strategy & Match & Wrapped & Curated & Text & Semantic & 8 games \\
PokerBench~\citeyearpar{PokerBench} & 2025/01 & L2 Strategy & Puzzle & Adapted & Single & Text & Semantic & 11K poker spots \\
Collab-Overcooked~\citeyearpar{CollabOvercooked} & 2025/02 & L3 Social & Encounter & Adapted & Single & Text & Semantic & 30 tasks \\
DSGBench~\citeyearpar{dsgbench} & 2025/03 & L2 Strategy & Mixed & Wrapped & Curated & Text & Semantic & 6 games \\
LVLM Game Players~\citeyearpar{LVLM10} & 2025/03 & L4 Visual & Mixed & Adapted & Curated & Mixed & Semantic & 6 games \\
LLM-Coordination~\citeyearpar{LLMCoordination} & 2025/04 & L3 Social & Match & Adapted & Curated & Text & Semantic & 4 games \\
TextArena~\citeyearpar{TextArena} & 2025/04 & L5 General & Mixed & Authored & Expandable & Text & Semantic & 57+ games \\
KORGym~\citeyearpar{KORGym} & 2025/05 & L2 Strategy & Mixed & Authored & Curated & Mixed & Semantic & 51 games \\
LMGAME-BENCH~\citeyearpar{lmgbench9} & 2025/05 & L4 Visual & Mixed & Wrapped & Curated & Mixed & Semantic & 6 games \\
VideoGameBench~\citeyearpar{videogamebench4} & 2025/05 & L4 Visual & Mixed & Embedded & Curated & Visual & Native & 23 games \\
MCU~\citeyearpar{MCU} & 2025/05 & L5 General & World & Generated & Open-ended & Visual & Native & 3452 tasks \\
Orak~\citeyearpar{oark5} & 2025/06 & L5 General & Mixed & Wrapped & Curated & Mixed & Semantic & 12 games \\
TextAtari~\citeyearpar{TextAtari} & 2025/06 & L5 General & Encounter & Adapted & Curated & Text & Semantic & 23 Atari games \\
TextQuests~\citeyearpar{TextQuests} & 2025/07 & L5 General & Arc & Wrapped & Curated & Text & Semantic & 25 IF games \\
StarDojo~\citeyearpar{stardojo11} & 2025/07 & L5 General & World & Adapted & Open-ended & Mixed & Semantic & 1000 tasks \\
GVGAI-LLM~\citeyearpar{GVGAILLM} & 2025/08 & L5 General & Mixed & Generated & Expandable & Text & Semantic & 118 games \\
FlashAdventure~\citeyearpar{flashadventure6} & 2025/09 & L4 Visual & Arc & Embedded & Family & Visual & Native & 34 games \\
StarBench~\citeyearpar{StarBench} & 2025/10 & L4 Visual & Encounter & Embedded & Single & Visual & Mixed & RPG battles \\
LLM-Hanabi~\citeyearpar{LLMHanabi} & 2025/10 & L3 Social & Match & Adapted & Single & Text & Semantic & Hanabi \\
PuzzlePlex~\citeyearpar{PuzzlePlex} & 2025/10 & L2 Strategy & Puzzle & Authored & Curated & Mixed & Semantic & 15 puzzle types \\
WOLF~\citeyearpar{Wolf} & 2025/12 & L3 Social & Dialogue & Adapted & Single & Text & Semantic & Werewolf \\
BotzoneBench~\citeyearpar{BotzoneBench} & 2026/01 & L1 Rule & Match & Wrapped & Curated & Text & Semantic & 8 games \\
Strategic Hanabi~\citeyearpar{StrategicHanabi} & 2026/01 & L3 Social & Match & Adapted & Single & Text & Semantic & Hanabi \\
EMemBench~\citeyearpar{EMemBench} & 2026/01 & L4 Visual & Puzzle & Generated & Curated & Mixed & Mixed & 16 games \\
MineNPC-Task~\citeyearpar{MineNPCTask} & 2026/01 & L2 Strategy & World & Adapted & Single & Text & Semantic & Minecraft \\
AI GAMESTORE~\citeyearpar{aigamescore57} & 2026/02 & L5 General & Mixed & Generated & Expandable & Mixed & Native & 100 games \\
Beyond Scaling~\citeyearpar{BeyondScaling} & 2026/03 & L2 Strategy & Match & Authored & Single & Text & Semantic & Zero-sum game \\
GameVerse~\citeyearpar{gameverse69} & 2026/03 & L5 General & Mixed & Wrapped & Curated & Mixed & Mixed & 15 games \\
ARC-AGI-3~\citeyearpar{ARCAGI3} & 2026/03 & L5 General & Puzzle & Authored & Curated & Visual & Mixed & 135 envs. \\
GameplayQA~\citeyearpar{GameplayQA} & 2026/03 & L4 Visual & Puzzle & Adapted & Curated & Visual & Semantic & 9 games \\
GameWorld~\citeyearpar{GameWorld} & 2026/04 & L4 Visual & Mixed & Wrapped & Curated & Visual & Mixed & 34 games \\
PokeGym~\citeyearpar{PokeGym} & 2026/04 & L4 Visual & World & Embedded & Single & Visual & Mixed & 30 tasks \\
Agent Island~\citeyearpar{agentisland2026} & 2026/05 & L3 Social & Dialogue & Authored & Single & Text & Semantic & Survivor-inspired social game \\
PTCG-Bench~\citeyearpar{ptcgbench2026} & 2026/05 & L2 Strategy & Match & Adapted & Single & Text & Semantic & Pokémon TCG: 5 decks \\
MINDGAMES~\citeyearpar{mindgames2026} & 2026/05 & L3 Social & Mixed & Wrapped & Curated & Text & Semantic & 4 strategic/social games \\
AgentOdyssey~\citeyearpar{agentodyssey2026} & 2026/05 & L5 General & World & Generated & Expandable & Text & Semantic & 6 generated text games \\
OmniGameArena~\citeyearpar{omnigamearena2026} & 2026/06 & L4 Visual & Mixed & Authored & Curated & Visual & Native & 12 games: 7/3/2 solo/PvP/coop \\
AgenticSTS~\citeyearpar{agenticsts2026} & 2026/07 & L2 Strategy & Arc & Wrapped & Single & Text & Semantic & Slay the Spire 2 \\
Gauntlet~\citeyearpar{compiledagency2026} & 2026/07 & N/A & Mixed & Authored / wrapped & Curated & Text / code / feedback & Program  ->  game API & Roguelike, SC2, Freeciv, others \\
GameHorizon~\citeyearpar{gamehorizon2026} & 2026/09 & L4 Visual & World & Wrapped & Single & Visual & Native & Minecraft: 20 tasks/62 subtasks \\
SpireBench~\citeyearpar{spirebench2026} & 2026/10 & L2 Strategy & Arc & Wrapped & Single & Text & Semantic & StS1: A20 Ironclad \\
AAArena~\citeyearpar{aaarena2026} & 2026/10 & N/A & Match & Wrapped & Curated & Text / source / replays & Program  ->  game API & 12 adversarial games \\
Video-Arena~\citeyearpar{videoarena2026} & 2026/10 & N/A & Video QA & Adapted & Curated & Visual & Answers / analysis tools & 9 games; 180 cases \\
Preference Model~\citeyearpar{preferencemodel_sts2026} & 2026/10 & L2 Strategy & Arc & Wrapped & Single & Text & Semantic & StS1: 5 scenarios \\
SpireClimb~\citeyearpar{spireclimb2026} & 2026/10 & L4 Visual & Arc & Wrapped & Family & Visual & Native & Slay the Spire 1/2 \\
\bottomrule
\end{tabular}
}
\endgroup
\end{table*}

While conventional game genres describe games from the perspective of human play, a benchmark taxonomy must describe what an environment preserves, abstracts, and makes measurable for an AI agent. The same genre label can hide very different evaluation contracts, and very different games can impose similar benchmark pressures once they are converted into text states, semantic actions, GUI control, or generated task instances. We therefore treat game benchmarks as a multidimensional design space rather than as a genre catalogue. Its primary axis is an evolutionary spine, which situates benchmarks by their dominant ambition \rev{and the minimum model class their interface admits}; three secondary axes then specify how each benchmark instantiates gameplay as an evaluable object: Structure, Scope, and Modality.

The primary axis of this design space is the evolutionary spine, which tracks the field's expanding benchmark ambition. Level 1 corresponds to early benchmarks built around formal, rule-grounded interaction, evaluating whether an agent can understand rules, track state, and select legal or useful actions in formalized interfaces \citep[e.g.,][]{SmartPlay,GTBench}. Level 2 reflects a shift toward broader strategic decision-making in evolving environments, pushing the ambition from basic rule compliance to dynamic planning and interactive reasoning \citep[e.g.,][]{PokerBench,dsgbench,BeyondScaling}. Level 3 marks the integration of multi-agent social reasoning, where the evaluation target becomes the ability to cooperate, negotiate, or deceive within language-mediated arenas \citep[e.g.,][]{WerewolfArena,Wolf}. Level 4 represents a leap toward ecological validity, shifting the environmental form to preserve more of the visual and interface burdens of human play while testing whether models can operate as grounded visual agents \citep[e.g.,][]{Balrog,StarBench}. Finally, Level 5 extends the frontier from competence in a single environment to adaptability across broader game spaces, generated task distributions, or first-contact settings \citep[e.g.,][]{gameverse69,aigamescore57,ARCAGI3}. \rev{The ordering follows a historical trajectory}: as models evolve, the game environments used to benchmark them tend to abstract away fewer of the native complexities of human play, or make the remaining abstractions more explicit.

\rev{Concretely, levels are ordered by the minimum model class, in the sense of Section 4, that an environment's interface admits. Levels 1--3 deliver rules, states, histories, and social exchange as symbols and are addressable by a text-only LLM; Level 4 delivers a rendered screen and expects coordinate or key events, so VLMs and VLAs are the minimum viable class; Level 5 lies beyond what any single fixed policy has yet achieved, and we conjecture that first-contact competence on unseen dynamics will require agents that carry an internal predictive model of $T$ (a WAM), although current attempts at this frontier are VLA-based. Adjacent levels probe complementary capabilities: Level 3 targets social intelligence and Level 4 visual agency, which differ in the observation and action channel they require. Where a benchmark exerts pressure at several levels, we assign it by its most demanding requirement and record the rest through the Structure, Scope, and Modality axes.}

Table~\ref{tab:benchmark_taxonomy} leaves levels unassigned for offline analysis in Video-Arena and policy development in AAArena and Gauntlet~\citep{videoarena2026,aaarena2026,compiledagency2026}. The GameHorizon row covers online Minecraft~\citep{gamehorizon2026}.

Structure specifies how gameplay becomes an evaluable unit through two fields: Form and Construction. Form identifies the playable unit on which progress, completion, and failure are defined. A Match has bounded contests and clear outcomes, as in formal and strategic game suites \citep[e.g.,][]{GTBench,GameBench}. A Puzzle isolates a constrained solving instance or question-like interactive problem \citep[e.g.,][]{PuzzlePlex,RuleOracles,ARCAGI3}. Dialogue benchmarks treat language interaction itself as the game loop, where debate, negotiation, reference, or persuasion drives the episode forward \citep[e.g.,][]{WerewolfArena,Clembench}. Encounter benchmarks focus on bounded tactical segments, while Arc benchmarks evaluate longer story or quest progressions \citep[e.g.,][]{StarBench,flashadventure6}. World benchmarks use persistent environments in which exploration, resources, tasks, or achievement graphs define progress \citep[e.g.,][]{MCU,civilbench61,long2024teamcraft}. Mixed applies when a suite intentionally spans several such playable units rather than fitting one dominant form \citep[e.g.,][]{SmartPlay,Balrog,GameWorld,oark5}.

Construction records the benchmark's relationship to the original game substrate. Embedded benchmarks preserve the native client or live play surface as much as possible, increasing ecological fidelity while also increasing execution noise and evaluation cost \citep[e.g.,][]{StarBench,flashadventure6}. Wrapped benchmarks reuse existing games or environments through a shared harness, API, or sandbox, improving instrumentation and comparability while abstracting parts of the native interface \citep[e.g.,][]{SmartPlay,Balrog,GameWorld,oark5}. Adapted benchmarks convert existing games, rules, scenes, or logs into targeted evaluation tasks, such as game-theoretic settings, social-deduction arenas, or grounded rule questions \citep[e.g.,][]{GTBench,WerewolfArena,RuleOracles}. Authored benchmarks create game-like environments specifically for measurement, while Generated benchmarks make new tasks, rules, levels, or game instances part of the benchmark's scaling strategy \citep[e.g.,][]{BeyondScaling,PuzzlePlex,aigamescore57,MCU,GVGAILLM}. This distinction is central because ecological validity, repeatability, contamination risk, and measurement cost are all shaped by how far the benchmark artifact sits from the original game.

A separate axis is Benchmark Scope, which describes how a benchmark packages diversity to evaluate distinct operational scales. \emph{Single game} benchmarks support deep diagnosis within one environment, often with clearer metrics and richer failure analysis \citep[e.g.,][]{WerewolfArena,StarBench,BeyondScaling}. \emph{Game family} benchmarks expand within a relatively coherent family of related environments, maintaining recognizable interfaces and evaluation logic \citep[e.g.,][]{flashadventure6}. \emph{Curated suites} intentionally assemble a fixed set of diverse games to evaluate general competence and cover multiple capability slices under a unified protocol \citep[e.g.,][]{SmartPlay,GameBench,dsgbench,Balrog,oark5}. \emph{Expandable suites} treat continuous growth as a core design principle, shifting the generalization focus from mastering a fixed game list to handling new instances, levels, or rules produced by the benchmark platform \citep[e.g.,][]{aigamescore57,GVGAILLM}. Finally, \emph{open-ended tasks} derive their breadth from large, combinatorial, or effectively unbounded task spaces within a persistent environment \citep[e.g.,][]{MCU,long2024teamcraft,civilbench61}. This axis prevents an important overclaim: broad coverage is not automatically cross-game generalization, and single-world task diversity should not be conflated with transfer across unrelated games.

The final axis is \emph{Modality}, which records which parts of the human perception-action loop remain, are scaffolded, or are abstracted away in the benchmark. On the observation side, benchmarks may be text-symbolic, converting game states into natural language or structured data to maximize diagnostic clarity \citep[e.g.,][]{SmartPlay,GTBench,dsgbench,GVGAILLM}; visual, using raw screenshots or pixel streams to retain more of the native perception burden \citep[e.g.,][]{StarBench,flashadventure6,PokeGym}; or mixed, providing multi-modal inputs or offering both raw and scaffolded tracks for comparison \citep[e.g.,][]{Balrog,GameWorld,gameverse69}. On the action side, benchmarks may rely on semantic controls, where the agent outputs high-level commands, action tuples, or API calls \citep[e.g.,][]{SmartPlay,GTBench,oark5}, or native control, where the agent must execute human-like GUI, keyboard, or mouse actions \citep[e.g.,][]{StarBench,flashadventure6,MCU}. At the taxonomic level, Modality is a factual coding of what burden the agent actually faces, not yet an explanation of how that burden changes score interpretation.

Taken together, these dimensions elevate the taxonomy table from a superficial genre catalog into a comprehensive design-space map. The five-level evolutionary spine \rev{traces the field's historical movement toward environments that abstract away less of human play}, while the Structure, Scope, and Modality axes deconstruct exactly how these environments are engineered. With this architectural framework established, the next section moves a step forward to address the core objective of these environments: what specific capabilities game benchmarks actually measure, and why they serve as a structured substrate for rigorous capability probes.

\subsection{Purpose: Core Capabilities Evaluated by Games}

The taxonomy above organizes game benchmarks by how they instantiate playable environments. Purpose shifts from that structural map to the capability claims those environments are meant to support. The same game substrate can become a different capability probe depending on which pressure the benchmark design foregrounds: rule exposure, state dynamics, information constraints, social interdependence, perception-action burden, or task novelty. This section explains why those structures can make particular capabilities visible, and where each kind of capability claim should remain bounded.

The central question is not whether games are harder than static tasks, but how a benchmark design makes a target capability necessary for play. Capability probing depends on design choices such as how rules are exposed, how states change, how opponents or collaborators constrain action, how much perception and control burden is preserved, and how task variation prevents success from collapsing into memorized routines. These are purpose-level questions: they explain why a game can serve as a probe for an ability. 

Accordingly, each subsection below follows three guiding questions:

\begin{itemize}
    \item Measurement target: What capability is the benchmark trying to put under pressure, and what should not be inferred from that target?
    \item Game affordance and boundary: Which properties of this game structure make the capability surface during play, and which aspects of real gameplay are abstracted away or only weakly tested?
    \item Probe design mechanisms: Which benchmark design choices make the target capability operationally necessary, such as rule-state-action formalization, information asymmetry, scenario selection, role structure, difficulty variation, scaffold contrasts, or generated task variation?
\end{itemize}

\rev{The five levels below follow the evolutionary spine: each names the capability a benchmark family is designed to expose given the interface it admits.}

\subsubsection{Level 1: Rule Understanding}

At the first level, game benchmarks treat games as rule-governed action systems rather than as demonstrations of broad intelligence. The target is the model's ability to enter the game correctly: interpret rules or manuals, map the current state to those rules, maintain state across turns, produce legal actions, and satisfy the output or tool protocol through which the environment accepts moves. This is an entry condition for later capability claims. If a model cannot stay inside the legal action space or reliably update the state after its own moves, poor downstream performance should not yet be read as weak strategy, social reasoning, or visual agency \citep{SmartPlay,GTBench,LLMChess}.

The strength of Level 1 settings is formal containment. Rules, states, actions, and outcomes can be made explicit enough that invalid moves, malformed outputs, tool-use failures, and weak move quality are separated instead of collapsed into a generic loss signal. Chess, grid-game, and game-theoretic benchmarks are useful precisely because simulators or engines can check participation while still leaving room for strategic choice. The same clarity also defines the boundary of the evidence: many such benchmarks expose manuals, state variables, histories, or legal moves through text or API interfaces, removing much of the perceptual and interface-discovery burden of human play  \citep{SmartPlay,GTBench,LLMChess,topsakal2025evaluating}.

At the mechanism level, Level 1 benchmarks make rule understanding observable through three connected designs. Rule-state-action containment makes rules, current states, histories, and admissible actions explicit enough that rule interpretation, state maintenance, and legal-action generation become the minimum conditions for entering the game system \citep{SmartPlay,GTBench,BotzoneBench}. Legal participation diagnostics then separate malformed outputs, illegal moves, tool-call failures, timeouts, and legal but strategically weak actions, preventing low scores from being prematurely interpreted as strategic failure \citep{LLMChess,topsakal2025evaluating,CompleteChessGames}. Finally, situated rule-application probes ask whether models can bind rules to the particular state now in front of them, using rulebook modality, board or grid representations, valid-action questions, and short-horizon optimization tasks \citep{LVLM10}. RuleOracles \citep{RuleOracles} makes this boundary especially clear: richer rulebook access can improve rule retrieval without guaranteeing correct rule application in a concrete game state.

\subsubsection{Level 2: Reasoning}

Level 2 shifts the target from legal participation to interactive decision quality. The question is not simply whether a model can produce a valid move, but whether it can sustain useful choices as states evolve, consequences are delayed, information remains incomplete, and opponents or environments respond. In this sense, reasoning is evaluated as a trajectory property: a plan must survive the action-consequence loop rather than remain plausible only as a one-step explanation \citep{GTBench,GameBench,dsgbench,BeyondScaling}.

The affordance of games at this level is that they turn reasoning into a sequence of commitments. A stated plan must survive contact with changing resources, opponent behavior, partial observability, spatial constraints, and accumulated earlier mistakes. This makes games useful probes of adaptive planning, opponent-aware choice, executable spatial reasoning, and long-horizon strategy. Yet the evidence remains bounded by the interface: many Level 2 benchmarks rely on textified states or API wrappers, so they should be read as controlled probes of strategic pressure rather than direct evidence of human-like play or validated cognitive factors \citep{GameBench,dsgbench,KORGym}.

At the mechanism level, Level 2 benchmarks first use formal strategic pressure to make reasoning consequential. Game-theoretic environments \citep{GTBench,GAMABench} expose incentive structure, information regime, and opponent response in a controlled form, while specialist poker settings \citep{PokerBench,GTOWizardBenchmark} add solver or strong-agent references for hidden-state decisions and action sizing. Coverage-oriented suites then broaden the pressure set across hidden information, stochasticity, cooperation, communication, adaptation, and planning under shared protocols, revealing uneven capability profiles rather than a single strategic score  \citep{GameBench,dsgbench,KORGym}. Finally, spatial and temporal execution probes test whether plans remain effective as they become trajectories: maze and traversal tasks isolate map maintenance and loop avoidance, while real-time or strategy-world environments expose timing, resource allocation, state growth, and the gap between legal and useful actions  \citep{MazeEval,GameTraversalBenchmark,BeyondScaling,towermind,civilbench61}.

\subsubsection{Level 3: Social Intelligence}

Level 3 treats social intelligence as interdependence in play, not as generic knowledge about social situations. A decision is good only relative to what other agents know, want, say, and may do next. The evidence is therefore behavioral: whether a model can turn social inference into consequential play, by identifying hidden roles, calibrating trust, coordinating with partners, or using language to shape later actions \citep{WerewolfArena,AvalonBench,LLMHanabi,LLMCoordination}.

Social games are useful because communication enters the causal loop of the game rather than remaining commentary about it. In deduction games, speech changes suspicion and votes; in cooperative games, a hint, request, or silence can determine whether partners converge on a shared plan; in Diplomacy-style play, negotiation matters only when it remains consistent with executable orders. The boundary is equally important: most current benchmarks make this loop measurable by textifying dialogue, fixing roles or turns, adding rule-based moderators, or scoring labels and offline human references. They are strong probes of instrumented social gameplay, not direct proxies for unconstrained human social competence \citep{WerewolfArena,LLMHanabi,cicero,Wolf,BeyondSurvival}.

At the mechanism level, designs make social reasoning consequential rather than merely reportable. Hidden-role settings turn private identity and role incentives into observable suspicion, voting, and deception dynamics; process-oriented variants then separate deception production, detection, calibration, and human-aligned judgment from final win rate \citep{AvalonBench,WerewolfArena,Wolf,BeyondSurvival}. Cooperative settings shift the pressure from deception to alignment: Hanabi, coordination games, and Overcooked-style tasks test whether agents infer partner knowledge and convert communication into joint action, with ToM scores, CoordQA, and initiating/responding metrics explaining failures beyond task success \citep{LLMHanabi,LLMCoordination,CollabOvercooked}. Negotiation-heavy play then bridges language and strategy by showing that promises and persuasion matter only when coupled to plans that other agents can later act on, while also exposing the standardization cost of ecological human play \citep{cicero}.

\subsubsection{Level 4: Visual Agency}

Level 4 targets the knowing-doing gap: whether a model can convert visual game states into effective action across time. The issue is not perception or planning in isolation, but their coupling with UI grounding, action localization, timing, memory, and recovery. A model may identify the right goal yet still fail by clicking the wrong region, misreading a status cue, repeating stale actions, becoming physically stuck, or losing track of quest progress \citep{StarBench,videogamebench4,PokeGym}.

Visual games are useful because they preserve the screen-to-action loop that textified or API-mediated benchmarks often remove. The agent must decide from a changing visual surface and execute through an interface where small grounding errors alter later states. This makes static visual understanding insufficient: perception, control, and trajectory repair become observable behavioral failures. The boundary, however, is that ecological fidelity also makes scores harder to interpret. Low performance may reflect visual misrecognition, UI grounding, latency, action-format errors, planning weakness, memory collapse, or failed recovery. Level 4 is therefore best understood not as ``more realistic is always better,'' but as the point where interface privilege determines which part of visual agency a benchmark can actually claim to measure \citep{StarBench,Balrog,GameWorld}.

Current benchmarks make this capability visible through three complementary designs. Perception-first probes remove or constrain control to show that gameplay-specific visual grounding and temporal attribution are already brittle before full agency is required \citep{GameplayQA,LVLM10,vmage7}. Matched-interface benchmarks then compare raw or computer-use control with semantic, tool-assisted, or harnessed variants, exposing how much apparent competence depends on the observation and action channel rather than on the game alone \citep{StarBench,GameWorld,lmgbench9,gameverse69}. Finally, long-horizon GUI and 3D-world benchmarks turn local mistakes into trajectories, using milestones, progress scores, deadlock categories, or achievement structures to test whether agents can remember, recover, and keep advancing after errors accumulate \citep{flashadventure6,PokeGym,Crafter}.

\subsubsection{Level 5: Cross-Game Generalization and Open-Ended Task Generalization}

Level 5 should not be equated with "many games." Its target is whether an agent can preserve useful game competence beyond a fixed, known task: when rules, levels, mechanics, task combinations, or even the game substrate change. Rather than naming a single evaluation format, Level 5 marks a family of generalization pressures: retaining competence as the task space expands, as the form of play changes, or as the agent encounters unfamiliar mechanics for the first time \citep{oark5,aigamescore57,ARCAGI3}.

Games are suitable for this purpose because their design space can expand in several non-equivalent directions. A benchmark may enlarge the task distribution inside one world, assemble heterogeneous games under a common protocol, generate new rules or levels, or protect future tests through private and out-of-distribution splits. As a result, current claims about game-agent generalization are still methodologically uneven: the field is moving toward broader and less saturable evaluation, but different benchmarks operationalize that goal through different kinds of novelty, breadth, and openness.

Within this family, three mechanisms should be distinguished. Single-world open-ended benchmarks use rich environments such as Minecraft, Stardew Valley, or survival worlds to multiply goals, resource chains, social or collaborative situations, and recovery demands; they are strongest as evidence for intra-world task generalization, memory, exploration, repair, and long-horizon task composition, not for cross-game transfer \citep{MCU,stardojo11,long2024teamcraft,Crafter}. Broad curated suites instead assemble multiple games or genres under a shared harness, asking whether a model, scaffold, or interface remains robust as the form of play changes; their main contribution is coverage across known game forms, so transfer claims should remain bounded by the fixed suite \citep{oark5,gameverse69,TextQuests,TextAtari,TextArena}. Generated, living, or first-contact benchmarks push the pressure further by making new rules, levels, private games, or hidden mechanics part of the evaluation protocol; these designs are the closest current evidence for anti-saturation, mechanic induction, and first-contact adaptation, although the novelty may still be symbolic, procedural, or abstract rather than unrestricted commercial-game transfer \citep{GVGAILLM,aigamescore57,ARCAGI3}.

The five-level discussion above explains why different game structures make different capabilities necessary for play. Table~\ref{tab:benchmark_results} selects representative reported results from each level to show what current systems can and cannot do under each benchmark's own protocol. The Condition column states the relevant setting or reference before the Result column reports the key performance signal. The table shows a recurring pattern: current models can often enter controlled game systems and sometimes perform well under privileged text or semantic-action interfaces, but their performance drops sharply when success requires search-level planning, social information conversion, visual grounding, long-horizon recovery, or first-contact adaptation.

\begin{table*}[!p]
\centering
\caption{Reported performance signals on representative game benchmarks. Conditions and metrics differ across rows; these results are diagnostic examples, not a cross-benchmark ranking.}
\label{tab:benchmark_results}
\footnotesize
\setlength{\tabcolsep}{4pt}
\renewcommand{\arraystretch}{1.12}
\begin{tabularx}{\textwidth}{>{\raggedright\arraybackslash}p{0.05\textwidth}>{\raggedright\arraybackslash}p{0.125\textwidth}>{\raggedright\arraybackslash}p{0.225\textwidth}>{\raggedright\arraybackslash}p{0.305\textwidth}>{\raggedright\arraybackslash}X}
\toprule
\textbf{Level} & \textbf{Benchmark} & \textbf{Condition} & \textbf{Result} & \textbf{Signal} \\
\midrule
L1 & SmartPlay\newline\citeyearpar{SmartPlay} & HN = human-normalized score, with human baseline = 1.00. & GPT-4-0613 reaches HN 1.00 on Bandit and 0.91 on RPS, but drops to 0.83 on Hanoi, 0.61 on Minecraft, and 0.26 on Crafter & Textified rule play is feasible, but planning, recovery, spatial reasoning, and long-horizon control remain far below humans. \\
\addlinespace[4pt]
L1 & GTBench\newline\citeyearpar{GTBench} & Opponent varies between random/MCTS/Tit-for-Tat and LLM opponents; NRA $\in [-1,1]$ means normalized relative advantage against the opponent. & LLM agents pass the completion-rate sanity check with at least 90\% completion overall, but in complete deterministic games against MCTS, all tested LLM agents report NRA $=-1$ & Legal participation and strategic strength separate sharply; rule-following does not imply search-level play. \\
\addlinespace[4pt]
L2 & KORGym\newline\citeyearpar{KORGym} & Text/API leaderboard reports average percentage over five reasoning dimensions; visual tasks are reported separately. & Text/API Avg.: o3-mini 82\%, Gemini-2.5-Pro 79\%, GPT-4o 22\%. In visual tasks, Gemini-2.5-Pro reaches 90\% on Bubble Ball but only 2\% on Square Addition. & Standardized game APIs reveal reasoning profiles, while visual grounding remains uneven even in controlled game tasks. \\
\addlinespace[4pt]
L2 & PuzzleJAX\newline(2025) & BFS = breadth-first search with a 1M-step cap & BFS solves all levels in 8/12 example games. LLMs are mostly 0\% across games; o3-mini reaches 100\% on Slidings but only 50\% on Sokoban Basic and 25\% on Sokoban Match3. & Puzzle games still favor explicit search; current LLM/RL agents struggle with rule interactions and deadlock-avoiding plans. \\
\addlinespace[4pt]
L3 & Collab-Overcooked\newline\citeyearpar{CollabOvercooked} & Levels 1--6 increase collaboration complexity; SR = success rate, PC = progress completeness. & Claude Sonnet 4 SR falls from 100/100/96/92/74/58\% across Levels 1--6; GPT-4o falls to 2\% at Level 5 and 4\% at Level 6. Humans under a 10s step limit still reach Level-6 SR 90\% and PC 85.6. & Collaboration fails under complexity before basic task understanding disappears; humans remain far more stable. \\
\addlinespace[4pt]
L3 & Werewolf Arena\newline\citeyearpar{WerewolfArena} & No-information and trusted-Seer simulations provide anchors; Seer reveal metrics measure whether social information is disclosed, correct, and believed. & No-information villagers win only 1.2\%; if the Seer automatically reveals a discovered Werewolf and villagers trust it, villager wins rise to 100\%. In played games, believed Seer reveal rates vary sharply. & Social deduction depends not only on hidden information, but on when it is revealed, whether it is trusted, and whether persuasion converts knowledge into votes. \\
\addlinespace[4pt]
L4 & Flash-Adventure\newline\citeyearpar{flashadventure6} & SR = success rate, MCR = milestone completion rate; COAST = native computer-use agent plus clue-memory Seek--Map--Solve scaffold. & Under a 1,000-step cap, COAST reaches highest SR 5.88\%, and best MCR 19.89\%, while human anchors reach SR/MCR 50.98/78.98\% with the same cap and 97.06/100.00\% without the cap. & Full-story GUI play remains far below human performance; clue memory improves progress slightly, but long-horizon observation--behavior gaps still dominate. \\
\addlinespace[4pt]
L4 & LMGame-Bench\newline\citeyearpar{lmgbench9} & No harness vs harness with perception, memory, and reasoning modules; Glass's effect size = standardized distance from random baseline. & Model--game mean scores above random: 36/60 without harness, 52/60 with harness (excluding text-only models). Positive Glass's effect size: 38/40 harnessed model-game pairs vs 26/40 without harness. & Harnesses can turn near-random play into discriminative scores, so reported ability is strongly scaffold-dependent. \\
\addlinespace[4pt]
L5 & ARC-AGI-3\newline\citeyearpar{ARCAGI3} & Official = no harness on semi-private environments; RHAE = Relative Human Action Efficiency. & Humans solve 100\% of included environments. The best official release score is only 0.37\% RHAE. & First-contact mechanic induction is still largely unsolved. \\
\addlinespace[4pt]
L5 & AI Game-Store\newline\citeyearpar{aigamescore57} & Scores are normalized to per-game human median 100, clipped to $[1,10{,}000]$, then geometrically averaged over 100 games. & GPT-5.2 has the highest reported aggregate, 8.5; the seven model aggregates range from 4.7 to 8.5. This aggregate does not identify the fraction of games exceeding a competence threshold. & Broad human-game coverage exposes a large human-relative gap, especially in memory, planning, and world-model learning. \\
\bottomrule
\end{tabularx}
\end{table*}

\subsection{Paradigm: From Interaction to Assessment}

A game environment does not automatically constitute a benchmark merely by being interactive or difficult. It becomes an evaluation instrument only when gameplay is mediated by an interaction contract and an evaluation contract. The former determines what part of the human play loop is exposed to the model, while the latter determines what kind of evidence can be extracted from the resulting behavior. In this sense, the methodological challenge of game benchmarking is not simply to select harder games, but to transform gameplay into a measurement pipeline that is interpretable, comparable, and sustainable as model capabilities evolve.

\subsubsection{Interaction}

Interface design operates as the front end of this evaluation pipeline. It specifies what the model is allowed to see, how it is allowed to act, and which cognitive or operational burdens of human play are preserved, simplified, or removed. A textified or API-mediated game can be a precise instrument for reasoning, while a visual native-control game can be a stronger test of end-to-end agency; neither is intrinsically superior unless its privilege level matches the claim being made. In this context, interaction design should be treated as a core part of benchmark validity rather than as an implementation detail.

\textbf{Observation} channels trace one of the clearest paradigm shifts in game benchmarks: the transition from abstract textual or symbolic state representations to raw visual streams. Textified interfaces align naturally with language-centric LLMs and excel in reasoning-oriented evaluations. By explicitly exposing rules, state variables, histories, and action constraints, they support highly controlled diagnostic probing \citep{SmartPlay,GTBench,dsgbench}. However, scaling such interfaces to complex visual or commercial games necessitates bespoke APIs, wrappers, or state-extraction procedures. This fundamentally bypasses the challenge of inferring game states directly from the play surface.

Therefore, the shift toward raw visual input is not simply a change in modality; it is a commitment to more human-like, end-to-end agent evaluation. Models are forced to interpret screen pixels, track dynamic visual changes, and make decisions under severe perceptual uncertainty. Unsurprisingly, current systems struggle in these settings, as raw visual observations reintroduce the state noise, grounding ambiguity, and temporal instability that text abstractions had neatly excised \citep{videogamebench4,PokeGym,GameplayQA}. To bridge this gap, hybrid designs---such as OCR-assisted, detector-assisted, or dual text-and-image tracks---occupy a critical middle ground. They recognize that while pure vision best approximates human play, providing intermediate perceptual scaffolding yields a more diagnostic benchmark, isolating higher-order reasoning failures from low-level perceptual bottlenecks \citep{StarBench,oark5}.

\textbf{Action} channels determine how a model's decisions manifest within the game environment. Semantic actions regularize the agent's output space by restricting choices to high-level moves, skills, targets, or predefined commands that are seamlessly parsed and executed by the benchmark harness \citep{SmartPlay,GTBench,dsgbench,oark5}. By eliminating the friction of motor execution and UI manipulation, this design directs evaluation toward pure reasoning, planning, and decision quality. Native control, by contrast, requires the model to interact through interfaces originally designed for human players, such as mouse, keyboard, controller, or GUI actions. This elevates the agentic requirements of the task: the model must map an intended strategy onto executable and time-sensitive interactions. Furthermore, native-control interfaces provide a scalable pathway for evaluating commercial and browser games. By leveraging standardized, human-facing control channels, researchers can bypass the prohibitive engineering cost of developing custom symbolic APIs for every new game environment \citep{videogamebench4,flashadventure6,StarBench}. Dual-interface benchmarks show why this distinction is substantive rather than cosmetic: the same or similar gameplay can look much easier when semantic actions replace low-level control \citep{gameverse69,GameWorld}.

Observation and action choices together form an interface-privilege hierarchy. High-privilege settings are valuable when the research question is strategic reasoning, rule application, or diagnostic comparability; low-privilege settings are valuable when the target is end-to-end gameplay which preserves visual grounding, UI operation, timing, and recovery. The risk is overclaiming across that boundary. Scores in high-privilege settings provide limited evidence about whether a model can understand and act within an unconstrained environment as presented to humans, while a low score in a raw visual benchmark may conflate planning failure with perception, latency, control formatting, or recovery failure \citep{videogamebench4,PokeGym,StarBench,GameWorld}. A rigorous game benchmark must explicitly document its privilege level and, ideally, provide comparative tracks between assisted and human-like settings to successfully disentangle reasoning competence from grounding and execution failures.

\subsubsection{Evaluation}

The evaluation contract determines how agent behavior is synthesized into evidence. Games naturally contain goals, rewards, progress, and failure states, but these native signals do not automatically constitute a rigorous benchmark. A game becomes a benchmark only when its outcomes are translated into metrics that are interpretable, comparable, and robust against the noise of stochastic interactions.

The simplest contract is result-based evaluation. Win rate, score, reward, survival, completion, and normalized progress preserve the native objective structure of play and are easy to automate at scale, which explains their prevalence from textified strategic suites to broader agent benchmarks \citep{SmartPlay,GTBench,Balrog,GameWorld}. Their weakness is diagnostic opacity. The same low score may reflect poor planning, rule misunderstanding, visual misrecognition, invalid actions, latency, memory failure, or weak recovery. This problem becomes especially severe in visually grounded or long-horizon games, where current models often achieve extremely low native progress and the metric loses diagnostic resolution \citep{videogamebench4,StarBench}. Furthermore, native scores suffer from weak semantic portability: a score in one game rarely maps to the same score in another.

Process-level evaluation addresses that opacity by instrumenting the trajectory rather than only the endpoint. These metrics decompose gameplay into intermediate milestones, trajectory analysis, sub-skill scores and reasoning checks. Validated trajectory scores can reveal whether agents fail through bad decisions, lost state, weak control, or poor recovery \citep{gamebot,dsgbench,GameWorld,MCU,gameverse69}. This makes process metrics especially valuable for long-horizon and open-ended games, where binary success hides most of the behavior. The trade-off is that decomposition is itself a design choice: once a benchmark decomposes gameplay into subskills or milestones, the decomposition itself reflects the benchmark creator's prior assumptions about positive behavior. Process metrics are therefore best treated as an explanatory layer around outcomes, not as a universal cross-game currency.

Relative and adversarial evaluation adds a different kind of evidence. By evaluating agents against other models, fixed opponents, or live human participants, these protocols expose capabilities that only appear in interaction, such as opponent modeling, deception, negotiation, and adaptation \citep{GTBench,WerewolfArena,gamearena27}. Consequently, adversarial frameworks maintain benchmark freshness and capture dimensions of social intelligence that isolated, single-agent tasks inherently miss. Yet the resulting scores are relational by construction. A tournament rank or arena win rate reports performance against a particular opponent pool under a particular protocol. Adversarial evaluation is therefore powerful for dynamic stress testing, but weak as an absolute measure unless its competitive context is made explicit.

Calibration is therefore necessary for giving game scores external semantics. By introducing fixed AI anchors, explicitly tiered opponent ladders, or calibrated solver references, benchmark designers can transform fluctuating relative performances into interpretable claims \citep{SmartPlay,BotzoneBench,ARCAGI3,GameWorld}. This ensures that the same numerical score is not overinterpreted across different games or varying interface privileges within a suite. A 60\% win rate, a progress score, or an Elo-like rating is meaningful only relative to the baseline, interface privilege, and task distribution that produced it.

Operationally, Eq.~\ref{eq:normalized-competence} requires per-game scores and compatible lower and expert anchors; Eq.~\ref{eq:generalist-criterion} additionally requires game weights and a competence threshold. A random baseline alone does not identify the expert-normalized scale. Glass's effect size, which divides by the random baseline's standard deviation, is a different statistic: zero random variance makes that statistic undefined, not necessarily the denominator in Eq.~\ref{eq:normalized-competence}. Estimating Eq.~\ref{eq:generalization-gap} also requires training provenance and a declared test split. Procedural levels test one form of transfer \citep{cobbe2020leveraging,NetHackLearningEnvironment}; held-out-title evaluations such as NitroGen's test another \citep{nitrogen}. Neither should be described as unrestricted cross-world transfer without specifying which observation, action, and dynamics changes are actually withheld.

Robustness then asks whether that meaning survives reasonable variation and benchmark pressure. Stronger protocols now use repeated runs, seeded duplicate matches, private or held-out environments, refreshed instances, contamination checks, scaffold ablations, and public/private leaderboard separation to reduce noise, leakage, and benchmark-specific optimization \citep{ARCAGI3,lmgbench9,Clembench2024,GameWorld}. The important point is not that any single defense solves robustness, but that evaluation design must report which threats it addresses and which remain. This is particularly important in game benchmarks, where interaction traces, agent scaffolds, and public gameplay knowledge can all become part of the measured system.

Taken together, these paradigms show that game evaluation should be read as a measurement stack rather than a scoreboard. Outcome metrics preserve the game's objective; process metrics explain the route through the game; adversarial protocols introduce interactive pressure; calibration supplies external reference points; and robustness protects those interpretations over time. The most rigorous game benchmarks are therefore not simply those built on difficult games, but those that make explicit what their scores can and cannot support as evidence about agent capability.

\subsection{Comparing Competence, Coverage, and Harness Dependence}
\label{sec:generality-comparison}

\textbf{Four benchmarks, four complementary questions.} Table~\ref{tab:generality-comparison} compares selected published results from GameVerse, GameWorld, OmniGameArena, and GameHorizon Suite. They test reflection, action interfaces, breadth, and temporal composition, respectively; their metrics cannot be pooled into one ranking. We distinguish paper versions from later official result snapshots and retain each source's evaluation unit. For $N$ equally weighted games with compatible anchors, the empirical counterpart of Eq.~\ref{eq:generalist-criterion} is
\[
\widehat q(\tau)=\frac{1}{N}\sum_{i=1}^{N}\mathbf 1\{\widehat{\mathcal C}_i\geq\tau\},
\qquad \widehat\delta(\tau)=1-\widehat q(\tau).
\]
The OmniGameArena block illustrates this calculation with $N=7$, conditional on its published random and human-expert reference rows.\footnote{For ObstacleRun2D, ObstacleRun3D, LastStand, MonsterShoot, SceneEscape, CueChase, and SoloCraft, respectively, the lower anchors are $(0.030,0,0.088,0.008,0,0,0)$ and expert anchors $(1,0.991,0.965,1,1,1,1)$. We compute $(s_i-b_i)/(h_i-b_i)$ without clipping and threshold before rounding. The data asset is \mbox{data.js?v=143}, accessed 21 October 2026 \citep{omnigamearenaWeb2026}.} Its human sample, practice allowance, and budget matching are incompletely documented, so this is a descriptive use of published references, not a verified matched-protocol generality guarantee.

\begin{table}[!p]
\centering
\caption{\textbf{Four complementary performance comparisons.} Rows select the top three models by after-reflection mean (A) or mean competence (C), three 27B checkpoints and Flash-Next under computer-use actions (B), and the top four by online subtask success (D); they do not form a cross-benchmark leaderboard. A: equal-weight means recomputed from the 15 per-game means in GameVerse v2, Table~3; its original ``Avg'' column reports rank. B: GameWorld's 9 October 2026 snapshot, 170 tasks in 34 games, three runs per task and a 400-step cap. C: OmniGameArena's official web snapshot, seven Solo games and five episodes per model--game cell; coverage counts games, not successful episodes. D: GameHorizon v2, Tables~4 and~9; offline accuracy covers 3,000 primary questions, whereas both online measures concern Minecraft. All values are descriptive point estimates under their source protocols.}
\label{tab:generality-comparison}
\fontsize{9}{11}\selectfont
\setlength{\tabcolsep}{5pt}
\renewcommand{\arraystretch}{1.13}
\begin{tabularx}{\textwidth}{@{}>{\raggedright\arraybackslash}Xrrr@{}}
\toprule
\multicolumn{4}{@{}l}{\textbf{A. GameVerse: video reflection at fixed model weights} \citep{gameverse69}} \\
\textbf{Model (GUI interface)} & \textbf{Before (/100)} & \textbf{After (/100)} & \textbf{Gain (points)} \\
\midrule
Qwen3-VL-32B & 29.86 & 32.94 & +3.08 \\
Seed-1.8 & 30.75 & 35.25 & +4.50 \\
Gemini-2.5-Pro & 30.26 & 36.76 & +6.50 \\
\addlinespace[7pt]
\multicolumn{4}{@{}l}{\textbf{B. GameWorld: computer-use performance} \citep{gameworldLeaderboard2026}} \\
\textbf{Model (CUA interface)} & \textbf{Success SR (\%)} & \multicolumn{2}{r}{\textbf{Progress PG (\%)}} \\
\midrule
Qwen3.5-27B & 21.96 & \multicolumn{2}{r}{41.79} \\
Qwen3.6-27B & 21.18 & \multicolumn{2}{r}{43.60} \\
Qwen3.8-27B & 29.22 & \multicolumn{2}{r}{54.64} \\
Qwen3.8-Flash-Next & 36.86 & \multicolumn{2}{r}{61.48} \\
\addlinespace[7pt]
\multicolumn{4}{@{}l}{\textbf{C. OmniGameArena: mean competence versus coverage} \citep{omnigamearenaWeb2026}} \\
\textbf{Model (VLM track)} & \textbf{Mean $\widehat{\mathcal C}$} & \textbf{$\widehat q(0.5)$} & \textbf{$\widehat q(0.9)$} \\
\midrule
GPT-5.5 & 0.457 & 3/7 & 0/7 \\
Claude Opus 4.6 & 0.392 & 2/7 & 0/7 \\
Gemini 3.1 Pro Preview & 0.375 & 3/7 & 0/7 \\
\addlinespace[7pt]
\multicolumn{4}{@{}l}{\textbf{D. GameHorizon Suite: offline and online performance} \citep{gamehorizon2026}} \\
\textbf{Model} & \textbf{Offline accuracy (\%)} & \textbf{Subtasks (/62; \%)} & \textbf{Long tasks (/20; \%)} \\
\midrule
GPT-6-Astra & 80.2 & 41/62; 66.1 & 9/20; 45.0 \\
Gemini 3.6 Flash & 75.3 & 35/62; 56.5 & 6/20; 30.0 \\
Kimi-K3 & 74.5 & 29/62; 46.8 & 2/20; 10.0 \\
GPT-5.6 Terra & 70.7 & 23/62; 37.1 & 2/20; 10.0 \\
\bottomrule
\end{tabularx}
\vspace{4pt}
\begin{minipage}{\textwidth}\fontsize{8.5}{10}\selectfont
A: scores use game-specific scales mapped to $[0,100]$, not human-normalized competence; trials and interaction budgets vary by game. Reflection retries the same target tasks. B: SR is mean task-run success (not pass@3); PG is clipped target-relative progress. C: VLM play pauses during inference; native real-time policies, PvP, and cooperation are excluded. Human calibration and uncertainty limit threshold interpretation. D: up to 400 calls per subtask; failure-triggered resets permit diagnosis of later subtasks, but disqualify whole-task success. Offline accuracy is not interactive completion.
\end{minipage}
\end{table}

\textbf{Reflection gains measure adaptation under a specified harness.} GameVerse's video reflections raise the selected models' average GUI scores, with Gemini-2.5-Pro increasing from 30.26 to 36.76. These gains use extra experience on the target tasks, with unchanged weights; they do not establish transfer to unseen games. The dataset pillar determines which failures and tutorials are available, while the model and harness determine how they enter the next attempt. Benchmark ablations should match the practice and inference budget and test fresh episodes after reflection, then revisit prior games to measure retention.

\textbf{An interface is part of the evaluated system.} GameWorld's selected checkpoints differ in success and progress under one CUA protocol (Table~\ref{tab:generality-comparison}); this comparison does not isolate the effects of training data or architecture. Holding Qwen3.8-Flash-Next fixed instead exposes the action-interface effect in Table~\ref{tab:intervention-comparison}. Both interfaces use screenshots, but semantic commands change how model outputs are translated into controls and do not uniformly improve performance. Progress is relative to task targets, and the original paper's human results use a shorter 100-step protocol, so they cannot calibrate this 400-step snapshot. Equal decision-step caps also do not establish equal token or wall-clock cost.

\textbf{Higher mean competence need not imply broader coverage.} In OmniGameArena, Opus~4.6 has a higher mean than Gemini~3.1 Pro Preview but reaches $0.5$ on fewer games ($2/7$ versus $3/7$). None of the ten VLMs in the saved snapshot reaches $0.9$ on any Solo game. These counts use rounded published means and are sensitive to trial variability: GPT-5.5's SoloCraft score is $0.504$ with reported standard deviation $0.046$, close to the $0.5$ boundary. They demonstrate the distinction between mean and coverage, not a confidence-certified $(\epsilon,\delta)$ claim or unseen-game transfer. Per-game data, compatible anchors, repeated runs, and declared game weights are necessary for stronger conclusions.

\textbf{Local success does not establish reliable temporal composition.} GameHorizon separates offline reasoning questions from online Minecraft control. GPT-6-Astra records 80.2\% offline accuracy, passes 41 of 62 diagnostic subtasks, and completes 9 of 20 long tasks. These denominators and task formats differ; subtracting the percentages would not measure a calibrated loss of competence. Failure-triggered intermediate resets make subsequent subtasks assessable, but do not demonstrate recovery. Across the four pillars, this motivates failure/recovery data, persistent state representations, budgeted planning, and whole-chain tests without oracle resets. The 21-game offline corpus cannot substitute for multi-game interactive evaluation.

\textbf{Compare interventions as well as model scores.} Table~\ref{tab:intervention-comparison} asks which supplied information or deployment component changes performance within each benchmark. It complements Table~\ref{tab:generality-comparison}: reflection-source ablations, interface changes, practice curves, and instruction-context variants reveal different dependencies of the effective model--harness system. The numbers retain their native scales; available budgets are reported separately from costs that the sources do not quantify.

\begin{table}[!p]
\centering
\caption{\textbf{Interventions, measured gains, and evaluation budgets.} Four within-benchmark comparisons complement the overall performance in Table~\ref{tab:generality-comparison}. They change reflection inputs, action interfaces, skill prompts, or instruction context while keeping the compared model weights fixed. They are not uniformly matched for computation. Values are reported point estimates and budgets are declared protocol settings, not measured total resource use; no cross-benchmark efficiency ranking or statistical superiority is inferred. Source versions follow Table~\ref{tab:generality-comparison}.}
\label{tab:intervention-comparison}
\fontsize{9}{11}\selectfont
\setlength{\tabcolsep}{3pt}
\renewcommand{\arraystretch}{1.08}
\begin{tabularx}{\textwidth}{@{}>{\raggedright\arraybackslash}p{0.18\textwidth}>{\raggedright\arraybackslash}p{0.17\textwidth}>{\raggedright\arraybackslash}p{0.31\textwidth}>{\raggedright\arraybackslash}X@{}}
\toprule
\textbf{Benchmark / model} & \textbf{What changes?} & \textbf{Measured outcome} & \textbf{Budget and interpretation} \\
\midrule
\textbf{GameVerse}\newline Qwen3-VL-32B\newline \citep{gameverse69}
& Reflection from failure video, expert tutorial, or both; three semantic-interface games.
& Mean score (/100):\newline None: 61.0; failure only: 72.0; tutorial only: 62.9; both: 76.7.\newline Source: v2, Table~5.
& Extra video analysis and a retry. Total tokens and latency are not reported. Same-task gains do not establish held-out-game transfer. \\
\addlinespace[8pt]
\textbf{GameWorld}\newline Qwen3.8-Flash-Next\newline \citep{gameworldLeaderboard2026}
& Semantic commands $\rightarrow$ direct computer-use actions, with screenshot observations in both.
& Success: $29.80\%\rightarrow36.86\%$ ($+7.06$ points).\newline Progress: $55.86\%\rightarrow61.48\%$ ($+5.62$ points).
& Same 400-decision-step cap; 170 tasks, three runs per task per interface. Equal step caps do not imply equal elapsed time, token use, or primitive-action duration. \\
\addlinespace[8pt]
\textbf{OmniGameArena}\newline GPT-5.5 / Opus~4.7\newline \citep{omnigamearenaWeb2026}
& Repeated skill-prompt revision on LastStand, then best-skill tests on four held-out variants.
& Native score, cold $\rightarrow$ best $\rightarrow$ final:\newline GPT-5.5: $0.408\rightarrow0.956\rightarrow0.937$; best-skill gains are positive on 4/4 variants.\newline Opus~4.7: $0.308\rightarrow0.928\rightarrow0.529$; best-skill gains are positive on 0/4 variants.
& Declared practice: $10\times5$ new episodes, plus five reused cold starts. Default variant tests add $4\times2\times5$ episodes. Reflector/judge costs are unreported; transfer remains within one game. \\
\addlinespace[8pt]
\textbf{GameHorizon}\newline Gemini~3.5 Flash; four-model mean\newline \citep{gamehorizon2026}
& Frames only $\rightarrow$ frames with short-horizon $\rightarrow$ multi-horizon instructions.
& Future-action MCQ accuracy:\newline Gemini~3.5 Flash: $34.0\%\rightarrow39.0\%\rightarrow47.5\%$.\newline Four-model mean: $34.6\%\rightarrow34.8\%\rightarrow41.8\%$ (source-rounded).\newline Source: v2, Table~7.
& 200 questions per model per variant. Extra goal context; tokens and latency are unreported. Offline accuracy gains do not establish improved online control or training-data benefits. \\
\bottomrule
\end{tabularx}
\vspace{4pt}
\begin{minipage}{\textwidth}\fontsize{8.5}{10}\selectfont
GameVerse's three games are TicTacToe, AngryBird, and PvZ. OmniGameArena's best/final rounds are R1/R10 for GPT-5.5 and R7/R10 for Opus~4.7; variant counts compare each selected best-round skill with the same model's no-skill baseline. GameHorizon's mean covers Qwen2.5-VL-32B-Instruct, Qwen3-VL-235B-A22B-Instruct, Gemini~3.1 Pro, and Gemini~3.5 Flash. These MCQ variants differ from its 3,000-question primary test in Table~\ref{tab:generality-comparison}.
\end{minipage}
\end{table}

The comparison connects the four pillars through testable interventions. GameVerse and GameHorizon vary the information supplied to a fixed model: which experience or goal context helps is a data and representation question, while reflection and prompt construction belong to the harness. GameWorld changes the action interface; its result concerns how the model and adapter work together. OmniGameArena shows why the benchmark must distinguish peak practice performance, final performance, and held-out variants: similar peak gains can have different transfer behavior. None of these comparisons alone demonstrates retained competence across prior games.

The remaining cost question is substantive. Additional video analysis, goal context, and skill revision consume resources, but the available results do not provide a common accounting of tokens, auxiliary-model calls, or elapsed time. A stronger comparison would freeze the model and harness rules, allocate the same total interaction and inference budget, then report fresh-test performance and coverage together with cost. This directly tests the reasoning--reactivity and model--harness trade-offs without treating every additional module or longer prompt as an improvement.

\section{Challenge and Future}

The preceding sections have traced the full pipeline of large foundation models in the multiverse of games, from datasets through models and harness to benchmarks. At every stage, a consistent pattern has emerged: progress on one axis exposes a bottleneck on another. This section distills these recurring bottlenecks into five fundamental trade-offs that define the current frontier, and then charts a five-step roadmap toward the generalist game player envisioned by this survey.

\subsection{Five Fundamental Trade-Offs}
\label{sec:tradeoffs}

The challenges facing game-playing foundation models recur across the four pillars. We organize them as five trade-offs, denoted T1--T5 below: improving one objective can increase costs elsewhere, but the strength of each tension must be tested under matched conditions rather than inferred from rankings of unrelated systems.

T1 concerns data scale, fidelity, and diversity; T2 breadth versus depth; T3 reasoning versus reactivity; T4 modular harnesses versus end-to-end models; and T5 code engines versus learned world models. They constrain different parts of an effective game-playing system: its training distribution, action representation, computational latency, division of responsibility between model and harness, and source of transition and reward signals.

\begin{figure*}[t]
  \centering
  \includegraphics[width=1\linewidth]{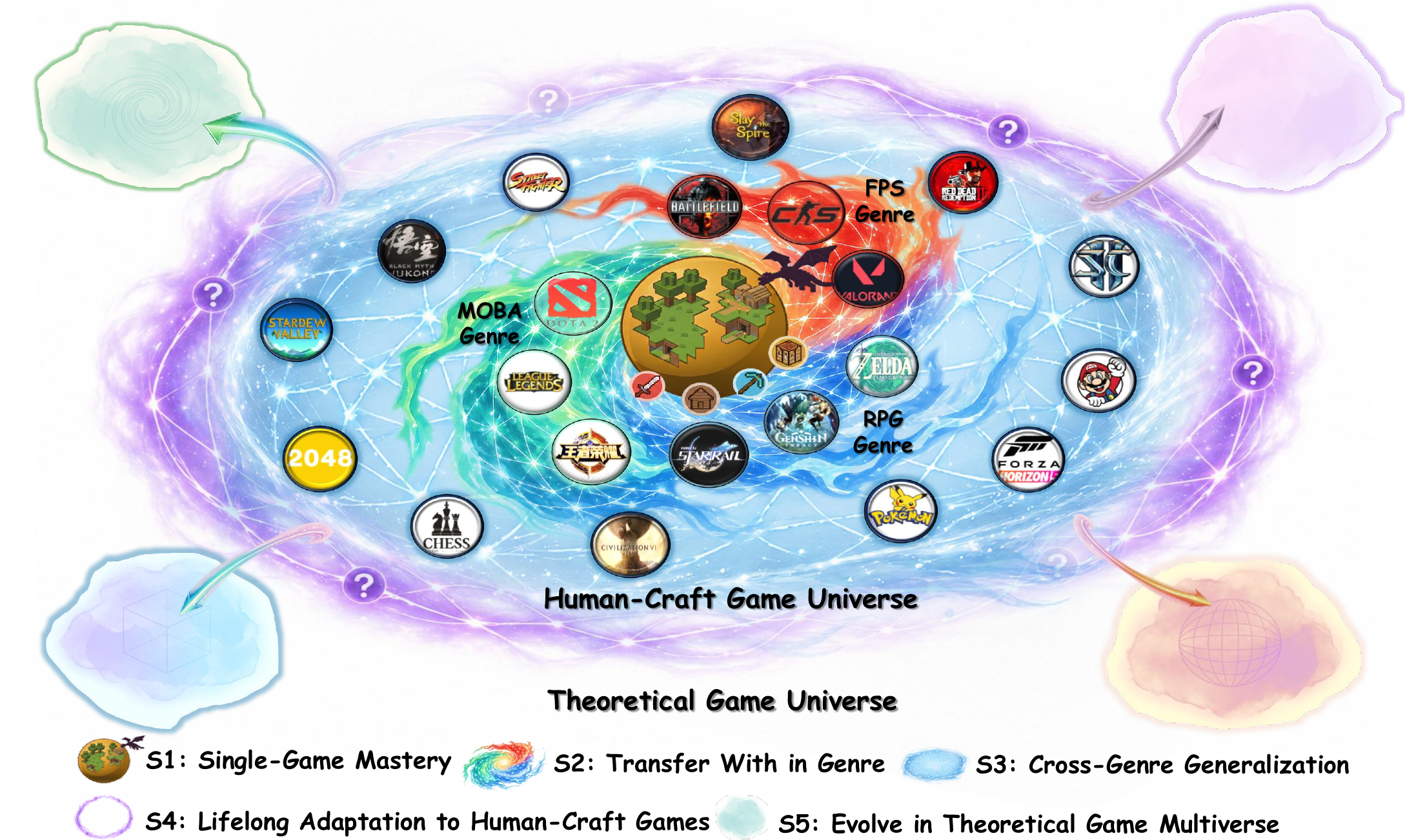}
  \caption{\textbf{Overview of five-step roadmap towards generalist game player.} Steps 1--5 appear as S1--S5 in the figure. The roadmap organizes research from single-game mastery (Step 1) to agents tested for cross-task transfer within genres (Step 2), cross-genre generalization (Step 3), and lifelong adaptation to unseen environments in human-craft game universe (Step 4), culminating in the ``Demiurge'' stage (Step 5), where agents would generate, simulate, and evolve game worlds. This final step is prospective. The steps reflect a widening scope of agency, from solving tasks to constructing and expanding the game multiverse.}
  \label{fig:challenge_teaser}
\end{figure*}

\subsubsection{Scale vs. Fidelity vs. Diversity: The Data Trilemma}

Section 3 identifies distinct data acquisition strategies rather than a single dominant dataset. NitroGen \citep{nitrogen} assembles 40,000 hours across 1,000+ games through controller-overlay extraction; validation against recorded inputs reports 0.96 per-frame button accuracy and joystick $R^2=0.84$. These per-frame measures do not establish trajectory-level fidelity. Open-P2P \citep{openp2p89} records 8,300 hours across 42 3D titles with logged actions, trading some breadth for a more direct supervision channel. VPT's \citep{vpt} 70,000 hours of IDM-pseudo-labeled data cover Minecraft, while PLAICraft's \citep{plaicraft} temporally aligned multimodal corpus is also confined to that game.

The research priority is to measure the contribution of each axis separately: hold the model and training budget fixed while varying label quality, trajectory volume, or game diversity. Learned annotation and synthetic data are possible routes to scale, but both require validation against recorded inputs or executable dynamics. Larger corpora alone do not certify better control, and cross-paper comparisons do not identify which data property caused a capability gain.

\subsubsection{Breadth vs. Depth: The Heterogeneity Wall}

Action semantics and task structure differ across games: the same key or visual cue may serve different purposes. Increasing the number of training titles therefore creates a representation problem as well as a data problem. The objective is to share useful control and planning structure without erasing the distinctions needed for expert play.

Specialists such as Metamon \citep{metamon} and CombatVLA \citep{combatvla} emphasize depth in competitive Pokemon and ARPG combat, respectively. NitroGen \citep{nitrogen} and Game-TARS \citep{gametars} emphasize broader game coverage, while Lumine \citep{lumine} reports storyline transfer among related RPGs. These systems differ in tasks, training, and evaluation interfaces, so they illustrate design choices rather than establish a causal law that breadth reduces depth. A useful comparison would train the same architecture on progressively broader, balanced game sets and measure both previously mastered and held-out tasks under one protocol.

Unified representations, such as NitroGen's \citep{nitrogen} gamepad vector or OmniJARVIS's \citep{omnijarvis} FSQ action tokens, should therefore be assessed for both transfer and within-game precision. A representation passes this test only if improvements on new games do not conceal regressions on earlier ones.

\subsubsection{Reasoning vs. Reactivity: The Latency-Intelligence Dilemma}

Strategic games and action games impose opposite demands on the same model. CICERO's \citep{cicero} multi-turn Diplomacy campaigns and LSPO's \citep{xu2025learning} equilibrium-finding in Werewolf require deep, multi-step reasoning that consumes hundreds of tokens. FPS and MOBA games require responses within 100 milliseconds.

Reported systems expose the cost of reconciling these demands. On VideoGameBench \citep{videogamebench4}, Gemini 2.5 Pro \citep{gemini2.526}, completes only 0.48\% of game tasks when operating in real time; the benchmark had to introduce a game-pausing mode to reach even 1.6\%. Existing solutions trade one end for the other: CombatVLA \citep{combatvla} truncates its Action-of-Thought chain to achieve a 50$\times$ speedup, reducing the explicit reasoning available at action time; Lumine \citep{lumine} reasons at 5Hz and acts at 30Hz by invoking deliberation only when needed; NitroGen's \citep{nitrogen} flow-matching DiT generates 16-step action chunks in parallel, eliminating the autoregressive bottleneck without an explicit language reasoning stage. Game-TARS \citep{gametars} introduces a "Greedy Thinking" strategy, but finds that excessive reasoning in fast-paced games triggers hallucinated reasoning loops that degrade performance below the no-reasoning baseline.

Autoregressive deliberation couples additional tokens to additional latency. Separating slow planning from fast control is a promising response, but its benefit must be measured with the environment clock running. Equal action budgets alone are insufficient: the evaluation must also report wall-clock limits, pausing rules, and the distribution of response delays.

\subsubsection{Modular Workflow vs. Model-as-Whole: The Harness Paradox}

Section 5 documented the harness as the nervous system connecting foundation models to game environments. The paradox is that current models need this external scaffolding to function, yet the scaffolding itself introduces fragility and limits generality.

VideoGameBench \citep{videogamebench4} and GameVerse \citep{gameverse69} report substantial failures when frontier VLMs act from screenshots. Cradle \citep{cradle14} and Voyager \citep{voyager8} instead combine models with explicit perception, planning, memory, or executable skill libraries. These results motivate scaffold ablations, not a conclusion that every improvement comes from modularity. The ``knowing-doing gap'' described in Section 5.2 may arise from perception, action grounding, latency, or errors at module boundaries. Isolating these sources requires holding the model, interface privileges, and evaluation budget fixed while adding or removing harness components.

Native memory and control could reduce external engineering, while an explicit harness can retain advantages for inspection, recovery, and budget enforcement. The relevant goal is a reliable effective system, not the elimination of modules as an end in itself. Model and harness advances should therefore be compared at matched capability and cost.

\subsubsection{Code Engine vs. World Model: The Simulation Gap}

Most training and evaluation depend on code-based game engines. These engines provide explicit rules, instrumentable rewards, and reproducible rollouts under controlled seeds, but they impose three ceilings: the action space is limited to predefined interfaces, the game diversity is limited to what humans have built, and each new game requires dedicated integration effort. World models offer a path beyond these ceilings. Genie \citep{genie} discovers latent action spaces from unlabeled video, removing the interface constraint. GameFactory \citep{gamefactory_wm} and GameGen-X \citep{gamegenx} generate interactive environments beyond the boundaries of existing games, pushing toward the theoretical multiverse of Era 4. PAN \citep{pan_wm} extends action conditioning to natural language, and Solaris \citep{solaris_wm} introduces multi-agent shared worlds with cross-player consistency. SIMA 2 \citep{sima2-87} also reports interaction and self-improvement in Genie 3 worlds \citep{genie3-86}, motivating further study of generated training environments.

Visual plausibility is not equivalent to a reliable training simulator. Genie 3 \citep{genie3-86} reports consistency over several minutes, but long-horizon state persistence and faithful consequences remain evaluation concerns. GameNGen \citep{gamengen_wm} studies simulation in DOOM; this does not by itself demonstrate reliable simulation across unrelated games. A generated scene can look coherent while violating an inventory constraint or changing a reward condition. Such errors may let an agent exploit the simulator rather than learn a transferable skill.

Synthetic-training claims should therefore pair long-rollout consistency tests with independent checks of dynamics and rewards, then evaluate trained agents in held-out code engines. A matched-budget comparison against training without synthetic trajectories is needed before attributing a transfer benefit to the world model.

\subsection{Five-Step Roadmap to the Future}
\label{sec:roadmap}

The roadmap organizes five research targets by what changes between training and evaluation: goals within one game (S1), related games with shared interaction structure (S2), structurally different games (S3), behavior after bounded interaction with an unknown game (S4), and generated environments (S5). Table~\ref{tab:roadmap-tradeoffs} uses GameVerse, GameWorld, OmniGameArena, and GameHorizon to connect these targets to measured limitations, binding trade-offs, four-pillar priorities, and evaluations that could substantiate progress. A benchmark need not demonstrate an entire step to expose a requirement for it.

These links are many-to-many. Long task chains need failure data (T1) and coordinated planning and control (T3--T4); cross-game transfer adds representation and coverage constraints (T2). Synthetic data may ease T1 while introducing simulator error (T5). The priorities are hypotheses for controlled experiments, not causal findings from rankings. Table~\ref{tab:generality-comparison} distinguishes the four benchmarks' measurement units: task completion, target-relative progress, and expert-normalized coverage cannot be interchanged. In the OmniGameArena calculation, Opus~4.6 has a higher mean competence than Gemini~3.1 Pro Preview but lower coverage at $0.5$ ($2/7$ versus $3/7$), illustrating why both quantities matter. Following Section~\ref{sec:generalist-definition}, each gate requires a declared protocol, held-out episodes, fixed model parameters and harness rules, and interaction and wall-clock budgets. Where compatible anchors exist, report per-game competence, threshold coverage and failure fraction with uncertainty; otherwise retain the source metric. S2--S3 require high held-out performance as well as a limited gap. S4 adds before/after adaptation and retention tests. None of these four benchmarks evaluates world generation, so S5 requires an extension.

\begin{table*}[!p]
\centering
\caption{\textbf{Four benchmarks inform five research steps.} T1: data scale/fidelity/diversity; T2: breadth/depth; T3: reasoning/reactivity; T4: modular harness/end-to-end model; T5: code engine/world model. D, M, H, B: datasets, models, harnesses, benchmarks. Gates require predeclared thresholds, budgets, and test splits; current evidence does not establish that a step is passed. S5 is an explicitly untested extension.}
\label{tab:roadmap-tradeoffs}
\fontsize{8.5}{10}\selectfont
\setlength{\tabcolsep}{3pt}
\renewcommand{\arraystretch}{1.05}
\begin{tabularx}{\textwidth}{@{}>{\raggedright\arraybackslash}p{0.12\textwidth}>{\raggedright\arraybackslash}p{0.19\textwidth}>{\raggedright\arraybackslash}p{0.085\textwidth}>{\raggedright\arraybackslash}p{0.26\textwidth}>{\raggedright\arraybackslash}X@{}}
\toprule
\textbf{Step and variation} & \textbf{Evidence and current limit} & \textbf{Binding trade-offs} & \textbf{Four-pillar priorities} & \textbf{Decisive evaluation} \\
\midrule
\textbf{S1: Single game}\newline Vary goals and task chains within one environment.
& GameHorizon~\citep{gamehorizon2026}: its strongest tested model completes $9/20$ Minecraft chains (45\%) versus $41/62$ diagnostic subtasks (66.1\%); intermediate resets do not count as recovery.
& T1, T3, T4
& \textbf{D:} failure and recovery trajectories. \textbf{M:} persistent memory and temporal control. \textbf{H:} event-driven planning and recovery. \textbf{B:} held-out goal chains and seeds.
& A frozen system meets the completion target on held-out chains and seeds without oracle resets. Report whole-chain completion, recovery, and latency under equal interaction and wall-clock budgets. \\
\addlinespace[6pt]
\textbf{S2: Related games}\newline Shared interaction structure; new content and mechanics.
& OmniGameArena~\citep{omnigamearena2026,omnigamearenaWeb2026}: learned skills transfer unevenly to held-out variants. These vary one game; they are not new related-game titles.
& T1, T2, T4
& \textbf{D:} separate level, mechanic, and title splits. \textbf{M:} disentangle rules from layouts. \textbf{H:} fixed adapters and skill selection. \textbf{B:} matched source/variant/title protocols.
& Freeze the learned skill before fresh tests. Require high held-out performance, not only practice gain; report source retention. Extend from variants to disjoint related titles without per-title retuning. \\
\addlinespace[6pt]
\textbf{S3: Distinct games}\newline Vary observations, actions, or dynamics, including unseen titles.
& GameVerse~\citep{gameverse69} and GameWorld~\citep{GameWorld}: cross-game evaluation and interface comparisons. Coverage of titles alone establishes neither unseen-game transfer nor expert competence.
& T1--T4; T5 if synthetic data are used
& \textbf{D:} balanced, auditable game splits. \textbf{M:} transferable action semantics. \textbf{H:} matched interfaces and memory. \textbf{B:} per-game scores, compatible anchors, and clock controls.
& One frozen system meets the coverage target on disjoint games. Report uncertainty and interface privileges; separate paused and real-time tracks. Attribute breadth gains only through matched model/data-budget ablations. \\
\addlinespace[6pt]
\textbf{S4: Adaptation}\newline Unknown games; bounded interaction with fixed weights and harness rules.
& GameVerse~\citep{gameverse69}: same-task video-reflection retries. OmniGameArena IDC~\citep{omnigamearenaWeb2026}: bounded skill updates. Neither establishes cross-game retention.
& T1--T4
& \textbf{D:} exploration and feedback traces. \textbf{M:} in-context belief updates. \textbf{H:} budgeted reflection and memory. \textbf{B:} fresh tests, selection controls, and retention.
& Compare before/after a fixed practice allowance, then test fresh episodes and prior games. Count reflector/judge cost; declare carried memory. Require retained competence and post-adaptation coverage. Separate weight-updating training. \\
\addlinespace[6pt]
\textbf{S5: Generated worlds}\newline Joint environment and policy design; prospective.
& None of these four benchmarks tests environment generation or a synthetic-training loop. Their executable games could provide external checks in a future extension.
& T1--T5
& \textbf{D:} validated synthetic trajectories. \textbf{M:} persistent dynamics. \textbf{H:} independent rule/reward checkers. \textbf{B:} held-out rule sets and code-engine transfer.
& Verify declared dynamics invariants and rewards with executable checks. At matched training cost, synthetic-data agents improve held-out code-engine coverage over a no-synthetic-data control; audit repeatability and reward exploitation. \\
\bottomrule
\end{tabularx}
\end{table*}

\subsubsection{Step 1: Single-Game Task Mastery}

\epigraph{\textit{``Complete all tasks within a single game, from atomic actions to long-horizon objectives.''}}

In developmental psychology \citep{fitts1967human,Anderson1982AcquisitionOC}, skill acquisition within a single rule system is the foundation of all higher-order transfer. A child must first master the rules, controls, and feedback loops of one game before any cross-game generalization becomes meaningful. For AI agents, this corresponds to achieving robust competence across the full task distribution of a single game environment, including both short-horizon atomic skills and long-horizon composite objectives that chain these skills over hundreds of sequential decisions.     
         
Minecraft has a substantial data-model ecosystem, including MineRL \citep{minerl}, MineDojo \citep{minedojo52}, VPT \citep{vpt}, GROOT \citep{groot51}, OpenHA \citep{openha}, JARVIS-VLA \citep{jarvis_vla}, and PLAICraft \citep{plaicraft}. GameHorizon \citep{gamehorizon2026} nevertheless exposes a gap between executing subtasks and completing their composition. Its strongest tested model, GPT-6-Astra, passes $41/62$ subtasks (66.1\%) but only $9/20$ long Minecraft tasks (45\%). The benchmark permits up to 400 model calls per subtask and, after a failure, supplies a successful intermediate state to diagnose subsequent subtasks. A whole task passes only without such failures or resets. Thus the two denominators measure different outcomes: diagnostic continuation is not evidence of autonomous recovery. The online test covers Minecraft, while its broader offline question-answering evaluation cannot establish interactive mastery of the other games.

The S1 gate in Table~\ref{tab:roadmap-tradeoffs} combines held-out goal coverage with long-horizon completion and recovery, rather than treating a collection of successful atomic actions as full-game mastery.

\subsubsection{Step 2: Cross-Task Transfer Within Genre}

\epigraph{\textit{``Transfer learned competence across games that share visual style, interface conventions, or interaction patterns.''}}

Analogical reasoning research \citep{Gentner1983StructureAT,Gick1983SchemaIA} shows that humans transfer skills most readily between domains that share surface features and relational structure. A player fluent in one open-world RPG navigates a new RPG almost immediately, because they share the similar interaction grammar. Step 2 tests whether AI agents can exploit this same structural overlap: given mastery of one game, can they generalize to a second game within the same genre without retraining?     

Lumine \citep{lumine} demonstrates the current frontier at this step. Trained on the Mondstadt region of Genshin Impact, it achieves zero-shot completion of the five-hour main storyline in Honkai: Star Rail and a 100-minute mission in Wuthering Waves. Its 5Hz perception and 30Hz action pipeline operates in real time without game-specific adaptation, and its selective reasoning mechanism activates deliberation only when needed.

These demonstrations cover only selected content in games with recognizable interface conventions; they do not establish uniform transfer across mechanics. OmniGameArena's current project results \citep{omnigamearenaWeb2026} isolate a narrower test: transfer of a selected skill to four held-out variants of the same game. On LastStand, the best-round gains on the practiced game are $0.620$ for Opus~4.7 and $0.548$ for GPT-5.5. Yet the selected Opus skill loses performance against its no-skill baseline on all four variants, while the GPT skill improves on all four. Changes include tile-removal order, clustered removal, and player-targeted removal. These results motivate testing whether skills encode reusable rules rather than a successful local tactic. They are within-game variant tests, not evidence of transfer to a new related-game title.

The S2 test should separate familiar interface conventions from genuinely transferred mechanics, and report both held-out coverage and regressions on the source games under the same protocol.

\subsubsection{Step 3: Cross-Genre Generalization}

\epigraph{\textit{``Operate across games with fundamentally different action spaces, visual styles, time scales, and mechanics.''}}

This step corresponds to what game studies call "ludic literacy" \citep{juul2005half,salen2003rules}: the ability to parse an unfamiliar rule system by recognizing abstract patterns (resource management, spatial navigation, turn economy) beneath surface-level differences. A chess expert picking up a new strategy board game does not start from zero, because the underlying combinatorial reasoning transfers even when pieces, boards, and victory conditions change entirely. For AI, Step 3 tests a single system across games that differ in action semantics, observations, or dynamics, including titles excluded from its task-specific training. Shared representations must retain the distinctions needed to pursue each game's goals.

GameVerse \citep{gameverse69} and GameWorld \citep{GameWorld} make cross-game evaluation and the deployment interface explicit. GameWorld evaluates 170 tasks in 34 games with screenshot observations and either computer-use actions or semantic commands translated to atomic controls. In its current 400-step results \citep{gameworldLeaderboard2026}, Qwen3.8-Flash-Next reaches 36.86\% task success with computer-use actions versus 29.80\% with semantic commands. These are task-run completion rates under one benchmark protocol, not percentages of games exceeding expert-normalized competence. A shared interface makes comparison possible, but does not establish that the games were excluded from every model's pretraining.

The model and harness pillars interact even when the game set stays fixed. In GameWorld's original memory ablation \citep{GameWorld}, increasing the retained interaction history from zero to two rounds raises Qwen3-VL-235B-A22B's per-step latency in both interfaces, while mean progress changes from 30.0\% to 30.6\% with semantic commands and from 30.3\% to 28.7\% with computer-use actions. More context is therefore not uniformly beneficial in this comparison. This motivates paired tests of memory content, action representation, and cost; it does not identify the cause of differences between unrelated model families or validate transfer to unseen games.

For S3, the missing evidence is high coverage over a declared split of independently authored games under a shared protocol, including stable performance on the original games. Table~\ref{tab:roadmap-tradeoffs} specifies the required controls.

\subsubsection{Step 4: Lifelong Adaptation}

\epigraph{\textit{``Rapidly adapt to entirely new environments through self-directed exploration and continuous self-improvement.''}}

S2--S3 can already include held-out games; S4 adds a different question: how much can behavior improve from a bounded amount of interaction with a previously unknown game? This resembles ``learning to learn'' \citep{harlow1949formation,Flavell1979MetacognitionAC}, but two mechanisms must be separated. In-context adaptation changes beliefs, memory, or plans while the checkpoint and harness remain fixed. Parameter-updating continual training changes the policy itself and requires a separate accounting of training data, compute, and retention. Only the former directly satisfies the fixed-parameter evaluation in Section~\ref{sec:generalist-definition}.

GameVerse \citep{gameverse69} adds textual reflections on failure videos and expert tutorials to a subsequent attempt's prompt without updating model weights. In its three-game semantic-interface ablation, Qwen3-VL-32B's reported mean score rises from 61.0 without reflection to 76.7 with both sources. This is improvement on the same target tasks, not a held-out-game transfer or retention test. It motivates testing which experience helps and whether the benefit survives new tasks.

OmniGameArena's current IDC results \citep{omnigamearenaWeb2026} extend this question to repeated skill revision: 42 of 72 model--game curves improve by more than 0.1 at their best round, but 10 of those 42 finish more than 0.1 below that peak. The protocol uses ten rounds after a cold start, five episodes per round, and a reflector that revises the skill prompt while player weights remain frozen. Validation and skill-selection rules belong to the harness; their calls and memory must be counted in the adaptation budget. Selecting the best observed round requires fresh testing, because that maximum is not an unbiased estimate of future performance. Neither a favorable practice curve nor variant transfer demonstrates retained competence across prior games.

A decisive S4 experiment should compare the same checkpoint before and after a fixed interaction allowance, evaluate on fresh episodes, and revisit previously tested games under the same memory budget. Context resets and memory carried between games must be specified. Updated-checkpoint results can be valuable, but should be reported in a separate continual-training track, including its compute and any loss of earlier competence.

Thus, the S4 target is improvement with high post-adaptation coverage and retained competence, not merely a favorable average after an unspecified adaptation phase.

\subsubsection{Step 5: The Demiurge, From Player to Creator}

\epigraph{\textit{``Become the environment itself: generate, reshape, and evolve game worlds.''}}

The preceding steps concern agents that play within fixed game environments. Step 5 envisions a qualitative shift in the agent's ontological role. In game design theory \citep{huizinga1938homo,salen2003rules}, the distinction between player and designer is fundamental: a player optimizes within the rules; a designer constructs the rules themselves. This is also the distinction between intelligence that adapts to a given world and intelligence that creates worlds. Within the POMDP formalism, the agent no longer optimizes a policy within a predefined game $\mathcal{M}$, but actively generates the state space $\mathcal{S}$, action space $\mathcal{A}$, transition dynamics $\mathcal{T}$, and reward structures $\mathcal{R}$. This is the Era 4 Demiurge outlined in Figure \ref{fig:adaptive_scale}.

GameVerse, GameWorld, OmniGameArena, and GameHorizon do not evaluate this capability: they assess agents in supplied games or on recorded gameplay. Their executable environments could support external checks for a future generative extension, but the current results supply no S5 demonstration. Outside these four benchmarks, Genie 3 \citep{genie3-86} produces interactive environments, and SIMA 2 \citep{sima2-87} demonstrates activity and self-improvement within generated worlds. These are useful precursors, but they do not establish an agent that creates arbitrary rule systems with verifiable goals and then masters them.

S5 remains conditional on reliable generated dynamics, externally checkable rewards, and persistent state. Multi-agent generation, explored by Solaris \citep{solaris_wm}, adds cross-player consistency to these requirements. Its evaluation must distinguish environment quality from the competence of a player trained in that environment. The S5 gate in Table~\ref{tab:roadmap-tradeoffs} therefore requires independently checked rules and rewards plus transfer to held-out code engines. Without such checks, improved scores inside a generated world may reflect simulator artifacts rather than better gameplay.

The roadmap is not a proof that each step subsumes the previous ones: an environment generator need not be an expert player, and strong transfer need not imply rapid adaptation. Its purpose is to expose the evidence needed for each claim and the dependencies that must be addressed jointly across datasets, models, harnesses, and benchmarks.

\begingroup
\color{revisionblue}
\section{Conclusion}
\label{sec:conclusion}

\subsection{Summary of Contributions}

This survey connects the evolution of game-playing AI to the systems and evidence needed for generality. The goal-conditioned POMDP provides a standard vocabulary for identifying changes in goals, observations, action semantics, and dynamics across the four eras. Normalized competence and threshold coverage distinguish performance depth from breadth, while a declared train--test split is needed to establish transfer. The four-benchmark comparison in Section~\ref{sec:generality-comparison} makes these distinctions concrete: a higher mean need not imply broader competence, interface changes can alter reported performance, and local or practiced-task improvements need not yield reliable long-horizon execution or transfer.

The four pillars explain why these outcomes cannot be attributed to model scale alone. Data must cover decision-relevant interactions, models must transfer representations and control, and harnesses must translate these capabilities into reliable action under a budget. Benchmarks then need to distinguish improved reasoning from added information, stronger scaffolding, or more interaction time. Treating the deployed model and harness as the evaluated policy connects these engineering choices to the generality claim.

Finally, Table~\ref{tab:roadmap-tradeoffs} links the five research steps to all five trade-offs, current limitations, advances across the four pillars, and decisive evaluations. The steps are capability targets rather than a guaranteed chronological sequence: expertise does not establish broad transfer, weight-updating adaptation must be distinguished from fixed-weight inference, and generating convincing worlds does not by itself demonstrate competence within them.

\subsection{Limitations}

\textbf{Evidence and protocols remain heterogeneous.} We conduct no new gameplay or training experiments. The quantitative reanalysis recomputes descriptive statistics from published scores and inherits their prompts, interfaces, harnesses, budgets, and sampling uncertainty. Many recent systems are documented in preprints, technical reports, or release notes rather than independently reproduced studies. We therefore compare models within a shared benchmark where possible and retain each source's original metric elsewhere; the tables do not establish a universal leaderboard.

\textbf{Observed coverage is narrower than generalization.} Threshold coverage has an exact interpretation on a specified finite game list, but its estimate from reported mean returns is sensitive to rounding, thresholds, and task weights. Our OmniGameArena calculation uses published mean scores and reference rows on seven Solo games. Its human-anchor protocol is incompletely documented, and near-threshold means may change classification across repeated trials. The resulting counts are descriptive rather than confidence-certified. GameWorld and GameVerse retain their source-specific progress, success, or normalized metrics; GameHorizon's online results concern tasks within Minecraft, not a distribution over games. Broader transfer claims require representative game sampling, repeated trials, training provenance, and explicit held-out structural changes. Undisclosed pretraining can obscure whether a game is new to the model. A small train--test gap is insufficient without high test competence, and online weight updates require reporting adaptation cost and forgetting. For generated worlds, the generator must not also define the only success criterion; independent rule and solvability checks are needed before player scores are interpretable.

\textbf{Coverage and taxonomy are curatorial choices.} The corpus is expert-curated, with a primary cutoff of April 2026 and targeted four-pillar and four-benchmark updates through 21 October 2026, and restricted to English-language work; in an area where the frontier moves monthly, some of what we describe as current will not be by the time this is read. Our taxonomies are likewise organizing devices rather than partitions: model classes are cumulative rather than mutually exclusive, and the benchmark levels of Section~\ref{sec:benchmarks} order environments by the minimum model class required to be evaluated at all rather than by strict difficulty. Different cuts through the same literature are defensible.

\section*{Broader Impact Statement}

The capabilities surveyed here carry risks that follow directly from what makes them interesting, and we state the three we consider most immediate.

\textbf{Automated botting and cheating.} The defining capability of a generalist game player, reading a game from pixels and acting through the ordinary keyboard-and-mouse or gamepad interface, could also support automated cheating through ordinary client inputs. An agent that needs no memory injection, privileged API access, or modified client may evade some client-integrity checks, although behavioral detection remains possible. Progress along the roadmap of Section~\ref{sec:roadmap} therefore also constitutes progress toward tools that degrade competitive integrity and the economics of games that depend on it. We encourage researchers releasing such systems to consider release practices that do not lower the barrier for this use, for instance withholding turnkey deployment tooling for live commercial titles. We also note that behavioural rather than integrity-based detection is a research problem this community is well placed to work on.

\textbf{Consent and privacy in gameplay datasets.} Several of the largest corpora in Section 3 are assembled from streams, ladder replays, and web video. The players in that footage generally did not consent to its use as training data, and the recordings can carry identifying material beyond gameplay: voice, chat text, account names, and social graphs. Platform terms of service permitting collection are not the same as informed consent from the people recorded, and re-identification is often trivial where account handles persist. We think dataset releases in this area should document provenance and consent status explicitly, strip incidental personal content rather than ship it, and provide a removal mechanism for identifiable individuals.

\textbf{Dual use of general computer control.} An agent that operates arbitrary game interfaces from pixels is not confined to games. The perception--action interface may also support automation beyond games, but competence and safety in those applications require separate evaluation. That is part of why this setting is a useful proxy for general computer use, and also why these methods could be adapted to automate actions on systems their owners did not intend to expose to automation. We regard this as an argument for evaluating such agents on refusal and boundary behaviour, not only on competence, and for treating the game setting as a place where control failures can be studied at low cost before they arise in consequential ones.
\endgroup

\bibliography{main}
\bibliographystyle{tmlr}

\end{document}